\RequirePackage{fix-cm}
\documentclass[twocolumn]{svjour3}          
\usepackage{times}
\usepackage{textcomp}
\usepackage{epsfig}
\usepackage{graphicx}
\usepackage{amsmath}
\usepackage{amssymb}
\usepackage{multirow}
\usepackage{algorithm}
\usepackage{algpseudocode}
\usepackage{pifont}
\usepackage{xcolor}
\usepackage{rotating}
\usepackage{epstopdf}
\usepackage{blindtext}
\usepackage{tabularx}
\usepackage[titletoc,title]{appendix}
\usepackage{hyperref}
\hypersetup{colorlinks,linkcolor={blue},citecolor={blue},urlcolor={blue}}
\usepackage{url}
\usepackage{lmodern}
\usepackage{xspace}
\usepackage{pdflscape}
\usepackage{enumitem}

\usepackage{amssymb}
\usepackage{amsmath,amsfonts}
\usepackage{amsopn}
\usepackage{bm} 

\newcommand{\vct}[1]{\boldsymbol{#1}} 
\newcommand{\mat}[1]{\boldsymbol{#1}} 
\newcommand{\cst}[1]{\mathsf{#1}}  

\newcommand{\field}[1]{\mathbb{#1}}
\newcommand{\R}{\field{R}} 
\newcommand{\I}{\field{I}} 
\newcommand{\T}{^{\textrm T}} 

\newcommand{\twonorm}[1]{\left\|#1\right\|_2^2}

\newcommand{\ProbOpr}[1]{\mathbb{#1}}

\newcommand{\expect}[2]{%
\ifthenelse{\equal{#2}{}}{\ProbOpr{E}_{#1}}
{\ifthenelse{\equal{#1}{}}{\ProbOpr{E}\left[#2\right]}{\ProbOpr{E}_{#1}\left[#2\right]}}} 
\newcommand{\var}[2]{%
\ifthenelse{\equal{#2}{}}{\ProbOpr{VAR}_{#1}}
{\ifthenelse{\equal{#1}{}}{\ProbOpr{VAR}\left[#2\right]}{\ProbOpr{VAR}_{#1}\left[#2\right]}}} 

\newcommand{\vtheta}{\vct{\theta}}

\newcommand{\vq}{\vct{q}}
\newcommand{\vx}{{\vct{x}}}

\newcommand{\vz}{{\vct{z}}}

\newcommand{\va}{\vct{a}}
\newcommand{\vb}{\vct{b}}

\newcommand{\vv}{\vct{v}}
\newcommand{\vw}{\vct{w}}

\newcommand{\vpsi}{\vct{\psi}}

\newcommand{\mD}{\mat{D}}
\newcommand{\mW}{\mat{W}}

\newcommand{\mM}{\mat{M}}

\newcommand{\mSigma}{\mat{\Sigma}}

\newcommand{\cN}{\cst{N}}

\newcommand{\cD}{\cst{D}}

\newcommand{\cR}{\cst{R}}
\newcommand{\cU}{\cst{U}}
\newcommand{\cS}{\cst{S}}

\newcommand{\eat}[1]{}

\newcommand{\bst}[1]{{{\color{red}\textbf{#1}}}}
\newcommand{\sbst}[1]{{{\color{blue}\textbf{#1}}}}
\definecolor{olive}{rgb}{0.0, 0.5, 0.0}
\newcommand{\tbst}[1]{{{\color{olive}\textbf{#1}}}}

\newcommand{\mtcl}[2]{\multicolumn{#1}{c|}{#2}}

\newcommand{\method}[1]{\textsc{#1}}
\newcommand{\dataset}[1]{\textbf{#1}}

\newcommand{\awa}{\dataset{AwA}\xspace}
\newcommand{\awatwo}{\dataset{AwA2}\xspace}

\newcommand{\cub}{\dataset{CUB}\xspace}
\newcommand{\sun}{\dataset{SUN}\xspace}

\newcommand{\imn}{\dataset{ImageNet}\xspace}

\newcommand{\zsldap}{\method{DAP} \cite{LampertNH14}\xspace}
\newcommand{\zsliap}{\method{IAP} \cite{LampertNH14}\xspace}
\newcommand{\zslale}{\method{ALE} \cite{AkataPHS13}\xspace}
\newcommand{\zslsje}{\method{SJE} \cite{AkataRWLS15}\xspace}
\newcommand{\zsleszsl}{\method{ESZSL} \cite{Bernardino15}\xspace}
\newcommand{\zslsse}{\method{SSE} \cite{ZhangS15}\xspace}

\newcommand{\zsldevise}{\method{DeViSE} \cite{FromeCSBDRM13}\xspace}
\newcommand{\zslconse}{\method{ConSE} \cite{NorouziMBSSFCD14}\xspace}
\newcommand{\zslbidilel}{\method{BiDiLEL} \cite{WangC17}\xspace}

\newcommand{\zsllatem}{\method{LatEm} \cite{XianASNHS16}\xspace}
\newcommand{\zslcca}{\method{CCA} \cite{Lu16}\xspace}
\newcommand{\zslsae}{\method{SAE} \cite{KodirovXG17}\xspace}
\newcommand{\zslgfzsl}{\method{GFZSL} \cite{VermaR17}\xspace}

\newcommand{\zslcmt}{\method{CMT} \cite{SocherGMN13}\xspace}

\newcommand{\zslhat}{\method{HAT} \cite{AlHalahS15}\xspace}

\newcommand{\zslcosta}{\method{COSTA} \cite{MensinkGS14}\xspace}

\newcommand{\zslsyncovo}{\method{SynC$^\textrm{o-vs-o}$}\xspace}
\newcommand{\zslsynccs}{\method{SynC$^\textrm{cs}$}\xspace}
\newcommand{\zslsyncstr}{\method{SynC$^\textrm{struct}$}\xspace}

\newcommand{\zslexemconse}{\method{EXEM (ConSE)}\xspace}

\newcommand{\zslexemeszsl}{\method{EXEM (ESZSL)}\xspace}
\newcommand{\zslexemsyncovo}{\method{EXEM (SynC$^\textrm{o-vs-o}$)}\xspace}
\newcommand{\zslexemsynccs}{\method{EXEM (SynC$^\textrm{cs}$)}\xspace}
\newcommand{\zslexemsyncstr}{\method{EXEM (SynC$^\textrm{struct}$)}\xspace}
\newcommand{\zslexemnn}{\method{EXEM (1NN)}\xspace}
\newcommand{\zslexemnns}{\method{EXEM (1NNs)}\xspace}

\newcommand{\tw}{us\xspace}

\newcommand{\mtmdl}{\cite{AlHalahS15}\xspace}

\newcommand{\bidilel}{\cite{WangC17}\xspace}
\newcommand{\gbu}{\cite{XianLSA17}\xspace}
\newcommand{\sync}{\cite{ChangpinyoCGS16}\xspace}
\newcommand{\exem}{\cite{ChangpinyoCS17}\xspace}
\newcommand{\gzsl}{\cite{ChaoCGS16}\xspace}

\journalname{International Journal of Computer Vision}

\begin{document}

\title{Classifier and Exemplar Synthesis for Zero-Shot Learning}



\author{Soravit Changpinyo*\and
	Wei-Lun Chao*\thanks{*\hspace{4pt} Equal contribution} \and
	Boqing Gong \and
	Fei Sha}


\institute{Soravit Changpinyo \at
	Google AI\\
	\email{schangpi@google.com}
	\and Wei-Lun Chao \at
	Cornell University, Department of Computer Science\\
	\email{weilunchao760414@gmail.com}
	\and Boqing Gong \at
	Tencent AI Lab\\
	\email{boqinggo@outlook.com}
	\and Fei Sha \at
	University of Southern California, Department of Computer Science\\
	\email{feisha@usc.edu}
}

\date{Received: date / Accepted: date}

\maketitle

\begin{abstract}
Zero-shot learning (ZSL) enables solving a task without the need to see its examples. In this paper, we propose two ZSL frameworks that learn to \emph{synthesize parameters} for novel unseen classes. First, we propose to cast the problem of ZSL as learning manifold embeddings from graphs composed of object classes, leading to a flexible approach that synthesizes ``classifiers" for the unseen classes. Then, we define an auxiliary task of synthesizing ``exemplars" for the unseen classes to be used as an automatic denoising mechanism for any existing ZSL approaches or as an effective ZSL model by itself. On five visual recognition benchmark datasets, we demonstrate the superior performances of our proposed frameworks in various scenarios of both conventional and generalized ZSL. Finally, we provide valuable insights through a series of empirical analyses, among which are a comparison of semantic representations on the full ImageNet benchmark as well as a comparison of metrics used in generalized ZSL. Our code and data are publicly available at \url{https://github.com/pujols/Zero-shot-learning-journal}.

\end{abstract}


\section{Introduction}
\label{sIntro}

Visual recognition has made a significant progress due to the widespread use of deep learning architectures~\cite{AlexNet,VGG,Inception,ResNet} that are optimized on large-scale datasets of human-labeled images~\cite{ILSVRC15}. Despite the exciting advances, to recognize objects ``in the wild'' remains a daunting challenge. In particular, the amount of annotation effort is vital to deep learning architectures in order to discover and exploit powerful discriminating visual features.

There are many application scenarios, however, where collecting and labeling training instances can be laboriously difficult and costly. For example, when the objects of interest are rare (e.g., only about a hundred of northern hairy-nosed wombats alive in the wild) or newly defined (e.g., images of futuristic products such as Tesla's Model Y), not only the number of labeled training images but also the statistical variation among them is limited. These restrictions prevent one from training robust systems for recognizing such objects. More importantly, the number of such objects could be significantly greater than the number of common objects. In other words, the frequencies of observing objects follow a long-tailed distribution~\cite{SalakhutdinovTT11,ZhuAR14,VanHornP17}.

Zero-shot learning (ZSL) has since emerged as a promising paradigm to remedy the above difficulties. Unlike supervised learning, ZSL distinguishes between two types of classes: \emph{seen} and \emph{unseen}. Labeled examples are only available for the seen classes whereas no (labeled or unlabeled) examples are available for the unseen ones. The main goal of zero-shot learning is to construct classifiers for the unseen classes, extrapolating from what we learned from the seen ones. To this end, we need to address two key interwoven challenges~\cite{PalatucciPHM09}: (1) how to relate unseen classes to seen ones and (2) how to attain optimal discriminative performance on the unseen classes even though we do not have access to their representative labeled data?

The first challenge can be overcome by the introduction of a shared semantic space that embeds all categories. Given access to this semantic space, zero-shot learners can exploit the \emph{semantic} relationship between seen and unseen classes to establish the \emph{visual} relationship. Multiple types of semantic information have been exploited in the literature: visual attributes \cite{FarhadiEHF09,LampertNH09}, word vector representations of class names \cite{FromeCSBDRM13,SocherGMN13,NorouziMBSSFCD14}, textual descriptions \cite{ElhoseinySE13,LeiSFS15,ReedKLS16}, hierarchical ontology of classes (such as WordNet \cite{Miller95}) \cite{AkataRWLS15,Lu16,XianASNHS16}, and human gazes \cite{KaressliABS17}.

The second challenge requires developing appropriate objectives or algorithmic procedures for ZSL. Many ZSL methods take a two-stage approach: (i) predicting the embedding of a visual input in the semantic space;  (ii) inferring the class labels by comparing the embedding to the unseen classes' semantic representations~\cite{FarhadiEHF09,LampertNH09,PalatucciPHM09,SocherGMN13,YuCFSC13,JayaramanG14,NorouziMBSSFCD14,Lu16}. More recent ZSL methods take a unified approach by jointly learning the functions to predict the semantic embeddings as well as to measure similarity in the embedding space~\cite{AkataPHS13,AkataRWLS15,FromeCSBDRM13,Bernardino15,ZhangS15,ZhangS16}. We refer the readers to Sect.~\ref{sRelated} and recent survey articles by \cite{XianSA17,XianLSA17,FuXJXSG18} for the descriptions and comparison of these representative methods.

In this paper, we propose two zero-shot learning frameworks, where the major common theme is to learn to ``synthesize" representative parameters --- a ``summary" for the unseen classes. One natural choice of such parameters are ``classifiers" that, as the name suggests, can be used to recognize object classes in a straightforward manner\footnote{In this work, classifiers are taken to be the normals of hyperplanes separating different classes (i.e., linear classifiers).}. Other choices of class summaries exist but additional steps may be needed to perform zero-shot recognition. We explore one such choice and define ``visual exemplars" as (average) dimensionality-reduced visual features of different classes. We learn to predict these exemplars and then use them to perform zero-shot recognition in two different manners. Below, we describe our concrete implementations of both frameworks.

In the first framework of \method{Syn}thesized \method{C}lassifiers (\method{SynC}; Fig.~\ref{fConceptSynC}), we take ideas from manifold learning~\cite{HintonR02,BelkinN03} and cast zero-shot learning as a graph alignment problem. On one end, we view the object classes in a semantic space as a weighted graph where the nodes correspond to object class names and the weights of the edges represent how much they are related. Semantic representations can be used to infer those weights. On the other end, we view models or classifiers for recognizing images of those classes as if they live in a space of models. The parameters for each object model are nothing but coordinates in this model space whose geometric configuration also reflects the relatedness among objects. To reduce the complexity of the alignment, we introduce a set of \emph{phantom} object classes --- interpreted as bases (classifiers) --- from which a large number of classifiers for real classes can be synthesized. In particular, the model for any real class is a convex combination of the coordinates of those phantom classes. Given these components, we learn to synthesize the classifier weights (i.e., coordinates in the model space) for the unseen classes via convex combinations of adjustable and optimized phantom coordinates and with the goal of preserving their semantic graph structures.

In the other framework of \method{EXEM}plar synthesis (\method{EXEM}; Fig.~\ref{fConceptExem}), we first define visual exemplars as target summaries of object classes and then learn to predict them from semantic representations. We then propose two ways to make use of these predicted exemplars for zero-shot recognition. One way is to use the exemplars as improved semantic representations in a separate zero-shot learning algorithm. This is motivated by the evidence that existing semantic representations are barely informative about \emph{visual} relatedness (cf. Sect.~\ref{sApproachEXEM}). Moreover, as the predicted visual exemplars already live in the visual feature space, we also use them to construct nearest-neighbor style classifiers, where we treat each of them as a data instance.

Our empirical studies extensively test the effectiveness of different variants of our approaches on five benchmark datasets for conventional and four for generalized zero-shot learning. We find that \method{SynC} performs competitively against many strong baselines. Moreover, \method{EXEM} enhances not only the performance of \method{SynC} but also those of other ZSL approaches. In general, we find that \method{EXEM}, albeit simple, is overall the most effective ZSL approach and that both \method{SynC} and \method{EXEM} achieve the best results on the large-scale ImageNet benchmark.

We complement our studies with a series of analysis on the effect of types of semantic representations and evaluation metrics on zero-shot classification performance. We obtain several interesting results. One is from an empirical comparison between the metrics used in generalized zero-shot learning; we identify shortcomings of the widely-used \emph{uncalibrated} harmonic mean and recommend that the \emph{calibrated} harmonic mean or the Area under Seen-Unseen Accuracy curve (AUSUC) be used instead. Another interesting result is that we obtain higher-quality semantic representations and use them to establish the new state-of-the-art performance on the large-scale ImageNet benchmark. Finally, based on the idea in \method{EXEM}, we investigate how much the ImageNet performance can be improved by \emph{ideal} semantic representations and see a large gap between those results and existing ones obtained by our algorithms.

This work unifies and extends our previously published conference papers \cite{ChangpinyoCGS16,ChangpinyoCS17}. Firstly, we unify our ZSL methods \method{SynC} and \method{EXEM} using the ``synthesis" theme, providing more consistent terminology, notation, and figures as well as extending the discussion of related work. Secondly, we provide more coherent experimental design and more comprehensive, updated results. Our experiments have been extended extensively to include results on an additional dataset (\awatwo\gbu), stronger visual features (ResNet), better semantic representations (our improved word vectors on ImageNet and ideal semantic representations), new and more rigorous training/validation/test data splits, recommended by \gbu, newly proposed metrics (per-class accuracy on ImageNet, AUSUC, uncalibrated and calibrated harmonic mean), additional variants of our methods, and additional baselines. We also provide a summarized comparison of ZSL methods (Sect.~\ref{sMainExp}). For more details on which results are newly reported by this work, please refer to our tables (``reported by us"). Thirdly, we extend our results and analysis on generalized ZSL. On selected multiple strong baselines, we provide empirical evidence of a shortcoming of the widely-used metric and propose its \emph{calibrated} version that is built on top of calibrated stacking \cite{ChaoCGS16}. Finally, we further empirically demonstrate the importance of high-quality semantic representations for ZSL, and establish upperbound performance on ImageNet in various scenarios of conventional ZSL.

The rest of the paper is organized as follows. We describe our classifier and exemplar systhesis frameworks in Sect.~\ref{sApproach}. We validate our approaches using the experimental setup in Sect.~\ref{sExpSetup} and present our results in Sect.~\ref{sExpRes}. We discuss related work in Sect.~\ref{sRelated}. Finally, we conclude in Sect.~\ref{sDiscuss}.


\section{Approach}
\label{sApproach}

We describe our methods for addressing (conventional) zero-shot learning, where the task is to classify images from unseen classes into the label space of unseen classes. We first describe, \method{SynC}, a manifold-learning-based method for synthesizing the classifiers of the unseen classes. We then describe, \method{EXEM}, an approach that automatically improves semantic representations through visual exemplar synthesis. \method{EXEM} can generally be combined with any zero-shot learning algorithms, and can by itself operate as a zero-shot learning algorithm.  

\paragraph{Notation:}
We denote by $\mathcal{D}= \{(\vx_n\in \R^{\cD},y_n)\}_{n=1}^\cN$ the training data with the labels coming from the label space of \emph{seen} classes $\mathcal{S} = \{1,2,\cdots,\cS\}$. Denote by $\mathcal{U} = \{\cS+1,\cdots,\cS+\cU\}$ the label space of \emph{unseen} classes. Let $\mathcal{T} = \mathcal{S} \cup \mathcal{U}$. For each class $c \in \mathcal{T}$, we assume that we have access to its semantic representation $\va_c$.

\subsection{Classifier Synthesis}
\label{sApproachSynC}

\begin{figure*}
\centering
\includegraphics[width=0.85\textwidth]{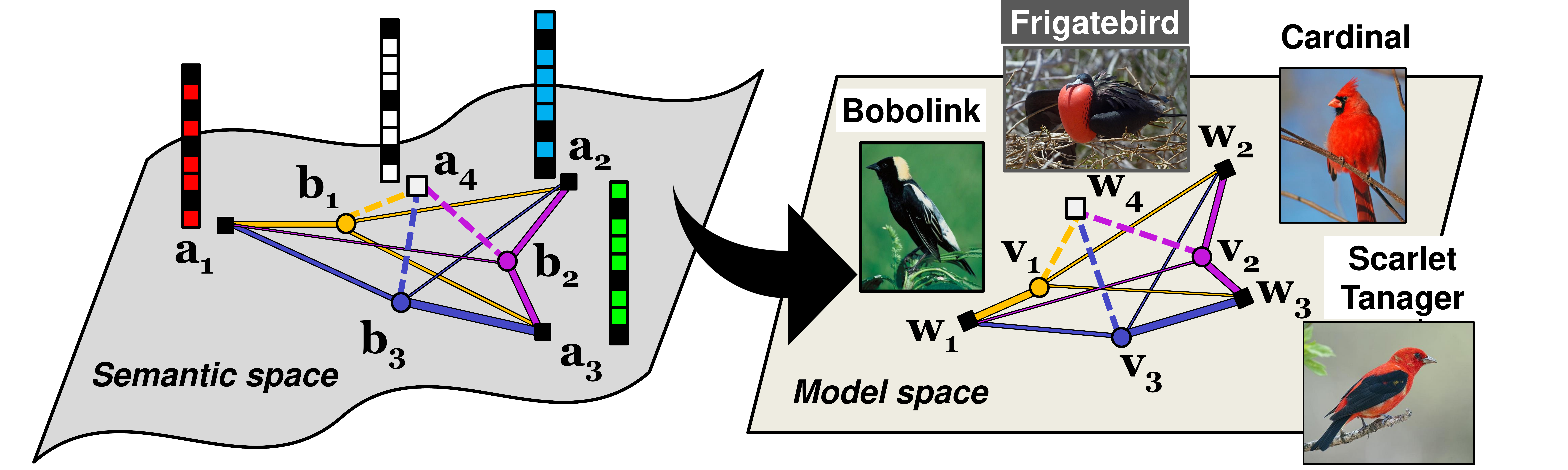}
\caption{Illustration of \method{SynC} for zero-shot learning. Object classes live in two spaces. They are characterized in the semantic space with semantic representations ($\va_s$) such as attributes or word vectors of their names. They are also represented as models for visual recognition ($\vw_s$) in the model space. In both spaces, those classes form weighted graphs. The main idea behind our approach is that these two spaces should be aligned. In particular, the coordinates in the model space should be the projection of the graph vertices from the semantic space to the model space --- preserving class relatedness encoded in the graph. We introduce adaptable phantom classes ($\vb$ and $\vv$) to connect seen ({black text: Bobolink, Cardinal, Scarlet Tanager}) and unseen ({white text: Frigatebird}) classes --- classifiers for the phantom classes are bases for synthesizing classifiers for real classes, including both the seen and unseen ones. In particular, the synthesis takes the form of convex combination. {We learn the phantom classes using seen classes' data (Sect.~\ref{sSynCLearnPhantom}), which are then used to synthesize unseen classes' classifiers (Sect.~\ref{sSynCSynthesis}).}}
\label{fConceptSynC}
\end{figure*}

We propose a zero-shot learning method of synthesized classifiers, called \method{SynC}. We focus on linear classifiers in the visual feature space $\R^\cD$ that assign a label $\hat{y}$ to a data point $\vx$ by
\begin{align}
\hat{y} = \arg\max_{c} \quad \vw_c\T\vx, 
\label{eSynCPredict}
\end{align}
where $\vw_c \in \R^{\cD}$, although our approach can be readily extended to nonlinear settings by the kernel trick~\cite{ScholkopfS02}\footnote{In the context of deep neural networks for classification, one can think of $\vw_c$ as the vector corresponding to class $c$ in the last fully-connected layer and $\vx$ as the input to that layer.}.

\subsubsection{Main idea: manifold learning}
\label{sSyncMainIdea}

The main idea behind our approach is to align the semantic space and the model space. The semantic space coordinates of objects are designated or derived based on external information (such as textual data) that do not directly examine visual appearances at the lowest level, while the model space concerns itself largely for recognizing low-level visual features. To align them, we view the coordinates in the model space as the projection of the vertices on the graph from the semantic space --- there is a wealth of literature on manifold learning for computing (low-dimensional) Euclidean space embeddings from the weighted graph, for example, the well-known algorithm of Laplacian eigenmaps~\cite{BelkinN03}.

This idea is shown by the conceptual diagram in Fig.~\ref{fConceptSynC}. Each class $c$ has a coordinate $\va_c$ and they live on a manifold in the semantic representation space. We use attributes to illustrate the idea here but in the experiments we test our approach on multiple types of semantic representations. Additionally, we introduce a set of \emph{phantom} classes associated with semantic representations $\vb_r, r =1, 2, \ldots, \cR$.  We stress that they are phantom as they themselves do \textbf{not} correspond to any real objects --- they are introduced to increase the modeling flexibility, as shown below.

The real and phantom classes form a weighted bipartite graph, with the weights defined as
\begin{align}
s_{cr} = \frac{\exp\{-d(\va_c,\vb_r)\}}{\sum_{r=1}^{\cR}\exp\{-d(\va_c,\vb_r)\}}
\label{eSynCSimForm}
\end{align}
to relate a real class $c$ and a phantom class $r$, where 
\begin{align}
d(\va_c,\vb_r)=(\va_c-\vb_r)^T\mSigma^{-1}(\va_c-\vb_r), \label{eSynCDist}
\end{align}
and $\mSigma^{-1}$ is a parameter that can be learned from data, modeling the correlation among attributes. For simplicity, we set $\mSigma = \sigma^2\mat{I}$ and tune the scalar, free hyper-parameter $\sigma$ by cross-validation (Appendix \ref{sSuppHyperTuning}).

The specific form of defining the weights is motivated by several manifold learning methods such as SNE~\cite{HintonR02}. In particular, $s_{cr}$ can be interpreted as the conditional probability of observing class $r$ in the neighborhood of class $c$. However, other forms can be explored and are left for future work.

In the model space, each real class is associated with a classifier $\vw_c$ and the phantom class $r$ is associated with a virtual classifier $\vv_r$. We align the semantic and the model spaces by viewing $\vw_c$ (or $\vv_r$) as the embedding of the weighted graph. In particular, we appeal to the idea behind Laplacian eigenmaps~\cite{BelkinN03}, which seeks the embedding that maintains the graph structure as much as possible. Equivalently, the distortion error
\begin{align}
\| \vw_c - \sum_{r=1}^{\cR} s_{cr} \vv_r\|_2^2
\end{align}
with respect to $\vw_c, \vv_r$ is minimized.  This objective has an analytical solution 
\begin{align}
\vw_c = \sum_{r=1}^\cR s_{cr}\vv_r, \quad \forall\, c\in\mathcal{T}=\{1,2,\cdots,\cS+\cU\}. \label{eSynCCombine}
\end{align}
In other words, the solution gives rise to the idea of \emph{synthesizing classifiers} from those virtual classifiers $\vv_r$. For conceptual clarity, from now on we refer to $\vv_r$ as base classifiers in a dictionary from which new classifiers can be synthesized. We identify several advantages. First,  we could construct an infinite number of classifiers as long as we know how to compute $s_{cr}$. Second, by  making $\cR\ll\cS$, the formulation can significantly reduce the learning cost as we only need to learn $\cR$ base classifiers.

\subsubsection{Learning phantom classes}
\label{sSynCLearnPhantom}

\paragraph{Learning base classifiers:} We learn the base classifiers $\{\vv_r\}_{r=1}^\cR$ from the training data (of the seen classes only). We experiment with two settings. To learn one-versus-other classifiers, we optimize,
\begin{align}
&\min_{\vv_1,\cdots,\vv_\cR} \sum_{c=1}^{\cS}\sum_{n=1}^\cN \ell({\vx_n}, \I_{y_n,c}; {\vw_c}) + \frac{\lambda}{2} \sum_{c=1}^{\cS} \twonorm{\vw_c}, \label{eSynCObj} \\
&\mathsf{s.t.}\quad \vw_c = \sum_{r=1}^\cR s_{cr}\vv_r, \quad \forall\, c\in\mathcal{S}=\{1,\cdots,\cS\},  \notag
\end{align}
where $\ell(\vx, y; \vw)=\max(0,1-y\vw\T\vx)^2$ is the squared hinge loss. The indicator $\I_{y_n,c}\in\{-1,1\}$ denotes whether or not $y_n=c$. Alternatively, we apply the Crammer-Singer multi-class SVM loss~\cite{CrammerS02}, given by
\begin{align}
&\ell_\text{cs}(\vx_n,\hspace{2pt}y_n; \{\vw_c\}_{c=1}^{\cS})  \\
= &\max (0, \max_{c\in \mathcal{S}-\{y_n\}} \Delta(c,y_n) + {\vw_c}\T{\vx_n} - {\vw_{y_n}}\T{\vx_n}). \notag
\end{align}
We have the standard Crammer-Singer loss when the structured loss $\Delta(c,y_n)=1$ if $c\neq y_n$, which ignores the semantic relatedness between classes. We additionally  use the $\ell_2$ distance for the structured loss $\Delta(c,y_n)=\left\|\va_c-\va_{y_n}\right\|_2$ to exploit the class relatedness in our experiments.
These two learning settings have separate strengths and weaknesses in our empirical studies.

\paragraph{Learning semantic representations:}
The weighted graph (Eq.~(\ref{eSynCSimForm})) is also parameterized by adaptable embeddings of the phantom classes $\vb_r$. For simplicity, we assume that each of them is a sparse linear combination of the seen classes' attribute vectors:
\begin{equation}
\vb_r = \sum_{c=1}^\cS\beta_{rc}\va_c,\forall r\in\{1,\cdots,\cR\}.
\end{equation}
Thus, to optimize those embeddings, we solve the following optimization problem
\begin{align}
&\min_{\{\vv_r\}_{r=1}^\cR,
\{\beta_{rc}\}_{r,c=1}^{\cR, \cS}} \sum_{c=1}^{\cS}\sum_{n=1}^\cN \ell({\vx_n}, \I_{y_n,c}; {\vw_c}) \label{eSynCSemObj} \\
&
+\frac{\lambda}{2} \sum_{c=1}^{\cS} \twonorm{\vw_c}
+\eta\sum_{r,c=1}^{\cR,\cS}|\beta_{rc}|
+\frac{\gamma}{2}\sum_{r=1}^{\cR} (\twonorm{\vb_r}-h^2)^2\notag,\\
&\mathsf{s.t.}\quad \vw_c = \sum_{r=1}^\cR s_{cr}\vv_r, \quad \forall\, c\in\mathcal{S}=\{1,\cdots,\cS\}  \notag,
\end{align}
where $h$ is a predefined scalar equal to the norm of real attribute vectors (i.e., 1 in our experiments since we perform $\ell_2$ normalization).
Note that in addition to learning $\{\vv_r\}_{r=1}^\cR$, we learn combination weights $\{\beta_{rc}\}_{r, c=1}^{\cR, \cS}.$
Clearly, the constraint together with the third term in the objective encourages the sparse linear combination of the seen classes' attribute vectors. The last term in the objective demands that the norm of $\vb_r$ is not too far from the norm of $\va_c$.

We perform alternating optimization for minimizing the objective function with respect to $\{\vv_r\}_{r=1}^\cR$ and $\{\beta_{rc}\}_{r,c=1}^{\cR, \cS}$. While this process is nonconvex, there are useful heuristics to initialize the optimization routine.  For example, if $\cR = \cS$, then the simplest setting is to let $\vb_r = \va_r$ for $r = 1, \ldots, \cR$. If  $\cR\le\cS$, we can let them be (randomly) selected from the seen classes' attribute vectors $\{\vb_1, \vb_2, \cdots, \vb_\cR\} \subseteq \{\va_1, \va_2, \cdots, \va_\cS\}$, or first perform clustering on $\{\va_1, \va_2, \cdots, \va_\cS\}$ and then let each $\vb_r$ be a combination of the seen classes' attribute vectors in cluster $r$. If $\cR > \cS$, we could use a combination of the above two strategies\footnote{In practice, we found these initializations to be highly effective --- even keeping the initial $\vb_r$ intact while only learning $\vv_r$ for $r = 1,\ldots,\cR$ can already achieve comparable results. In most of our experiments, we thus only learn $\vv_r$ for $r = 1,\ldots,\cR$.}. There are four hyper-parameters $\lambda, \sigma, \eta,$ and $\gamma$ to be tuned. To reduce the search space during cross-validation, we first tune $\lambda, \sigma$ while fixing $\vb_r$ for $r = 1,\ldots,\cR$ to the initial values as mentioned above. We then fix $\lambda$ and $\sigma$ and tune $\eta$ and $\gamma$.

\subsubsection{Zero-shot classification with synthesized classifiers}
\label{sSynCSynthesis}
{
Given the attribute vectors $\{\va_c\}_{c=\cS+1}^{\cS+\cU}$ of $\cU$ unseen classes, we synthesize their classifiers $\{\vw_c\}_{c=\cS+1}^{\cS+\cU}$ according to Eq.~(\ref{eSynCCombine}) and Eq.~(\ref{eSynCSimForm}) using the learned phantom classes $\{(\vv_r, \vb_r)\}_{r=1}^\cR$ from Eq.~(\ref{eSynCSemObj}):
\begin{align}
& \vw_c = \sum_{r=1}^\cR s_{cr}\vv_r, \quad \forall\, c\in\mathcal{U}=\{\cS+1,\cdots,\cS+\cU\}, \label{eSynCFinalCla}\\
& \mathsf{s.t.}\quad s_{cr} = \frac{\exp\{-d(\va_c,\vb_r)\}}{\sum_{r=1}^{\cR}\exp\{-d(\va_c,\vb_r)\}}. \nonumber
\end{align}
That is, we apply the exact same rule to synthesize classifiers $\vw_c$ for both seen and unseen classes.

During testing, as in Eq. (\ref{eSynCPredict}), we then classify $\vx$ from unseen classes into the label space $\mathcal{U}$ by
\begin{align}
\hat{y} = \arg\max_{c\;\in\;\mathcal{U}}\quad{\vw_c}\T{\vx}.
\label{eSynCPredictNormal}
\end{align}
}
\subsection{Exemplar Synthesis}
\label{sApproachEXEM}

\begin{figure*}
\centering
\includegraphics[width=0.85\textwidth]{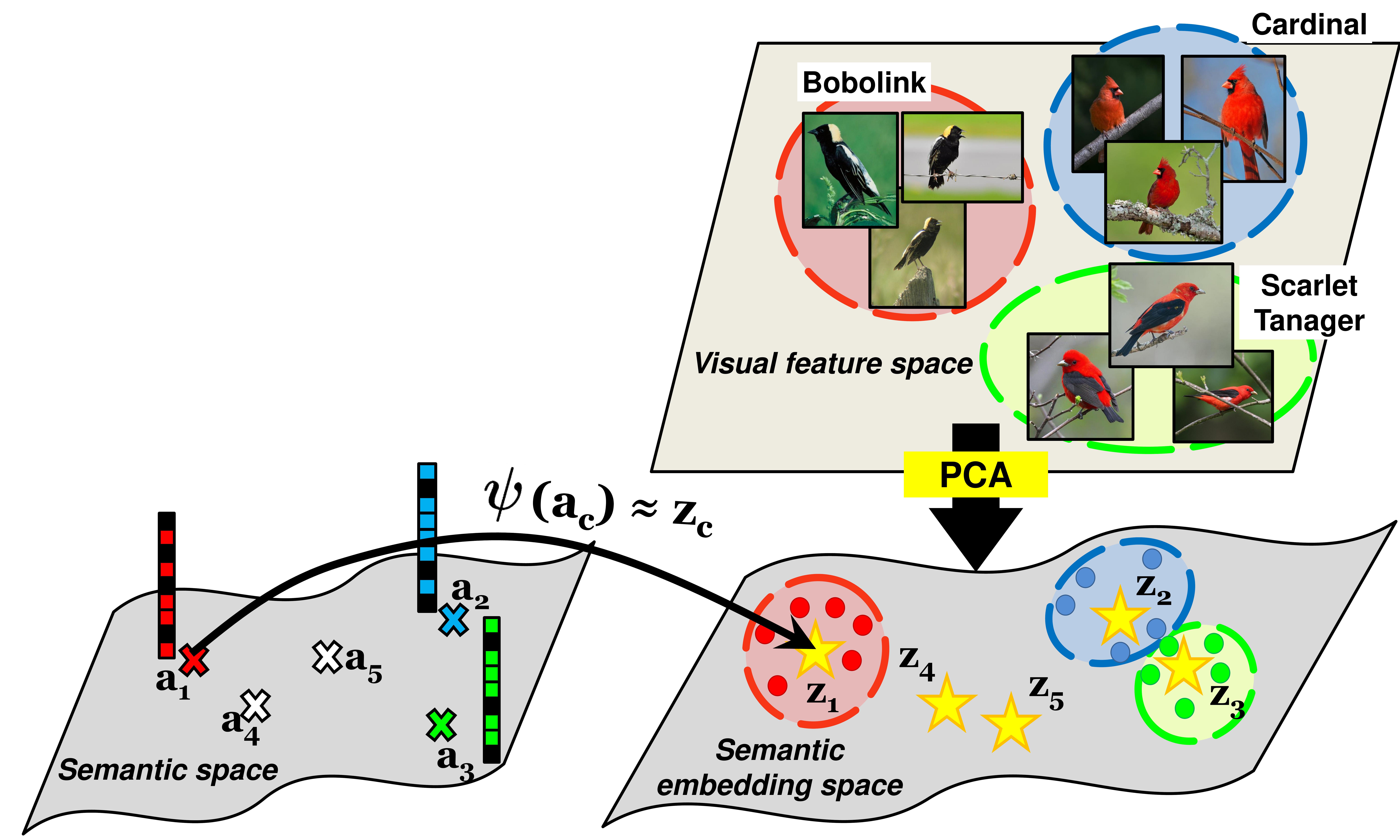}
\caption{Illustration of our method \method{EXEM} for improving semantic representations as well as for zero-shot learning. Given semantic information and visual features of the seen classes, we learn a \textbf{kernel-based regressor $\vpsi(\cdot)$} such that the semantic representation $\va_c$ of class $c$ can \textbf{predict well} its visual exemplar (center) $\vz_c$ that characterizes the clustering structure. The learned $\vpsi(\cdot)$ can be used to predict the visual feature vectors of unseen classes for nearest-neighbor (NN) classification, or to improve the semantic representations for existing ZSL approaches.}
\label{fConceptExem}
\end{figure*}

The previous subsection describes, \method{SynC}, an approach for synthesizing the classifiers of the unseen classes in zero-shot learning. \method{SynC} preserves graph structures in the semantic representation space. This subsection describes another route for constructing representative parameters for the unseen classes. We define the \emph{visual exemplar} of a class to be the target ``cluster center" of that class, characterized by the average of visual feature vectors. We then learn to predict the object classes' visual exemplars.

One motivation for this is the evidence that class semantic representations are hard to get right. While they may capture high-level \emph{semantic} relationships between classes, they are not well-informed about \emph{visual} relationships. For example, visual attributes are human-understandable so they correspond well with our object class definition. However, they are not always discriminative \cite{ParikhG11a,YuCFSC13}, not necessarily machine detectable \cite{DuanPCG12,JayaramanG14}, often correlated among themselves (``brown" and ``wooden") \cite{JayaramanSG14}, and possibly not category-independent (``fluffy" animals and ``fluffy" towels) \cite{ChenG14}. Word vectors of class names have been shown to be inferior to attributes \cite{AkataRWLS15,ChangpinyoCGS16}. Derived from texts, they have little knowledge about or are barely aligned with visual information. Intuitively, this problem would weaken zero-shot learning methods that rely heavily on \emph{semantic} relationships of classes (such as \method{SynC}).

We therefore propose the method of predicting visual exemplars (\method{EXEM}) to transform the (original) semantic representations into \emph{semantic embeddings} in another space to which visual information is injected. More specifically, the main computation step of \method{EXEM} is reduced to learning (from the seen classes) a predictive function from semantic representations to their corresponding centers of visual feature vectors. This function is used to predict the locations of visual exemplars of the unseen classes. Once predicted, they can be effectively used in any zero-shot learning algorithms as improved semantic representations. For instance, we could use the predicted visual exemplars in \method{SynC} to alleviate its naive reliance on the object classes' semantic representations. As another example, as the predicted visual exemplars live in the visual feature space, we could use them to construct nearest-neighbor style classifiers, where we treat each of them as a data instance.

Fig.~\ref{fConceptExem} illustrates the conceptual diagram of our approach. Our two-stage approach for zero-shot learning consists of learning a function to predict visual exemplars from semantic representations (Sect.~\ref{sEXEMPredExem}) and then apply this function to perform zero-shot learning given novel semantic representations (Sect.~\ref{sEXEMZsl}).

\subsubsection{Learning a function to predict visual exemplars from semantic representations}
\label{sEXEMPredExem}

For each class $c$, we would like to find a transformation function $\vpsi(\cdot)$ such that $\vpsi(\va_c) \approx \vz_c$, where $\vz_c \in \R^\cst{d}$ is the visual exemplar for the class. 
In this paper, we create the visual exemplar of a class by averaging the PCA projections of data belonging to that class. That is, we consider $\vz_c = \frac{1}{|I_c|}\sum_{n \in I_c} \mM \vx_n$, where $I_c = \{i: y_i = c\}$ and $\mM \in \R^{\cst{d} \times \cD} $ is the PCA projection matrix computed over training data of the seen classes. We note that $\mM$ is fixed for all data points (i.e., not class-specific) and is used in Eq. (\ref{eExemPredict}).

Given training visual exemplars and semantic representations, we learn $\cst{d}$ support vector regressors (SVR) with the RBF kernel --- each of them predicts each dimension of visual exemplars from their corresponding semantic representations. Specifically, for each dimension $d = 1, \ldots, \cst{d}$, we use the $\nu$-SVR formulation \cite{ScholkopfSWB00}.
\begin{align}
\min_{\vq, \xi, \xi', \epsilon} \frac{1}{2} & \vq^T\vq + \lambda(\nu \epsilon + \frac{1}{\cS} \sum_{c = 1}^{\cS} (\xi_c + \xi'_c)) \nonumber\\
\mathsf{s.t.} & \vq^T \vtheta^{\mathsf{rbf}}(\va_c)  -  \vz_{c} \leq \epsilon + \xi_c \label{eEXEMlearn}\\
& \vz_{c} - \vq^T \vtheta^{\mathsf{rbf}}(\va_c)  \leq \epsilon + \xi'_c \nonumber\\
& \xi_c \geq 0, \xi'_c \geq 0, \nonumber
\end{align}
where $\vtheta^{\mathsf{rbf}}$ is an implicit nonlinear mapping based on the RBF kernel. We have dropped the subscript $d$ for aesthetic reasons but readers are reminded that each regressor is trained independently with its own parameters. $\lambda$ and $\nu \in (0,1]$ (along with hyper-parameters of the kernel) are the hyper-parameters to be tuned. The resulting $\vpsi(\cdot) = [\vq_1^{T}\vtheta^{\mathsf{rbf}}(\cdot), \cdots, \vq_\cst{d}^{T}\vtheta^{\mathsf{rbf}}(\cdot)]^T$, where $\vq_d$ is from the $d$-th regressor.

Note that the PCA step is introduced for both computational and statistical benefits. In addition to reducing dimensionality for faster computation, PCA decorrelates the dimensions of visual features such that we can predict these dimensions independently rather than jointly.

\subsubsection{Zero-shot classification based on predicted visual exemplars}
\label{sEXEMZsl}
Now that we learn the transformation function $\vpsi(\cdot)$, how do we use it to perform zero-shot classification? We first apply $\vpsi(\cdot)$ to all semantic representations $\va_u$ of the unseen classes. We then consider two main approaches that depend on how we interpret these predicted exemplars $\vpsi(\va_u)$. 

\paragraph{Predicted exemplars as training data:}
An obvious approach is to use $\vpsi(\va_u)$ as data directly. Since there is only one data point per class, a natural choice is to use a nearest neighbor classifier. Then, the classifier outputs the label of the closest exemplar for each novel data point $\vx$ that we would like to classify:
\begin{align}
\hat{y} = \arg\min_{u} \quad \textrm{dis}_{NN}(\mM\vx, \vpsi(\va_u)), 
\label{eExemPredict}
\end{align}
where we adopt the Euclidean distance or the standardized Euclidean distance as $\textrm{dis}_{NN}$ in the experiments.

\paragraph{Predicted exemplars as improved semantic representations:}

The other approach is to use $\vpsi(\va_u)$ as the \emph{improved} semantic representations (``improved" in the sense that they have knowledge about visual features) and plug them into any existing zero-shot learning framework. We provide two examples.

In the method of convex combination of semantic embeddings (\method{ConSE}) \cite{NorouziMBSSFCD14}, their original class semantic embeddings are replaced with the corresponding predicted exemplars, while the combining coefficients remain the same.  
In \method{SynC} described in the previous section, the predicted exemplars are used to define the similarity values between the unseen classes and the bases, which in turn are used to compute the combination weights for constructing classifiers. In particular, their similarity measure is of the form in Eq.~(\ref{eSynCSimForm}). In this case, we simply need to change such a similarity measure to
\begin{align}
s_{cr} = \frac{\exp\{-\textrm{dis}(\vpsi(\va_c),\vpsi(\vb_r))\}}{\sum_{r=1}^{\cR}\exp\{-\textrm{dis}(\vpsi(\va_c),\vpsi(\vb_r))\}}.
\end{align}

In the experiments, we empirically show that existing semantic representations for ZSL are far from the optimal. Our approach can thus be considered as a way to improve semantic representations for ZSL.


\section{Experimental Setup}
\label{sExpSetup}

In this section, we describe experimental setup and protocols for evaluating zero-shot learning methods, including details on datasets and their splits, semantic representations, visual features, and metrics. We make distinctions between different settings to ensure fair comparison.

\subsection{Datasets and Splits}
\label{sExpSetupDataSplits}

We use five benchmark datasets in our experiments. Table~\ref{tDatasets} summarizes their key characteristics and splits. More details are provided below. 
\begin{itemize}
	\item The \textbf{Animals with Attributes (AwA)} dataset \cite{LampertNH14} consists of 30,475 images of 50 animal classes. 
	\item The \textbf{Animals with Attributes 2 (AwA2)} dataset \cite{XianLSA17} consists of 37,322 images of 50 animal classes. This dataset has been recently introduced as a replacement to \awa, whose images may not be licensed for free use and redistribution.  
	\item The \textbf{CUB-200-2011 Birds (CUB)} dataset~\cite{CUB} consists of 11,788 images of 200 fine-grained bird classes.
  \item The \textbf{SUN Attribute (SUN)} dataset~\cite{PattersonH14} consists of 14,340 images of 717 scene categories (20 images from each category). The dataset is drawn from the the \textbf{SUN} database~\cite{SUN}. 
	\item The \textbf{ImageNet} dataset~\cite{Imagenet} consists of two disjoint subsets. (i) The ILSVRC 2012 1K dataset~\cite{ILSVRC15} contains 1,281,167 training and 50,000 validation images from 1,000 categories and is treated as the seen-class data. (ii) Images of unseen classes come from the rest of the ImageNet Fall 2011 release dataset~\cite{Imagenet} that do not overlap with any of the 1,000 categories. We will call this release the ImageNet 2011 21K dataset (as in \cite{FromeCSBDRM13,NorouziMBSSFCD14}). Overall, this dataset contains 14,197,122 images from 21,841 classes, and we conduct our experiment on \textbf{20,842 unseen classes}\footnote{There is one class in the ILSVRC 2012 1K dataset that does not appear in the ImageNet 2011 21K dataset. Thus, we have a total of 20,842 unseen classes to evaluate.}.
\end{itemize}

\begin{table}[t]
\centering
\caption{Key characteristics of datasets and their class splits. SS (SS0) indicates the standard splits adopted in almost all previous ZSL methods. NS means the new splits proposed by~\cite{XianLSA17}.} \label{tDatasets}
\scriptsize
\begin{tabular}{c|c|c|c|c|c|}
Dataset         & Number of & \mtcl{2}{Class splits}                  & \mtcl{2}{\# of classes}     \\ \cline{3-6}
name            & images& Name                        & \#    & $\cS$         & $\cU$              \\ \hline
\multirow{2}{*}{\awa} & \multirow{2}{*}{30,475} & SS \cite{LampertNH14}       & 1         & 40            & 10                   \\ 
                & & NS \cite{XianLSA17}         & 1         & 40            & 10                   \\ \hline
\multirow{2}{*}{\awatwo} & \multirow{2}{*}{37,322} & SS \cite{LampertNH14}       & 1         & 40            & 10                    \\ 
                & & NS \cite{XianLSA17}         & 1         & 40            & 10                    \\ \hline					
\multirow{3}{*}{\cub} & \multirow{3}{*}{11,788} & SS \cite{ChangpinyoCGS16}   & 4         & 150           & 50                   \\
                & & SS0 \cite{AkataPHS13}       & 1         & 150           & 50                    \\ 
                & & NS \cite{XianLSA17}         & 1         & 150           & 50                    \\ \hline
\multirow{4}{*}{\sun} & \multirow{3}{*}{14,340} & SS$^\dagger$ \cite{ChangpinyoCGS16} & 10 & 645/646      & 72/71               \\
								&& SS0 \cite{XianLSA17}        & 1         & 645           & 72                    \\
                && NS \cite{XianLSA17}         & 1         & 645           & 72                    \\ \hline 
\imn            & 14,197,122& SS \cite{FromeCSBDRM13}     & 1         & 1,000         & 20,842    \\ \hline
\end{tabular}
\begin{flushleft}
$^\dagger$: Publicly available splits that follow \cite{LampertNH14} to do 10 splits.
\end{flushleft}
\end{table}

For each dataset, we select popular class splits in existing literature and make distinctions between them. On \cub and \sun, Changpinyo et al. \cite{ChangpinyoCGS16} randomly split each dataset into 4 and 10 disjoint subsets, respectively. In this case, we report the average score over those subsets; when computing a score on one subset, we use the rest as training classes.
Moreover, we differentiate between \emph{standard} and \emph{new} splits. Test classes in standard splits (SS or SS0) may overlap with classes used to pre-train deep neural networks for feature extraction (cf. Sect.~5.2 in \cite{XianLSA17} for details), but almost all previous ZSL methods have adopted them for evaluation. On the other hand, \emph{new} splits (NS), recently proposed by \cite{XianLSA17}, avoid such problematic class overlapping.
We summarize different class splits in Table~\ref{tDatasets}. We use SS0 on \cub and \sun to denote splits proposed by \cite{AkataPHS13} and \cite{XianLSA17}, respectively. On \imn, only SS exists as we do not have the problem of unseen classes ``leaking" during pre-training. The seen classes are selected from ImageNet ILSVRC 2012 1K~\cite{ILSVRC15} and are normally used for the pre-training of feature extractors.

For the generalized zero-shot learning (GZSL) setting (cf. Sect.~\ref{sExpSetupGZSL}) on \awa, \awatwo, \cub, and \sun, the test set must be the union of the seen classes' instances and the unseen classes' instances. The NS splits remain the same as before as they already reserve a portion of seen classes' instances for testing. For the SS or SS0 splits, we modify their original train and test sets following~\cite{ChaoCGS16}; we train the models using the 80\% of the seen classes' instances and test on the remaining 20\% (and the original unseen classes' instances).

\subsection{Semantic Representations}
\label{sExpSetupSem}

In our main experiments, we focus on attributes as semantic representations on \awa, \awatwo, \cub, and \sun, and word vectors as semantic representations on \imn. We use 85-, 312- and 102-dimensional continuous-valued attributes for the classes in \awa (and \awatwo), \cub, and \sun, respectively. For each class in \sun, we average attribute vectors over all images belonging to that class to obtain a class-level attribute vector. For \imn, we train a skip-gram model~\cite{MikolovCCD13,MikolovSCCD13} on the Wikipedia dump corpus\footnote{\url{http://dumps.wikimedia.org/enwiki/latest/enwiki-latest-pages-articles.xml.bz2} on September 1, 2015} consisting of more than 3 billion words to extract a 500-dimensional word vector for each class. Following \cite{FromeCSBDRM13,ChangpinyoCGS16}, we train the model for a single epoch. We ignore classes without word vectors in the experiments, resulting in \textbf{20,345 (out of 20,842) unseen classes}. Other details are in Appendix~\ref{sSuppSkip}. For both the continuous attribute vectors and the word vector embeddings of the class names, we normalize them to have unit $\ell_2$ norms unless stated otherwise.
Additional experimental setup and results on the effect of semantic representations can be found in Sect.~\ref{sExpResAddSem}.

\subsection{Visual Features}

We employ the strongest and most popular deep visual features in the literature: GoogLeNet~\cite{Inception} and ResNet~\cite{ResNet}. 
On all datasets but \awatwo, GoogLeNet features are 1,024-dimensional activations of the pooling units of the Inception v1 pre-trained on the ILSVRC 2012 1K dataset (\awa, \cub, \imn)~\cite{ILSVRC15} or the Places database (\sun) \cite{Places,Places18}, extracted using the Caffe package~\cite{Caffe}. We perform pre-processing on \cub by cropping all images with the provided bounding boxes following~\cite{FuHXFG14} and on \imn by center-cropping all images (without data augmentation or other preprocessing). We obtained the ResNet features on all datasets from \cite{XianSA17,XianLSA17}. These features are 2,048-dimensional activations of the pooling units of the ResNet-101 pretrained on the ILSVRC 2012 1K dataset~\cite{ILSVRC15}. Throughout the experiments, we denote GoogLeNet v1 features with G and ResNet features with R.

\subsection{Evaluation Protocols}

Denote by $A_{\mathcal{O} \rightarrow \mathcal{Y}}$ the accuracy of classifying test data whose labels come from $\mathcal{O}$ into the label space $\mathcal{Y}$. Note that the accuracy denotes the ``per-class" multi-way classification accuracy (defined below). 

\subsubsection{Conventional zero-shot learning}
\label{sExpSetupZSL}

The performance of ZSL methods on infrequent unseen classes whose examples are scarce (i.e., the tail) may not be reflected if we use \emph{per-sample} multi-way classification accuracy (averaged over all test images):
\begin{align}
A^{ps}_{\mathcal{U} \rightarrow \mathcal{U}} = \frac{\sum_{c \in \mathcal{U}} \text{\# correct predictions in c}}{\sum_{c \in \mathcal{U}} \text{\# test images in c}}.
\end{align}

For this reason, as in most previous work, on all datasets (with some exceptions on \imn below), we use the \emph{per-class} multi-way classification accuracy (averaged over all classes, and averaged over all test images in each class): 
\begin{align}
A_{\mathcal{U} \rightarrow \mathcal{U}} := A^{pc}_{\mathcal{U} \rightarrow \mathcal{U}} = \frac{1}{\left|\mathcal{U}\right|} \sum_{c \in \mathcal{U}} \frac{\text{\# correct predictions in c}}{\text{\# test images in c}}.
\end{align}
Note that we use $A_{\mathcal{U} \rightarrow \mathcal{U}}$ to denote $A^{pc}_{\mathcal{U} \rightarrow \mathcal{U}}$ in this paper.

Evaluating zero-shot learning on the large-scale \imn allows for different scenarios from evaluating on the other four datasets. We consider multiple subsets of the test set of \imn based on different characteristics.
Following the procedure in \cite{FromeCSBDRM13,NorouziMBSSFCD14}, we evaluate on the following subsets of increasing difficulty: 2-hop and 3-hop. These, respectively, correspond to 1,509 and 7,678 unseen classes that are within two and three tree hops of the 1K seen classes according to the ImageNet label hierarchy\footnote{\url{http://www.image-net.org/api/xml/structure_released.xml}}. Furthermore, following the procedure in \cite{XianLSA17}, we evaluate on the 500, 1K, and 5K most populated and least populated unseen classes. Finally, we evaluate on All: all 20,345 unseen classes in the ImageNet 2011 21K dataset that are not in the ILSVRC 2012 1K dataset. Note that the numbers of unseen classes are slightly different from what are used in~\cite{FromeCSBDRM13,NorouziMBSSFCD14} due to the missing semantic representations (i.e., word vectors) for certain class names.

To aid comparison with previous work on AlexNet and GoogLeNet features \cite{FromeCSBDRM13,NorouziMBSSFCD14,ChangpinyoCGS16,ChangpinyoCS17}, we also adopt two additional evaluation metrics: Flat hit@K (F@K) and Hierarchical precision@K (HP@K). F@K is defined as the percentage of test images for which the model returns the true label in its top K predictions. Note that F@1 is the \emph{per-sample} multi-way classification accuracy, which we report in the main text. We refer the reader to Appendix~\ref{sSuppExpIMN} for the details on HP@K and the rest of the results.

\subsubsection{Generalized zero-shot learning (GZSL)}
\label{sExpSetupGZSL}

In the generalized zero-shot learning (GZSL) setting, test data come from both seen and unseen classes. The label space is thus $\mathcal{T} = \mathcal{S} \cup \mathcal{U}$. This setting is of practical importance as real-world data should not be unrealistically assumed (as in conventional ZSL) to come from the unseen classes only. Since no labeled training data of the unseen classes are available during training, the bias of the classifiers toward the seen classes are difficult to avoid, making GZSL extremely challenging \cite{ChaoCGS16}.

Following \cite{ChaoCGS16}, we use the Area Under Seen-Unseen accuracy Curve (AUSUC) to evaluate ZSL methods in the GZSL setting. Below we describe briefly how to compute AUSUC, given a ZSL method. We assume that the ZSL method has a scoring function $f_c$ for each class $c \in \mathcal{T}$\footnote{In \method{SynC}, $f_c(\vx) = \vw_c\T\vx= (\sum_{r=1}^\cR s_{cr}\vv_r)\T\vx$ (cf. Sect.~\ref{sSynCSynthesis} and Eq.~(\ref{eSynCFinalCla})). In \method{EXEM}, $f_c(\vx) = \textrm{dis}_{NN}(\mM\vx, \vpsi(\va_c))$ if we treat $\vpsi(\va_c)$ as data and apply a nearest neighbor classifier (cf. Sect.~\ref{sEXEMZsl} and Eq.~(\ref{eExemPredict})).}. The approach of calibrated stacking \cite{ChaoCGS16} adapts the ZSL method so the prediction in the GZSL setting is  
\begin{align}
\hat{y} = \arg\max_{c\;\in\;\mathcal{T}}\quad f_c(\vx) - \gamma\I[c\in\mathcal{S}],
\label{eGZSLPredictCalib}
\end{align}
where $\gamma$ is the calibration factor. Adjusting $\gamma$ can balance two conflicting forces: recognizing data from seen classes versus those from unseen ones.

Recall that $\mathcal{T} = \mathcal{S} \cup \mathcal{U}$ is the union of the seen set $\mathcal{S}$ and the unseen set $\mathcal{U}$ of classes, where $\mathcal{S} = \{1,\cdots,\cS\}$ and $\mathcal{U} = \{\cS+1,\cdots,\cS+\cU\}$. Varying $\gamma$, we can compute a series of classification accuracies ($A_{\mathcal{U} \rightarrow \mathcal{T}}$, $A_{\mathcal{S} \rightarrow \mathcal{T}}$). We then can create the \emph{Seen-Unseen accuracy Curve (SUC)} with two ends for the extreme cases ($\gamma \rightarrow -\infty$ and $\gamma \rightarrow +\infty$). The Area Under SUC (AUSUC) summarizes this curve, similar to many curves whose axes representing conflicting goals, such as the Precision-Recall (PR) curve and the Receiving Operator Characteristic (ROC) curve. 

Recently, \cite{XianLSA17} alternatively proposed the harmonic mean of \emph{seen} and \emph{unseen} accuracies defined as
\begin{align}
H = \frac{ 2 * A_{\mathcal{S} \rightarrow \mathcal{T}} * A_{\mathcal{U} \rightarrow \mathcal{T}}  }{ A_{\mathcal{S} \rightarrow \mathcal{T}} + A_{\mathcal{U} \rightarrow \mathcal{T}}}.
\end{align}

While easier to implement and faster to compute than AUSUC, the harmonic mean may not be an accurate measure for the GZSL setting. It captures the performance of a zero-shot learning algorithm given a fixed degree of bias toward seen (or unseen) classes. This bias can vary across zero-shot learning algorithms and limit us to fairly compare them. We expand this point through our experiments in Sect. \ref{sExpResGZSL}.

\subsection{Baselines}
\label{sExpBaselines}

We consider 12 zero-shot learning baseline methods in \cite{XianLSA17} (cf. Table 3), including \zsldap, \zsliap, \zslcmt, \zsldevise, \zslconse, \zslale, \zslsje, \zsllatem, \zsleszsl, \zslsse, \zslsae, \zslgfzsl. Additionally, we consider \zslcosta, \zslhat, \zslbidilel, and \zslcca in some of our experiments. These baselines are diverse in their approaches to zero-shot learning. Note that \zsldap and \zsliap require binary semantic representations, and we follow the setup in~\cite{ChangpinyoCGS16} to obtain them. For further discussion of these methods, see Sect.~\ref{sRelated} as well as \cite{XianLSA17,FuXJXSG18}. 

\subsection{Summary of Variants of Our Methods}
\label{sExpOurVariants}

We consider the following variants of \method{SynC} that are different in the type of loss used in the objective function (cf. Sect.~\ref{sSynCLearnPhantom}).
\begin{itemize}[noitemsep]
\item \method{SynC$^\textrm{o-vs-o}$}: one-versus-other with the squared hinge loss.
\item \method{SynC$^\textrm{cs}$}: Crammer-Singer multi-class SVM loss \cite{CrammerS02} with $\Delta(c,y_n)=1$ if $c\neq y_n$ and $0$ otherwise.
\item \method{SynC$^\textrm{struct}$}: Crammer-Singer multi-class SVM loss \cite{CrammerS02} with $\Delta(c,y_n)=\left\|{\va_c-\va_{y_n}}\right\|_2$.
\end{itemize}
\emph{Unless stated otherwise, we adopt the version of \method{SynC} that sets the number of base classifiers $\cR$ to be the number of seen classes $\cS$, and sets $\vb_r = \va_c$ for $r=c$ (i.e., without learning semantic representations). The results with learned representations are in Appendix~\ref{sSuppExpResAddSynC}.} 

Furthermore, we consider the following variants of \method{EXEM} (cf. Sect~\ref{sEXEMZsl}).
\begin{itemize}[noitemsep]
\item \method{EXEM (\emph{ZSL method})}: A ZSL method with predicted exemplars as semantic representations, where \emph{ZSL method} $=$ \method{ConSE} \cite{NorouziMBSSFCD14}, \method{ESZSL} \cite{Bernardino15}, or the variants of \method{SynC}.
\item \method{EXEM (1NN)}: 1-nearest neighbor classifier with the Euclidean distance to the exemplars.
\item \method{EXEM (1NNs)}: 1-nearest neighbor classifier with the \emph{standardized} Euclidean distance to the exemplars, where the standard deviation is obtained by averaging the intra-class standard deviations of all seen classes.
\end{itemize}

\method{EXEM (\emph{ZSL method})} regards the predicted exemplars as the improved semantic representations.
On the other hand, \method{EXEM (1NN)} treats predicted exemplars as data prototypes.
The standardized Euclidean distance in \method{EXEM (1NNs)} is introduced as a way to scale the variance of different dimensions of visual features. In other words, it helps reduce the effect of \emph{collapsing} data that is caused by our usage of the average of each class' data as cluster centers.


\section{Experimental Results}
\label{sExpRes}
 
The outline of our experimental results in this section is as follows. We first provide a summary of our main results (Sect.~\ref{sMainExp}, Table~\ref{tbMain}), followed by detailed results in various experimental scenarios. We provide detailed conventional ZSL results on 4 small datasets \awa, \awatwo, \cub, \sun (Sect.~\ref{sExpResZSL}, Table~\ref{tbZSLMainSmall}) and on the large-scale \imn (Sect.~\ref{sExpResZSL}, Table~\ref{tbZSLMainIMN}). We separate GZSL results on small datasets (Sect.~\ref{sExpResGZSL}) into two parts: one using AUSUC (Table~\ref{tbGZSLMain}) and the other comparing multiple metrics (Table~\ref{tbGZSLHarmonic}). The rest are additional results on \imn (Sect.~\ref{sExpResAdd}), including an empirical comparison between semantic representations (Table~\ref{tbZSLSemanticPCIMN}), a comparison to recent state-of-the-art with \emph{per-sample} accuracy (Table~\ref{tbZSLSemanticPSIMN}), and results with ideal semantic representations (Table~\ref{tbZSLSemanticIdealIMN}). Results on \imn using an earlier experimental setup \cite{FromeCSBDRM13,NorouziMBSSFCD14,ChangpinyoCGS16} can be found in Appendix~\ref{sSuppExpIMN}. Finally, further analyses on \method{SynC} and \method{EXEM} are in Appendix~\ref{sSuppExpResAddSynC} and Appendix~\ref{sSuppExpResAddExem}, respectively.  

\subsection{Main Experimental Results}
\label{sMainExp}

\begin{table*}
\centering
\tabcolsep 3pt
\caption{\small Main results: our results and the previously \emph{published} ones on the conventional ZSL task based on per-class multi-way classification accuracy (in \%) and on the generalized ZSL task based on AUSUC. The ResNet features are used with the new splits (NS) on small datasets and standard split (SS) on \imn (All: 20,345 unseen classes). For each dataset, the best is in \bst{red}; the second best in \sbst{blue}; the third best in \tbst{green}. We also summarize the results on small datasets by ordering zero-shot methods based on their mean ranks (in brackets) in ZSL (bottom left) and GZSL (bottom right) settings. Element (i, j) indicates the number of times method i ranks at j-th.} \label{tbMain}
\vskip -0.5em
\begin{tabular}{c|c|c|c|c|c|c|c|c|c|c|c|}
& \mtcl{6}{ZSL (Per-class Accuracy $A_{\mathcal{U} \rightarrow \mathcal{U}}$)} & \mtcl{5}{Generalized ZSL (AUSUC)} \\ \hline
Approach/Datasets & {Reported by}& \awa & \awatwo & \cub & \sun & \imn & {Reported by}& \awa & \awatwo & \cub	& \sun \\ \hline
\zsldap & \gbu & 44.1 & 46.1 & 40.0 & 39.9 & - & \tw & 0.341 & 0.353 & 0.200 & 0.094 \\ 
\zsliap & \gbu & 35.9 & 35.9 & 24.0 & 19.4 & - & \tw & 0.376 & 0.392 & 0.209 & 0.121 \\ 
\zslcmt & \gbu & 39.5 & 37.9 & 34.6 & 39.9 & 0.29 & - & - & - & - & - \\ 
\zsldevise & \gbu & 54.2 & 59.7 & 52.0 & 56.5 & 0.49 & - & - & - & - & - \\ 
\zslconse & \gbu & 45.6 & 44.5 & 34.3 & 38.8 & 0.95 & \tw & 0.350 & 0.344 & 0.214 & 0.170 \\ 
\zslale & \gbu & 59.9 & 62.5 & 54.9 & 58.1 & 0.50 & \tw & 0.504 & 0.538 & 0.338 & 0.193 \\ 
\zslsje & \gbu & 65.6 & 61.9 & 53.9 & 53.7 & 0.52 & - & - & - & - & - \\ 
\zsllatem & \gbu & 55.1 & 55.8 & 49.3 & 55.3 & 0.50 & \tw & 0.506 & 0.514 & 0.276 & 0.171 \\ 
\zslsse & \gbu & 60.1 & 61.0 & 43.9 & 51.5 & - & - & - & - & - & - \\ 
\zsleszsl & \gbu & 58.2 & 58.6 & 53.9 & 54.5 & 0.62 & \tw & 0.452 & 0.454 & 0.303 & 0.138 \\ 
\zslsae & \gbu & 53.0 & 54.1 & 33.3 & 40.3 & 0.56 & - & - & - & - & - \\ 
\zslgfzsl & \gbu & \sbst{68.3} & 63.8 & 49.3 & \tbst{60.6} & - & - & - & - & - & - \\ 
\zslcosta & \tw & 49.0 & 53.2 & 44.6 & 43.0 & - & - & - & - & - & - \\ \hline
\zslsyncovo & \tw & 57.0 & 52.6 & 54.6 & 55.7 & 0.98 & \tw & 0.454 & 0.438 & 0.353 & 0.220 \\ 
\zslsynccs & \tw & 58.4 & 53.7 & 51.5 & 47.4 & - & \tw & 0.477 & 0.463 & 0.359 & 0.189 \\ 
\zslsyncstr & \tw & 60.4 & 59.7 & 53.4 & 55.9 & 0.99 & \tw & 0.505 & 0.504 & 0.337 & 0.241 \\ \hline
\zslexemconse & \tw & 57.6 & 57.9 & 44.5 & 51.5 & - & \tw & 0.439 & 0.425 & 0.266 & 0.189 \\ 
\zslexemeszsl & \tw & 65.2 & 63.6 & \tbst{56.9} & 57.1 & - & \tw & 0.522 & 0.538 & 0.346 & 0.191 \\ 
\zslexemsyncovo & \tw & 60.0 & 56.1 & \tbst{56.9} & 57.4 & 1.25 & \tw & 0.481 & 0.474 & \tbst{0.361} & 0.221 \\ 
\zslexemsynccs & \tw & 60.5 & 57.9 & 54.2 & 51.1 & - & \tw & 0.497 & 0.481 & 0.360 & 0.205 \\ 
\zslexemsyncstr & \tw & 65.5 & \sbst{64.8} & \bst{60.5} & 60.1 & \bst{1.29} & \tw & \tbst{0.533} & \tbst{0.552} & \bst{0.397} & \sbst{0.251} \\ 
\zslexemnn & \tw & \bst{68.5} & \bst{66.7} & 54.2 & \bst{63.0} & \tbst{1.26} & \tw & \sbst{0.565} & \bst{0.565} & 0.298 & \bst{0.253} \\ 
\zslexemnns & \tw & \tbst{68.1} & \tbst{64.6} & \sbst{58.0} & \sbst{62.9} & \bst{1.29} & \tw & \bst{0.575} & \sbst{0.559} & \sbst{0.366} & \sbst{0.251} \\ \hline
\end{tabular}
\begin{tabular}{cc}
\includegraphics[width=0.45\textwidth]{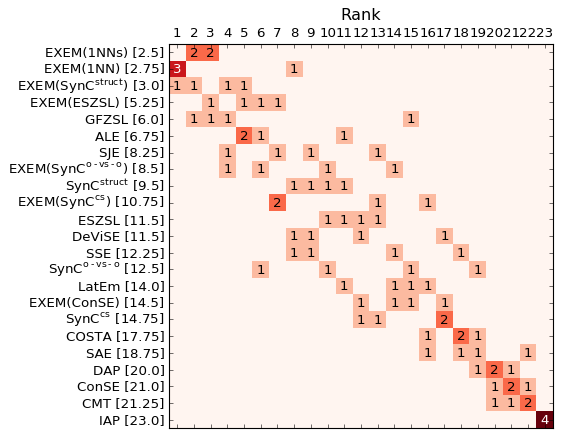} & 
\includegraphics[width=0.45\textwidth]{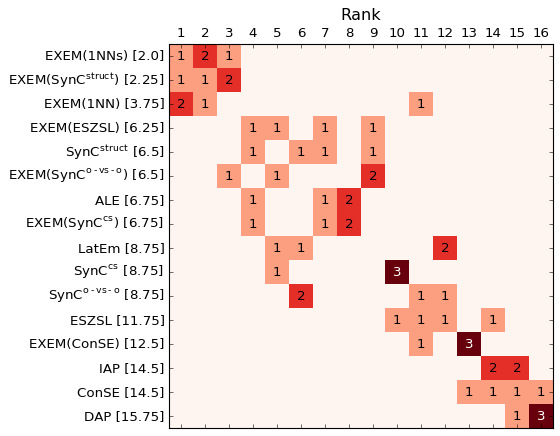} \\
ZSL \label{fRankZSLSmall} & GZSL \label{fRankGZSLSmall}
\end{tabular}
\end{table*}

Table~\ref{tbMain} summarizes our results on both conventional and generalized zero-shot learning using the ResNet features. On \awa, \awatwo, \cub, \sun, we use visual attributes and the new splits (where the unseen/test classes does not overlap with those used for feature extraction; see Sect.~\ref{sExpSetupDataSplits}). On \imn, we use word vectors of the class names and the standard split. We use per-class multi-way classification accuracy for the conventional zero-shot learning task and AUSUC for the generalized zero-shot learning task.

On the word-vector-based \imn (All: 20,345 unseen classes), our \method{SynC} and \method{EXEM} outperform baselines by a significant margin.
To encapsulate the general performances of various ZSL methods on small visual-attribute-based datasets, we adopt the non-parametric Friedman test \cite{GarciaH08} as in \cite{XianLSA17}; we compute the mean rank of each method across small datasets and use it to order 23 methods in the conventional task and 16 methods in the generalized task. We find that on small datasets \method{EXEM (1NN)}, \method{EXEM (1NNs)} and \method{EXEM (SynC$^\textrm{struct}$)} are the top three methods in each setting. Additionally, \method{SynC$^\textrm{struct}$} performs competitively with the rest of the baselines. Notable strong baselines are \method{GFZSL}, \method{ALE}, and \method{SJE}. 

The rankings in both settings demonstrate that a positive correlation appears to exist between the ZSL and GZSL performances, but this is not always the case. For example, \method{ALE} outperforms \method{SynC$^\textrm{struct}$} on \sun in ZSL (58.1 vs. 55.9) but it underperforms in GZSL (0.193 vs. 0.241). The same is true in many other cases such as \method{DAP} vs. \method{IAP} on all datasets and \method{EXEM (SynC$^\textrm{cs}$)} vs. \method{EXEM (ESZSL)} on \cub and \sun. This observation stresses the importance of GZSL as an evaluation setting.   

\subsection{Conventional Zero-Shot Learning Results}
\label{sExpResZSL}

\begin{table*}
\centering
\caption{Comparison between existing ZSL approaches in per-class multi-way classification accuracy (in \%) on small datasets. All methods use visual attributes as semantic representations. Each row corresponds to a ZSL method. Each column corresponds to a \emph{scenario} with a particular combination of dataset, its class split, and visual features. We use GoogLeNet features (G) and ResNet features (R). Class splits include both standard (SS or SS0) and new (NS) splits. For each scenario, the best is in \bst{red} and the second best in \sbst{blue}.} \label{tbZSLMainSmall}
\vskip -0.5em
\begin{tabular}{c|c|c|c|c|c|c|c|c|c|c|c|c|c|}
& \mtcl{2}{Reported by} & \mtcl{3}{\awa} & \mtcl{2}{\awatwo} & \mtcl{3}{\cub} & \mtcl{3}{\sun} \\ \hline
Features  & G & R & G & \mtcl{2}{R} & \mtcl{2}{R} & G & \mtcl{2}{R} &\mtcl{1}{G} & \mtcl{2}{R} \\ \hline
Approach/Splits	& \mtcl{2}{-} & SS & SS & NS & SS & NS  & SS  & SS0 & NS & SS & SS0 & NS \\ \hline
\zsldap & \sync & \gbu & 60.5 & 57.1 & 44.1 & 58.7 & 46.1 & 39.1 & 37.5 & 40.0 & 44.5 & 38.9 & 39.9 \\ 
\zsliap & \sync & \gbu & 57.2 & 48.1 & 35.9 & 46.9 & 35.9 & 36.7 & 27.1 & 24.0 & 40.8 & 17.4 & 19.4 \\ 
\zslhat & \mtmdl & - & 74.9 & - & - & - & - & - & - & - & - & - & - \\ 
\zslcmt & - & \gbu & - & 58.9 & 39.5 & 66.3 & 37.9 & - & 37.3 & 34.6 & - & 41.9 & 39.9 \\ 
\zsldevise & - & \gbu & - & 72.9 & 54.2 & 68.6 & 59.7 & - & 53.2 & 52.0 & - & 57.5 & 56.5 \\ 
\zslconse & \sync & \gbu & 63.3 & 63.6 & 45.6 & 67.9 & 44.5 & 36.2 & 36.7 & 34.3 & 51.9 & 44.2 & 38.8 \\ 
\zslale & \tw & \gbu & 74.8 & 78.6 & 59.9 & 80.3 & 62.5 & 53.8 & 53.2 & 54.9 & 66.7 & 59.1 & 58.1 \\ 
\zslsje & \sync & \gbu & 66.3 & 76.7 & 65.6 & 69.5 & 61.9 & 46.5 & 55.3 & 53.9 & 56.1 & 57.1 & 53.7 \\ 
\zsllatem & \exem & \gbu & 72.1 & 74.8 & 55.1 & 68.7 & 55.8 & 48.0 & 49.4 & 49.3 & 64.5 & 56.9 & 55.3 \\ 
\zslsse & - & \gbu & - & 68.8 & 60.1 & 67.5 & 61.0 & - & 43.7 & 43.9 & - & 54.5 & 51.5 \\ 
\zsleszsl & \tw & \gbu & 73.2 & 74.7 & 58.2 & 75.6 & 58.6 & 54.7 & 55.1 & 53.9 & 58.7 & 57.3 & 54.5 \\ 
\zslsae & - & \gbu & - & 80.6 & 53.0 & \sbst{80.7} & 54.1 & - & 33.4 & 33.3 & - & 42.4 & 40.3 \\ 
\zslgfzsl & - & \gbu & - & 80.5 & \sbst{68.3} & 79.3 & 63.8 & - & 53.0 & 49.3 & - & \sbst{62.9} & 60.6 \\ 
\zslbidilel & \bidilel & - & 72.4 & - & - & - & - & - & - & - & - & - & - \\ 
\zslcosta & \sync & \tw & 61.8 & 70.1 & 49.0 & 63.0 & 53.2 & 40.8 & 42.1 & 44.6 & 47.9 & 46.7 & 43.0 \\ \hline
\zslsyncovo & \sync & \tw & 69.7 & 75.2 & 57.0 & 71.0 & 52.6 & 53.4 & 53.5 & 54.6 & 62.8 & 59.4 & 55.7 \\ 
\zslsynccs & \sync & \tw & 72.1 & 77.9 & 58.4 & 66.7 & 53.7 & 51.6 & 49.6 & 51.5 & 53.3 & 54.7 & 47.4 \\ 
\zslsyncstr & \sync & \tw & 72.9 & 78.4 & 60.4 & 75.4 & 59.7 & 54.5 & 53.5 & 53.4 & 62.7 & 59.1 & 55.9 \\ \hline
\zslexemconse & \exem & \tw & 70.5 & 74.6 & 57.6 & 76.6 & 57.9 & 46.2 & 47.4 & 44.5 & 60.0 & 55.6 & 51.5 \\ 
\zslexemeszsl & \tw & \tw & \bst{78.1} & \sbst{80.9} & 65.2 & 80.4 & 63.6 & 57.5 & 59.3 & 56.9 & 63.4 & 58.2 & 57.1 \\ 
\zslexemsyncovo & \exem & \tw & 73.8 & 77.7 & 60.0 & 77.1 & 56.1 & 56.2 & 58.3 & 56.9 & 66.5 & 60.9 & 57.4 \\ 
\zslexemsynccs & \exem & \tw & 75.0 & 79.5 & 60.5 & 75.3 & 57.9 & 56.1 & 56.2 & 54.2 & 58.4 & 57.2 & 51.1 \\ 
\zslexemsyncstr & \exem & \tw & \sbst{77.2} & \bst{82.4} & 65.5 & 80.2 & \sbst{64.8} & \bst{59.8} & \bst{60.1} & \bst{60.5} & 66.1 & 62.2 & 60.1 \\ 
\zslexemnn & \exem & \tw & 76.2 & \sbst{80.9} & \bst{68.5} & 78.1 & \bst{66.7} & 56.3 & 57.1 & 54.2 & \bst{69.6} & \bst{64.2} & \bst{63.0} \\ 
\zslexemnns & \exem & \tw & 76.5 & 77.8 & 68.1 & \bst{81.4} & 64.6 & \sbst{58.5} & \sbst{59.7} & \sbst{58.0} & \sbst{67.3} & 62.7 & \sbst{62.9} \\ \hline
\end{tabular}
\end{table*}

In Table~\ref{tbZSLMainSmall}, we provide detailed results on small datasets (\awa, \awatwo, \cub, and \sun), including other popular scenarios for zero-shot learning that were investigated by past work. In particular, we include results for other visual features and data splits. All zero-shot learning methods use visual attributes as semantic representations. Similar to before, we find that the variants of \method{EXEM} consistently outperform other ZSL approaches. Other observations are discussed below. 

\paragraph{Variants of \method{SynC}:}
On \awa and \awatwo, \method{SynC$^\textrm{struct}$} outperforms \method{SynC$^\textrm{cs}$} and \method{SynC$^\textrm{o-vs-o}$} consistently, but it is inconclusive whether \method{SynC$^\textrm{cs}$} or \method{SynC$^\textrm{o-vs-o}$} is more effective. On \cub and \sun, \method{SynC$^\textrm{o-vs-o}$} and \method{SynC$^\textrm{struct}$} clearly outperform \method{SynC$^\textrm{cs}$}.

\paragraph{Variants of \method{EXEM}:}
We find that there is no clear winner between using predicted exemplars as improved semantic representations or as data prototypes. The former seems to perform better on datasets with fewer seen classes. Nonetheless, we note that using 1-nearest-neighbor classifiers clearly scales much better than using most zero-shot learning methods; \method{EXEM (1NN)} and \method{EXEM (1NNs)} are more efficient than \method{EXEM (SynC)}, \method{EXEM (ESZSL)}, \method{EXEM (ConSE)} in training. Finally, while we expect that using the standardized Euclidean distance (\method{EXEM (1NNs)}) instead of the Euclidean distance (\method{EXEM (1NN)}) for nearest neighbor classifiers would help improve the accuracy, this is the case only on \cub (and on \imn as we will show in Sect.~\ref{sExpResZSLLarge}). We hypothesize that accounting for the variance of visual exemplars' dimensions is important in fine-grained ZSL recognition.

\paragraph{\method{EXEM (\emph{ZSL method})} improves over ZSL method:}
Our approach of treating predicted visual exemplars as the improved semantic representations significantly outperforms taking semantic representations as given. \method{EXEM (SynC)}, \method{EXEM (ConSE)}, and \method{EXEM (ESZSL)} outperform their corresponding \emph{base} ZSL methods by relatively 2.1-13.3\%, 11.4-32.7\%, and 1.6-12.0\%, respectively. 
Thus, we conclude that the semantic representations (on the predicted exemplar space) are indeed improved by \method{EXEM}. We further qualitatively and quantitatively analyze the nature of the predicted exemplars in Appendix \ref{sSuppExpResAddExem}.

\paragraph{Visual features and class splits:}
The choice of visual features and class splits affect performance greatly, suggesting that these choices should be made explicit or controlled in zero-shot learning studies. 

On the same standard split of \awa, we observe that the ResNet features are generally stronger than the GoogLeNet features, but not always (\zsldap and \zsliap), suggesting that further investigation on the algorithm-specific transferability of different types of features may be needed.

As observed in \cite{XianLSA17}, zero-shot learning on the new splits is a more difficult task because pre-trained visual features have not seen the test classes. However, the effect of class splits on the fine-grained benchmark \cub is not apparent as in other datasets. This suggests that, when object classes are very different, class splits create very different zero-shot learning tasks where some are much harder than others. Evaluating the ``possibility" of transfer of these different tasks is important and likely can be more easily approached using coarse-grained benchmarks.

\begin{figure*}
	\centering
	\includegraphics[width=1.0\textwidth]{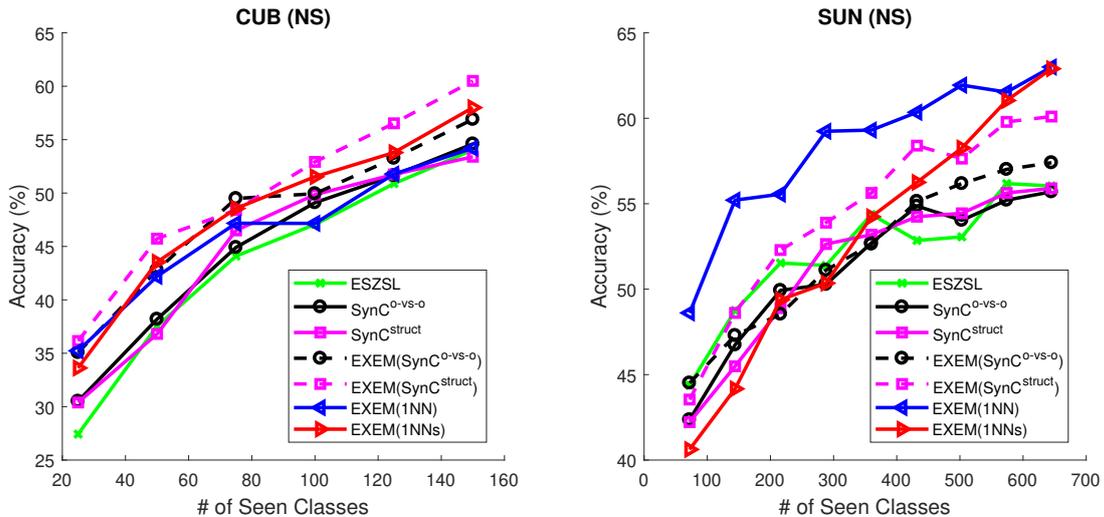}
	\caption{{The effect of varying the numbers of seen classes on per-class accuracy on the new splits (NS) of \cub and \sun.}}
	\label{fDiffSeen}
\end{figure*}

\paragraph{Different numbers of seen classes:}
{
We further examine the effect of varying the number of seen classes on the zero-shot classification accuracy. We focus on the new splits (NS) of \cub and \sun, as their default numbers of seen classes are reasonably large (cf. Table~\ref{tDatasets}). We construct subsets of the default seen classes with decreasing sizes: classes are selected uniformly at random and a larger subset is a superset of a smaller one. The unseen classes remain the same as before. We experiment with our proposed methods \zslsyncovo, \zslsyncstr, \zslexemsyncovo, \zslexemsyncstr, \zslexemnn, and \zslexemnns, as well as a strong baseline \zsleszsl. We report the results in Fig.~\ref{fDiffSeen}.

As expected, we see that the ZSL accuracies of all ZSL methods degrade as the number of seen classes is reduced. When comparing between methods, we see that our previous observation that \method{EXEM (\emph{ZSL method})} generally outperforms the corresponding \emph{ZSL method} (dashed vs. solid curves) still holds, despite the fact that the quality of predicted exemplars is expected to suffer from the reduced number of training semantic representation-exemplars pairs. Another interesting observation is that \zslexemnn is the most robust ZSL method with an absolute $14.4\%$ drop (relative $22.8\%$) on \sun when the number of seen classes decreases from $645$ to $72$ and an absolute $19.0\%$ (relative $35.0\%$) on \cub when the number decreases from $150$ to $25$. In comparison, the performances of other methods degrade faster than that of \zslexemnn; $22.3\%$ for \zslexemnns on \sun and $26.6\%$ for \zsleszsl on \cub. We think that this observation is likely caused by complex models' overfitting. For instance, \zslexemnns adds to the complexity of \zslexemnn, computing the standard deviation by averaging the intra-class standard deviations of seen classes. Based on this, we suggest that \zslexemnn be the first ``go-to" method in scenarios where the number of seen classes is extremely small.
}

\begin{table*}
	\centering
	\caption{Comparison between existing ZSL approaches in multi-way classification accuracy (in \%) on \imn. Normal text denotes per-class accuracy and \emph{italicized} text denotes per-sample accuracy (following previous work). All methods use word vectors of the class names as semantic representations. Each row corresponds to a ZSL method. Each column corresponds to a \emph{scenario} with a particular combination of a selected test set and visual features. We use GoogLeNet features (G) and ResNet features (R). For each scenario, the best is in \bst{red} and the second best in \sbst{blue}.} \label{tbZSLMainIMN}
	\vskip -0.5em
	\scriptsize
	\begin{tabular}{c|c|c|c|c|c|c|c|c|c|c|c|c|c|c|c|c|c|}
		& \multicolumn{2}{c|}{Reported by} & \multicolumn{4}{c|}{Hierarchy} & \multicolumn{3}{c|}{Most populated} & \multicolumn{3}{c|}{Least populated} & \multicolumn{2}{c|}{All} \\ \hline
		Splits & \multicolumn{2}{c|}{} & \multicolumn{2}{c|}{2-hop} & \multicolumn{2}{c|}{3-hop} & 500 & 1K & 5K & 500 & 1K & 5K & \multicolumn{2}{c|}{} \\ \hline
		Approach/Features & G & R & G & R & G & R & R & R & R & R & R & R & G & R \\ \hline
		\zslcmt & - & \gbu & \textit{-} & 2.88 & \textit{-} & 0.67 & 5.10 & 3.04 & 1.04 & 1.87 & 1.08 & 0.33 & \textit{-} & 0.29 \\ 
		\zsldevise & - & \gbu & \textit{-} & 5.25 & \textit{-} & 1.29 & 10.36 & 6.68 & 1.94 & 4.23 & 2.86 & 0.78 & \textit{-} & 0.49 \\ 
		\zslconse & \sync & \gbu & \textit{8.3} & 7.63 & \textit{2.6} & 2.18 & 12.33 & 8.31 & 3.22 & 3.53 & 2.69 & 1.05 & \textit{1.3} & 0.95 \\ 
		\zslale & - & \gbu & \textit{-} & 5.38 & \textit{-} & 1.32 & 10.40 & 6.77 & 2.00 & 4.27 & 2.85 & 0.79 & \textit{-} & 0.50 \\ 
		\zslsje & - & \gbu & \textit{-} & 5.31 & \textit{-} & 1.33 & 9.88 & 6.53 & 1.99 & 4.93 & 2.93 & 0.78 & \textit{-} & 0.52 \\ 
		\zsllatem & - & \gbu & \textit{-} & 5.45 & \textit{-} & 1.32 & 10.81 & 6.63 & 1.90 & 4.53 & 2.74 & 0.76 & \textit{-} & 0.50 \\ 
		\zsleszsl & - & \gbu & \textit{-} & 6.35 & \textit{-} & 1.51 & 11.91 & 7.69 & 2.34 & 4.50 & 3.23 & 0.94 & \textit{-} & 0.62 \\ 
		\zslsae & - & \gbu & \textit{-} & 4.89 & \textit{-} & 1.26 & 9.96 & 6.57 & 2.09 & 2.50 & 2.17 & 0.72 & \textit{-} & 0.56 \\ 
		\zslgfzsl & - & \gbu & \textit{-} & 1.45 & \textit{-} & - & 2.01 & 1.35 & - & 1.40 & 1.11 & 0.13 & \textit{-} & - \\ \hline
		\zslsyncovo & \sync & \tw & \textit{10.5} & 9.60 & \textit{2.9} & 2.31 & 16.38 & 11.14 & 3.50 & 5.47 & 3.83 & 1.34 & \textit{1.4} & 0.98 \\ 
		\zslsyncstr & \sync & \tw & \textit{9.8} & 8.76 & \textit{2.9} & 2.25 & 14.93 & 10.33 & 3.44 & 4.20 & 3.22 & 1.26 & \textit{1.5} & 0.99 \\ \hline
		\zslexemsyncovo & \exem & \tw & \textit{11.8} & 11.15 & \textit{\sbst{3.4}} & 2.95 & \bst{19.26} & \bst{13.37} & 4.50 & \sbst{6.33} & 4.48 & 1.62 & \textit{1.6} & 1.25 \\ 
		\zslexemsyncstr & \tw & \tw & \textit{\sbst{12.1}} & \sbst{11.28} & \textit{\sbst{3.4}} & \bst{3.02} & 18.97 & \sbst{13.26} & \bst{4.57} & 6.10 & \bst{4.74} & 1.67 & \textit{\sbst{1.7}} & \bst{1.29} \\ 
		\zslexemnn & \exem & \tw & \textit{11.7} & 10.60 & \textit{\sbst{3.4}} & 2.87 & 18.15 & 12.68 & 4.40 & 6.20 & 4.61 & \sbst{1.68} & \textit{\sbst{1.7}} & 1.26 \\ 
		\zslexemnns & \exem & \tw & \textit{\bst{12.5}} & \bst{11.58} & \textit{\bst{3.6}} & \sbst{2.99} & \sbst{19.09} & 13.16 & \sbst{4.55} & \bst{6.70} & \sbst{4.67} & \bst{1.70} & \textit{\bst{1.8}} & \bst{1.29} \\ \hline
	\end{tabular}
\end{table*}

\subsection{Large-Scale Conventional Zero-Shot Learning Results}
\label{sExpResZSLLarge}

In Table~\ref{tbZSLMainIMN}, we provide detailed results on the large-scale \imn, including scenarios for zero-shot learning that were investigated by \cite{XianLSA17} and \cite{FromeCSBDRM13,NorouziMBSSFCD14}. In particular, we include results for other visual features and other test subsets of \imn. All zero-shot learning methods use word vectors of the class names as semantic representations. To aid comparison with previous work, we use per-class accuracy when evaluating on ResNet features (R) and per-sample accuracy when evaluating on GoogLeNet features (G). We compare the two types of accuracy in Appendix~\ref{sSuppExpIMN} and find that the per-sample accuracy is a more optimistic metric than the per-class accuracy is, reasonably reflected by the fact that \imn's classes are highly unbalanced. Furthermore, in Sect.~\ref{sExpResAddSem}, we analyze the effect of different types of semantic representations on zero-shot performance and report the best published results on this dataset.

We compare and contrast these results with the one on small datasets (Table~\ref{tbZSLMainSmall}).
We observe that, while \method{SynC} does not clearly outperform other baselines on small datasets, it does so on \imn, in line with the observation in \cite{XianLSA17} (cf. Table 5 which only tested on \method{SynC$^\textrm{o-vs-o}$}). Any variants of \method{EXEM} further improves over \method{SynC} in all scenarios (in each column). As in small datasets, \method{EXEM (\emph{ZSL method})} improves over \emph{method}.

\paragraph{Variants of \method{SynC}:}
\method{SynC$^\textrm{o-vs-o}$} generally outperforms \method{SynC$^\textrm{struct}$} in all scenarios but the settings ``3-hop (G)" and ``All." This is reasonable as the semantic distances needed in \method{SynC$^\textrm{struct}$} may not be reliable as they are based on word vectors. This hypothesis is supported by the fact that \method{EXEM (SynC$^\textrm{struct}$)} manages to reduce the gap to or even outperforms \method{EXEM (SynC$^\textrm{o-vs-o}$)} \textbf{after} semantic representations have been improved by the task of predicting visual exemplars. 

\method{SynC$^\textrm{struct}$} becomes more effective against \method{SynC$^\textrm{o-vs-o}$} when the labeling space grows larger (i.e., when we move from 2-hop to 3-hop to All or when we move from 500 to 1K to 5K). In fact, it even achieves slightly better performance when we consider All classes (\textit{1.5} vs. \textit{1.4} and 0.99 vs. 0.98). One hypothesis is that, when the labeling space is large, the ZSL task becomes so difficult that both methods become equally bad. Another hypothesis is that the semantic distances in \method{SynC$^\textrm{struct}$} only helps when we consider a large number of classes.

\paragraph{Variants of \method{EXEM}:}
First, we find that, as in \cub, using the standardized Euclidean distance instead of the Euclidean distance for nearest neighbor classifiers helps improve the accuracy --- \method{EXEM (1NNs)} outperforms \method{EXEM (1NN)} in all cases. This suggests that there is a certain effect of collapsing actual data during training. Second, \method{EXEM (SynC$^\textrm{struct}$)} generally outperforms \method{EXEM (SynC$^\textrm{o-vs-o}$)}, except when classes are very frequent or very rare. Third, \method{EXEM (1NNs)} is better than \method{EXEM (SynC$^\textrm{o-vs-o}$)} on rare classes, but worse on frequent classes. Finally, \method{EXEM (1NNs)} and \method{EXEM (SynC$^\textrm{struct}$)} are in general the best approaches but one does not clearly outperform the other.

\paragraph{Visual features:}
Comparing the columns ``G" (GoogLeNet) of Table~\ref{tbZSLMainIMN} against the row ``wv-v1" of Table~\ref{tbZSLSemanticPSFullIMN} in Appendix~\ref{sSuppExpIMN} (ResNet), we show that, when evaluated with the same ``per-sample" metrics on 2-hop, 3-hop, and All test subsets of \imn, ResNet features are clearly stronger (i.e., more transferable) than GoogLeNet features.

\subsection{Generalized Zero-Shot Learning Results}
\label{sExpResGZSL}

\subsubsection{Comparison among ZSL approaches}

\begin{figure}[t]
	\centering
	\includegraphics[width=0.5\textwidth]{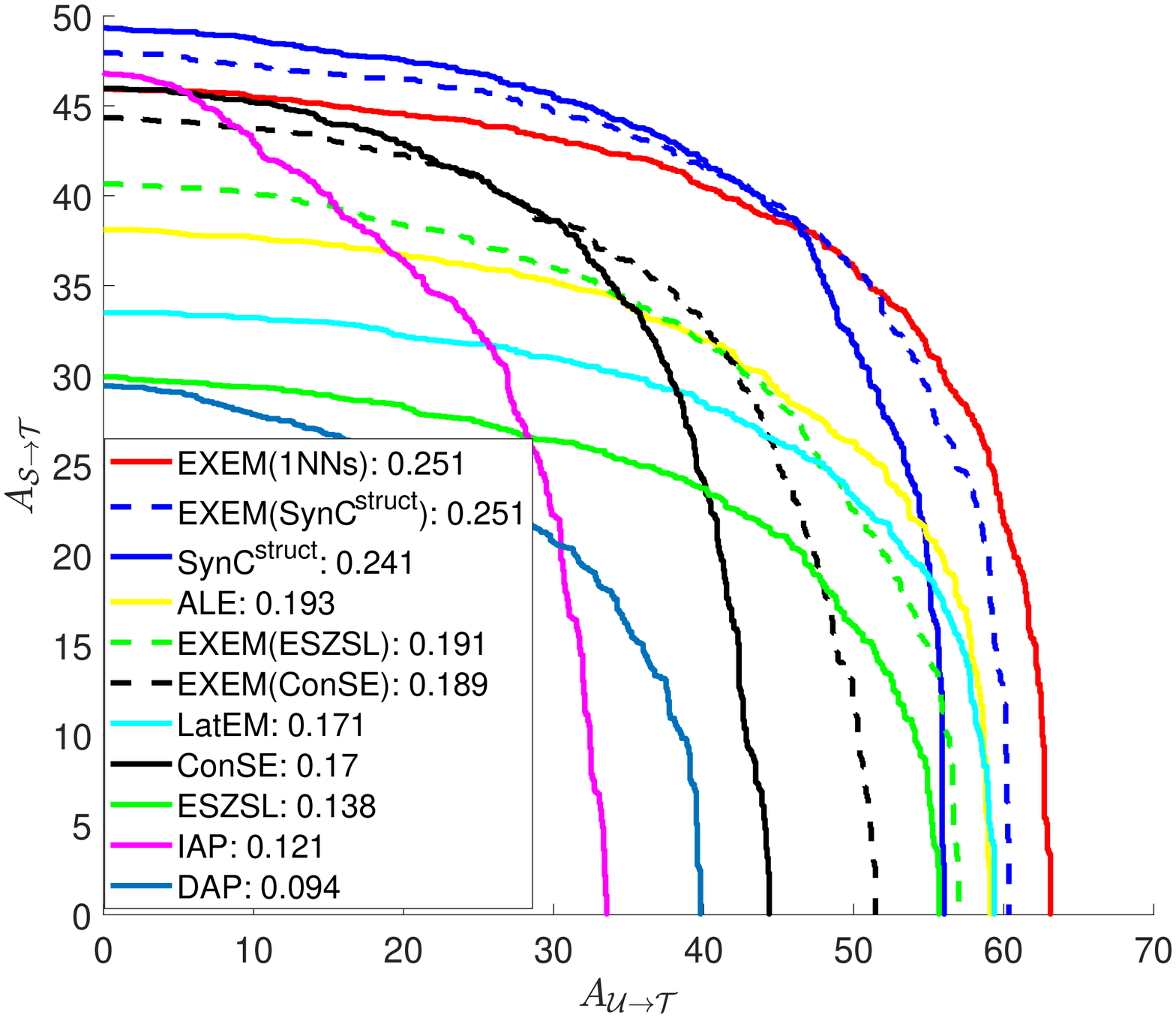}
	\caption{Seen-Unseen accuracy curves of ZSL methods on \sun. The area under each curve is also included in the legend. All approaches use calibrated stacking~\cite{ChaoCGS16} to combine the scores for seen and unseen classes, leading to curves of $(A_{\mathcal{U} \rightarrow \mathcal{T}}$, $A_{\mathcal{S} \rightarrow \mathcal{T}})$. Dashed lines correspond to ZSL methods that involve \method{EXEM}. We use ResNet features on the new split in all cases. Best viewed in color.}
	\label{fAUSUC}
	\vskip -0.5em
\end{figure}

\begin{table*}
\centering
\caption{Comparison between existing ZSL approaches on the task of GZSL based on the Area Under Seen-Unseen accuracy Curve (AUSUC)~\cite{ChaoCGS16} on small datasets. Each row corresponds to a ZSL method. Each column corresponds to a \emph{scenario} with a particular combination of dataset, its class split, and visual features. We use GoogLeNet features (G) and ResNet features (R). Class splits include both standard (SS or SS0) and new (NS) split. All approaches use calibrated stacking~\cite{ChaoCGS16} to combine the scores for seen and unseen classes. For each scenario, the best is in \bst{red} and the second best in \sbst{blue}.} \label{tbGZSLMain}
\vskip -0.5em
\scriptsize
\begin{tabular}{c|c|c|c|c|c|c|c|c|c|c|c|c|c|}
& \mtcl{2}{Reported by} & \mtcl{3}{\awa} & \mtcl{2}{\awatwo} & \mtcl{3}{\cub} & \mtcl{3}{\sun} \\ \hline
Features  & G & R & G & \mtcl{2}{R} & \mtcl{2}{R} & G & \mtcl{2}{R} &\mtcl{1}{G} & \mtcl{2}{R} \\ \hline
Approach/Splits	& \mtcl{2}{-} & SS & SS & NS & SS & NS  & SS  & SS0 & NS & SS & SS0 & NS \\ \hline
\zsldap & \gzsl & \tw & 0.366 & 0.402 & 0.341 & 0.423 & 0.353 & 0.194 & 0.204 & 0.200 & 0.096 & 0.087 & 0.094 \\ 
\zsliap & \gzsl & \tw & 0.394 & 0.452 & 0.376 & 0.466 & 0.392 & 0.199 & 0.215 & 0.209 & 0.145 & 0.128 & 0.121 \\ 
\zslconse & \gzsl & \tw & 0.428 & 0.486 & 0.350 & 0.521 & 0.344 & 0.212 & 0.226 & 0.214 & 0.200 & 0.182 & 0.170 \\ 
\zslale & \tw & \tw & 0.566 & 0.632 & 0.504 & 0.639 & 0.538 & 0.298 & 0.312 & 0.338 & 0.228 & 0.195 & 0.193 \\ 
\zsllatem & \tw & \tw & 0.551 & 0.632 & 0.506 & 0.639 & 0.514 & 0.284 & 0.290 & 0.276 & 0.201 & 0.169 & 0.171 \\ 
\zsleszsl & \gzsl & \tw & 0.490 & 0.591 & 0.452 & 0.625 & 0.454 & 0.304 & 0.311 & 0.303 & 0.168 & 0.138 & 0.138 \\ \hline
\zslsyncovo & \gzsl & \tw & 0.568 & 0.626 & 0.454 & 0.627 & 0.438 & 0.336 & 0.328 & 0.353 & 0.242 & 0.231 & 0.220 \\ 
\zslsynccs & \tw & \tw & \bst{0.593} & \sbst{0.651} & 0.477 & 0.658 & 0.463 & 0.322 & 0.329 & 0.359 & 0.212 & 0.209 & 0.189 \\ 
\zslsyncstr & \gzsl & \tw & 0.583 & 0.642 & 0.505 & 0.642 & 0.504 & 0.356 & 0.327 & 0.337 & 0.260 & \sbst{0.254} & 0.241 \\ \hline
\zslexemconse & \tw & \tw & 0.462 & 0.517 & 0.439 & 0.561 & 0.425 & 0.271 & 0.283 & 0.266 & 0.240 & 0.213 & 0.189 \\ 
\zslexemeszsl & \tw & \tw & 0.532 & 0.629 & 0.522 & 0.656 & 0.538 & 0.342 & 0.346 & 0.346 & 0.237 & 0.196 & 0.191 \\ 
\zslexemsyncovo & \exem & \tw & 0.553 & 0.643 & 0.481 & 0.668 & 0.474 & 0.365 & 0.360 & 0.361 & 0.265 & 0.243 & 0.221 \\ 
\zslexemsynccs & \tw & \tw & 0.563 & 0.649 & 0.497 & 0.670 & 0.481 & 0.347 & 0.321 & 0.360 & 0.230 & 0.205 & 0.205 \\ 
\zslexemsyncstr & \exem & \tw & \sbst{0.587} & \bst{0.674} & 0.533 & \sbst{0.687} & 0.552 & \bst{0.397} & \bst{0.397} & \bst{0.397} & \bst{0.288} & \bst{0.259} & \sbst{0.251} \\ 
\zslexemnn & \exem & \tw & 0.570 & 0.628 & \sbst{0.565} & 0.652 & \bst{0.565} & 0.318 & 0.313 & 0.298 & 0.284 & \sbst{0.254} & \bst{0.253} \\ 
\zslexemnns & \exem & \tw & 0.584 & 0.650 & \bst{0.575} & \bst{0.688} & \sbst{0.559} & \sbst{0.373} & \sbst{0.365} & \sbst{0.366} & \sbst{0.287} & 0.250 & \sbst{0.251} \\ \hline
\end{tabular}
\end{table*}

\begin{table*}
	\centering
	\caption{Comparison between different metrics on the task of GZSL on small datasets. We consider (i) ``H w/o calibration": the harmonic mean without calibrated stacking, (ii) ``H w/ calibration": the harmonic mean with calibrated stacking where the calibrating factor is selected with cross-validation, and (iii) the Area Under Seen-Unseen accuracy Curve (AUSUC) with calibrated stacking where all possible calibrating factors are integrated over. $A_{\mathcal{U} \rightarrow \mathcal{T}}$ and $A_{\mathcal{S} \rightarrow \mathcal{T}}$used to compute H are also included. Calibrated stacking was introduced in \cite{ChaoCGS16}. We focus on ResNet features (R) and the new split (NS) in all cases. For each scenario, the best is in \bst{red} and the second best in \sbst{blue}.} \label{tbGZSLHarmonic}
	\vskip -0.5em
	\tabcolsep 2.3pt
	\scriptsize
\begin{tabular}{c|c|c|c|c|c|c|c|c|c|c|c|c|c|c|c|c|}
& \mtcl{2}{Reported by} & \mtcl{7}{\awa} & \mtcl{7}{\awatwo} \\\hline
& w/o & w/ & \mtcl{3}{w/o calibration} & \mtcl{4}{w/ calibration} & \mtcl{3}{w/o calibration} & \mtcl{4}{w/ calibration} \\ \hline
Approach/Metric & \mtcl{2}{-} &$A_{\mathcal{U} \rightarrow \mathcal{T}}$ & $A_{\mathcal{S} \rightarrow \mathcal{T}}$ & H & $A_{\mathcal{U} \rightarrow \mathcal{T}}$ & $A_{\mathcal{S} \rightarrow \mathcal{T}}$ & H & AUSUC & $A_{\mathcal{U} \rightarrow \mathcal{T}}$ & $A_{\mathcal{S} \rightarrow \mathcal{T}}$ & H & $A_{\mathcal{U} \rightarrow \mathcal{T}}$ & $A_{\mathcal{S} \rightarrow \mathcal{T}}$ & H & AUSUC \\ \hline 
\zsldap & \gbu & \tw & 0.0 & \bst{88.7} & 0.0 & 37.8 & 63.9 & 47.5 & 0.341 & 0.0 & 84.7 & 0.0 & 39.3 & 67.5 & 49.7 & 0.353 \\ 
\zsliap & \gbu & \tw & 2.1 & 78.2 & 4.1 & 38.6 & 71.6 & 50.1 & 0.376 & 0.9 & 87.6 & 1.8 & 41.5 & 71.9 & 52.7 & 0.392 \\ 
\zslconse & \gbu & \tw & 0.4 & \sbst{88.6} & 0.8 & 37.8 & 59.9 & 46.4 & 0.350 & 0.5 & 90.6 & 1.0 & 35.8 & 62.9 & 45.6 & 0.344 \\ 
\zslale & \gbu & \tw & 16.8 & 76.1 & 27.5 & 48.6 & 76.6 & 59.4 & 0.504 & 14.0 & 81.8 & 23.9 & 42.9 & \bst{84.2} & 56.8 & 0.538 \\ 
\zsllatem & \gbu & \tw & 7.3 & 71.7 & 13.3 & 48.5 & 78.3 & 60.0 & 0.506 & 11.5 & 77.3 & 20.0 & 39.9 & \sbst{83.8} & 54.0 & 0.514 \\ 
\zsleszsl & \gbu & \tw & 6.6 & 75.6 & 12.1 & 50.4 & 70.7 & 58.8 & 0.452 & 5.9 & 77.8 & 11.0 & 49.4 & 73.4 & 59.1 & 0.454 \\ \hline
\zslsyncovo & \tw & \tw & 4.1 & 80.7 & 7.7 & 52.7 & 68.7 & 59.6 & 0.454 & 0.5 & 89.0 & 0.9 & 48.1 & 74.0 & 58.3 & 0.438 \\ 
\zslsynccs & \tw & \tw & 3.3 & 88.2 & 6.4 & 50.8 & 75.9 & 60.9 & 0.477 & 0.7 & \bst{92.3} & 1.4 & 47.6 & 79.9 & 59.6 & 0.463 \\ 
\zslsyncstr & \tw & \tw & 7.7 & 84.6 & 14.1 & 52.7 & 76.4 & 62.4 & 0.505 & 4.5 & 90.8 & 8.7 & 50.1 & 81.8 & 62.1 & 0.504 \\ \hline
\zslexemconse & \tw & \tw & 12.5 & 86.5 & 21.9 & 54.9 & 45.9 & 50.0 & 0.439 & \sbst{21.8} & 79.5 & \sbst{34.2} & 52.0 & 53.7 & 52.9 & 0.425 \\ 
\zslexemeszsl & \tw & \tw & 12.4 & 85.7 & 21.7 & \bst{59.9} & 66.7 & 63.1 & 0.522 & 10.4 & 88.3 & 18.7 & \bst{59.6} & 68.0 & 63.5 & 0.538 \\ 
\zslexemsyncovo & \tw & \tw & 6.2 & 87.5 & 11.6 & 46.7 & \bst{79.9} & 58.9 & 0.481 & 5.8 & 89.1 & 10.8 & 48.1 & 79.8 & 60.0 & 0.474 \\ 
\zslexemsynccs & \tw & \tw & 8.8 & 84.4 & 15.9 & 48.4 & 78.6 & 59.9 & 0.497 & 9.9 & 90.7 & 17.8 & 47.4 & 81.8 & 59.8 & 0.481 \\ 
\zslexemsyncstr & \tw & \tw & 14.8 & 85.6 & 25.3 & 54.6 & 76.0 & 63.5 & 0.533 & 14.5 & \sbst{91.7} & 25.1 & 54.0 & 78.5 & 64.0 & 0.552 \\ 
\zslexemnn & \tw & \tw & \sbst{21.8} & 83.8 & \sbst{34.6} & 57.2 & 75.7 & \sbst{65.2} & \sbst{0.565} & 18.3 & 86.9 & 30.2 & \sbst{55.7} & 78.3 & \sbst{65.1} & \bst{0.565} \\ 
\zslexemnns & \tw & \tw & \bst{31.6} & 88.1 & \bst{46.5} & \sbst{57.4} & \sbst{78.7} & \bst{66.4} & \bst{0.575} & \bst{30.8} & 89.3 & \bst{45.8} & 55.0 & 82.2 & \bst{65.9} & \sbst{0.559} \\ \hline
\end{tabular}
\vskip 5pt
\begin{tabular}{c|c|c|c|c|c|c|c|c|c|c|c|c|c|c|c|c|}
& \mtcl{2}{Reported by} & \mtcl{7}{\cub} & \mtcl{7}{\sun} \\\hline
& w/o & w/ & \mtcl{3}{w/o calibration} & \mtcl{4}{w/ calibration} & \mtcl{3}{w/o calibration} & \mtcl{4}{w/ calibration} \\ \hline
Approach/Metric & \mtcl{2}{-} &$A_{\mathcal{U} \rightarrow \mathcal{T}}$ & $A_{\mathcal{S} \rightarrow \mathcal{T}}$ & H & $A_{\mathcal{U} \rightarrow \mathcal{T}}$ & $A_{\mathcal{S} \rightarrow \mathcal{T}}$ & H & AUSUC &$A_{\mathcal{U} \rightarrow \mathcal{T}}$ & $A_{\mathcal{S} \rightarrow \mathcal{T}}$ & H & $A_{\mathcal{U} \rightarrow \mathcal{T}}$ & $A_{\mathcal{S} \rightarrow \mathcal{T}}$ & H & AUSUC \\ \hline 
\zsldap & \gbu & \tw & 1.7 & 67.9 & 3.3 & 33.9 & 35.8 & 34.8 & 0.200 & 4.2 & 25.1 & 7.2 & 23.8 & 23.5 & 23.7 & 0.094 \\ 
\zsliap & \gbu & \tw & 0.2 & \bst{72.8} & 0.4 & 31.4 & 41.7 & 35.7 & 0.209 & 1.0 & 37.8 & 1.8 & 25.1 & 32.4 & 28.3 & 0.121 \\ 
\zslconse & \gbu & \tw & 1.6 & \sbst{72.2} & 3.1 & 33.7 & 38.9 & 36.1 & 0.214 & 6.8 & 39.9 & 11.6 & 34.0 & 34.8 & 34.4 & 0.170 \\ 
\zslale & \gbu & \tw & \sbst{23.7} & 62.8 & \sbst{34.4} & 50.7 & 45.7 & 48.1 & 0.338 & \bst{21.8} & 33.1 & \sbst{26.3} & 40.6 & 31.9 & 35.7 & 0.193 \\ 
\zsllatem & \gbu & \tw & 15.2 & 57.3 & 24.0 & 47.0 & 38.2 & 42.2 & 0.276 & 14.7 & 28.8 & 19.5 & 36.3 & 29.6 & 32.6 & 0.171 \\ 
\zsleszsl & \gbu & \tw & 12.6 & 63.8 & 21.0 & 48.9 & 41.9 & 45.1 & 0.303 & 11.0 & 27.9 & 15.8 & 35.0 & 25.2 & 29.3 & 0.138 \\ \hline
\zslsyncovo & \tw & \tw & 9.8 & 66.7 & 17.0 & \bst{53.1} & 41.6 & 46.7 & 0.353 & 8.8 & 44.5 & 14.6 & \bst{46.6} & 30.9 & 37.1 & 0.220 \\ 
\zslsynccs & \tw & \tw & 9.7 & 69.7 & 17.0 & \sbst{52.0} & 44.9 & 48.2 & 0.359 & 6.1 & 46.5 & 10.8 & 38.6 & 33.8 & 36.1 & 0.189 \\ 
\zslsyncstr & \tw & \tw & 15.4 & 69.0 & 25.2 & 50.1 & 45.4 & 47.6 & 0.337 & 7.8 & 45.7 & 13.3 & 41.0 & \bst{41.4} & \sbst{41.2} & 0.241 \\ \hline
\zslexemconse & \tw & \tw & 13.4 & 69.8 & 22.5 & 40.5 & 36.4 & 38.3 & 0.266 & 7.7 & 43.9 & 13.1 & 35.3 & 36.2 & 35.8 & 0.189 \\ 
\zslexemeszsl & \tw & \tw & 8.8 & 68.9 & 15.6 & 45.9 & 54.5 & 49.9 & 0.346 & 5.8 & 40.4 & 10.2 & 22.8 & 37.9 & 28.5 & 0.191 \\ 
\zslexemsyncovo & \tw & \tw & 16.1 & 70.2 & 26.2 & 47.4 & 56.3 & \sbst{51.5} & 0.361 & 11.0 & 44.6 & 17.6 & 37.8 & 39.4 & 38.6 & 0.221 \\ 
\zslexemsynccs & \tw & \tw & 18.4 & 68.6 & 29.1 & 44.9 & \bst{58.4} & 50.8 & 0.360 & 7.4 & \sbst{46.7} & 12.8 & 36.9 & 37.6 & 37.3 & 0.205 \\ 
\zslexemsyncstr & \tw & \tw & 22.5 & 71.1 & 34.1 & 50.8 & \sbst{58.1} & \bst{54.2} & \bst{0.397} & 12.3 & \bst{47.2} & 19.5 & 41.0 & \sbst{41.2} & 41.1 & \sbst{0.251} \\ 
\zslexemnn & \tw & \tw & 21.4 & 58.7 & 31.3 & 46.5 & 45.7 & 46.1 & 0.298 & \sbst{20.1} & 39.0 & \bst{26.6} & 42.9 & 40.4 & \bst{41.6} & \bst{0.253} \\ 
\zslexemnns & \tw & \tw & \bst{28.0} & 67.8 & \bst{39.6} & 49.8 & 52.1 & 50.9 & \sbst{0.366} & 14.6 & 42.0 & 21.6 & \sbst{43.5} & 39.1 & \sbst{41.2} & \sbst{0.251} \\ \hline
\end{tabular}
\end{table*}

We now present our results on generalized zero-shot learning (GZSL). We focus on \awa, \awatwo, \cub, and \sun because all \imn's images from seen classes are used either for pre-training for feature extraction or for hyper-parameter tuning.
As in conventional zero-shot learning experiments, we include popular scenarios for ZSL that were investigated by past work and all ZSL methods use visual attributes as semantic representations.

We first present our main results on GZSL in Table~\ref{tbGZSLMain}. We use the Area Under Seen-Unseen accuracy curve with calibrated stacking (AUSUC)~\cite{ChaoCGS16}. Calibrated stacking introduces a calibrating factor that adaptively changes how we combine the scores for seen and unseen classes. AUSUC is the final score that integrates over all possible values of this factor. Besides similar trends stated in Sect.~\ref{sMainExp} and Sect.~\ref{sExpResZSL}, we notice that \method{EXEM (SynC$^\textrm{struct}$)} performs particularly well in the GZSL setting. For instance, the relative performance of \method{EXEM (1NN)} against \method{EXEM (SynC$^\textrm{struct}$)} drops when moving from conventional ZSL to generalized ZSL.

To illustrate why one method may perform better or worse than another, we show the Seen-Unseen accuracy curves of ZSL methods on \sun in Fig.~\ref{fAUSUC}. We observe that a method might perform well on one axis but poorly on the other. For example, \method{IAP} outperforms \method{DAP} on $A_{\mathcal{S} \rightarrow \mathcal{T}}$ but not on $A_{\mathcal{U} \rightarrow \mathcal{T}}$. As another example, \method{ESZSL}, \method{ALE}, and \method{LatEm} achieve similar $A_{\mathcal{U} \rightarrow \mathcal{T}}$ to the one by \method{SynC$^\textrm{struct}$} but perform significantly worse on $A_{\mathcal{S} \rightarrow \mathcal{T}}$, resulting in a lower AUSUC. Our results hence emphasize once again the importance of the GZSL setting evaluation.

\subsubsection{Comparison among evaluation metrics}

\begin{figure}
	\centering
	\includegraphics[width=0.45\textwidth]{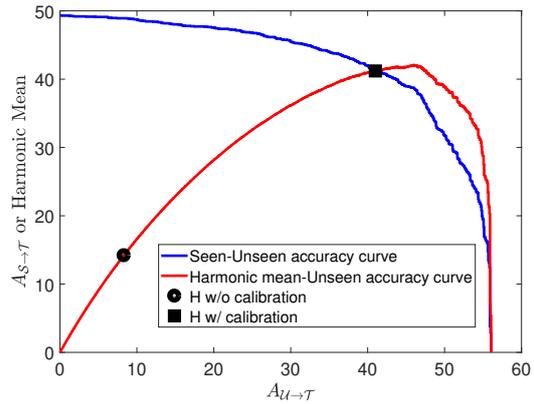}
	\caption{How calibrated stacking affects the harmonic mean of \method{SynC$^\textrm{struct}$} on \sun. The blue curve is the Seen-Unseen accuracy curve, where we plot $A_{\mathcal{U} \rightarrow \mathcal{T}}$ (y-axis) vs. $A_{\mathcal{S} \rightarrow \mathcal{T}}$ (x-axis). The red curve is the Harmonic mean-Unseen accuracy curve, where we plot the harmonic mean (y-axis) vs. $A_{\mathcal{U} \rightarrow \mathcal{T}}$ (x-axis). The heart and square correspond to the harmonic mean without and with calibrated stacking, respectively.
	We see that calibrated stacking drastically improves the value of harmonic mean and that the Harmonic mean-Unseen accuracy curve is biased toward $A_{\mathcal{U} \rightarrow \mathcal{T}}$ (left-skewed). We use ResNet features on the new split. Best viewed in color.}
	\label{fAUSUC_vs_HM}
\end{figure}

We then focus on ResNet features and the new splits and present an empirical comparison between different metrics used for GZSL in Table~\ref{tbGZSLHarmonic}. First, we consider the harmonic mean of $A_{\mathcal{U} \rightarrow \mathcal{T}}$ and $A_{\mathcal{S} \rightarrow \mathcal{T}}$, as in \cite{XianLSA17}. Second, we consider the ``calibrated" harmonic mean; we propose to select the calibrating factor (cf. Eq.~(\ref{eGZSLPredictCalib})) for each ZSL method using cross-validation, resulting in new values of $A_{\mathcal{U} \rightarrow \mathcal{T}}$ and $A_{\mathcal{S} \rightarrow \mathcal{T}}$ and hence a new value for the harmonic mean. Finally, we use the AUSUC with calibrated stacking \cite{ChaoCGS16} as in Table~\ref{tbGZSLMain}. See Appendix~\ref{sSuppHyperTuning} for details on hyper-parameter tuning for each metric.

Fig.~\ref{fAUSUC_vs_HM} illustrates what happens after we apply calibrated stacking. We plot two curves based on results by \method{SynC$^\textrm{struct}$} on \sun. One (blue) is the Seen-Unseen accuracy curve \cite{ChaoCGS16}. The other (red) is the harmonic mean vs. $A_{\mathcal{U} \rightarrow \mathcal{T}}$, which we will call the Harmonic mean-Unseen accuracy curve. In other words, as we vary the calibrating factor, the harmonic mean changes. The uncalibrated harmonic mean (heart) and the calibrated harmonic mean (square) reported in Table~\ref{tbGZSLMain} are also shown.
Clearly, we see a large improvement in the harmonic mean with calibration. Furthermore, we see that the harmonic mean curve is left-skewed; it goes up until $A_{\mathcal{U} \rightarrow \mathcal{T}}$ nearly reaches 50\%.   

We have the following important observations and implications.
First, we discuss critical issues with the uncalibrated harmonic mean metric. We observe that it is correlated with the standard metric used in zero-shot learning $A_{\mathcal{U} \rightarrow \mathcal{T}}$ and that it can be made much higher after calibration. ZSL methods that have bias toward predicting a label from unseen classes can perform well under this metric, while in fact other methods may do just as well or better when this bias is calibrated. For instance, \method{ConSE} and \method{SynC$^\textrm{struct}$} become much more competitive under the calibrated harmonic mean metric. Fig.~\ref{fAUSUC_vs_HM} also evidently supports this observation. \emph{We therefore conclude that the uncalibrated harmonic mean may be a misleading metric in the GZSL evaluation.}

Second, we discuss the calibrated harmonic mean and AUSUC. We observe a certain degree of positive correlation between the two metrics, but exceptions exist. For example, on \awa, one might mistakenly conclude that \method{EXEM (SynC$^\textrm{o-vs-o}$)} does not improve over \method{SynC$^\textrm{o-vs-o}$} (i.e., predicting exemplars does not help) while AUSUC says the opposite. \emph{We therefore advocate using both evaluation metrics in the GZSL evaluation.}


\subsection{Additional Results on \imn}
\label{sExpResAdd}

\subsubsection{Our approaches with other types of semantic representations}
\label{sExpResAddSem}

\begin{table*}
\centering
\caption{\small Comparison between different types of semantic representations on \imn. We consider (i) ``wv-v1": word vectors of the class names trained for one epoch used in Table~\ref{tbZSLMainIMN}, (ii) ``wv-v2": word vectors of the class names trained for 10 epochs, (iii) ``hie": the WordNet-hierarchy embeddings obtained using multidimensional scaling \cite{Lu16}, (iv) ``wv-v1 + hie": the combination of (i) and (iii), (v) ``wv-v2 + hie": the combination of (ii) and (iii). We use \textbf{``per-class"} accuracy (in \%) and ResNet features in all cases. For each scenario, the best is in \bst{red} and the second best in \sbst{blue}.} \label{tbZSLSemanticPCIMN}
\vskip -0.5em
\begin{tabular}{c|c|c|c|c|c|c|c|c|c|c|c|c|}
Approach & Semantic & \multicolumn{2}{c|}{Hierarchy} & \multicolumn{3}{c|}{Most populated} & \multicolumn{3}{c|}{Least populated} & All \\ \cline{3-10}
& types & 2-hop & 3-hop & 500 & 1K & 5K & 500 & 1K & 5K & \\ \hline
\zslsyncovo &  & 9.60 & 2.31 & 16.38 & 11.14 & 3.50 & 5.47 & 3.83 & 1.34 & 0.98 \\ 
\zslsyncstr &  & 8.76 & 2.25 & 14.93 & 10.33 & 3.44 & 4.20 & 3.22 & 1.26 & 0.99 \\ 
\zslexemsyncovo & wv-v1 & 11.15 & 2.95 & \bst{19.26} & \bst{13.37} & 4.50 & \sbst{6.33} & 4.48 & 1.62 & 1.25 \\ 
\zslexemsyncstr &  & \sbst{11.28} & \bst{3.02} & 18.97 & \sbst{13.26} & \bst{4.57} & 6.10 & \bst{4.74} & 1.67 & \bst{1.29} \\ 
\zslexemnn &  & 10.60 & 2.87 & 18.15 & 12.68 & 4.40 & 6.20 & 4.61 & \sbst{1.68} & 1.26 \\ 
\zslexemnns &  & \bst{11.58} & \sbst{2.99} & \sbst{19.09} & 13.16 & \sbst{4.55} & \bst{6.70} & \sbst{4.67} & \bst{1.70} & \bst{1.29} \\ \hline
\zslsyncovo &  & 12.56 & 3.04 & 19.04 & 13.01 & 4.39 & 5.33 & 3.77 & 1.79 & 1.25 \\ 
\zslsyncstr &  & 11.59 & 2.98 & 17.78 & 12.53 & 4.24 & 6.07 & 3.87 & 1.69 & 1.25 \\ 
\zslexemsyncovo & wv-v2 & 13.79 & 3.58 & \bst{21.63} & \bst{14.86} & 5.14 & 6.60 & 5.03 & 2.14 & 1.48 \\ 
\zslexemsyncstr &  & \bst{14.12} & \bst{3.67} & \sbst{21.47} & 14.81 & \bst{5.27} & 6.73 & 4.38 & \bst{2.15} & \bst{1.52} \\ 
\zslexemnn &  & 13.19 & 3.47 & 20.80 & 14.40 & 5.10 & \bst{7.23} & \sbst{5.05} & 2.02 & 1.48 \\ 
\zslexemnns &  & \sbst{14.09} & \sbst{3.62} & 21.31 & \sbst{14.84} & \sbst{5.24} & \sbst{7.17} & \bst{5.39} & \bst{2.15} & \sbst{1.51} \\ \hline
\zslsyncovo &  & 19.77 & 3.90 & 15.37 & 11.53 & 4.73 & 9.30 & 6.71 & 2.41 & 1.47 \\ 
\zslsyncstr &  & 19.37 & 3.81 & 14.58 & 11.13 & 4.64 & 8.27 & 6.20 & 2.45 & 1.44 \\ 
\zslexemsyncovo & hie & 21.65 & 4.29 & \sbst{16.94} & \sbst{12.79} & 5.07 & \bst{10.07} & \bst{7.22} & \sbst{2.73} & 1.62 \\ 
\zslexemsyncstr &  & \bst{22.30} & \bst{4.42} & \bst{17.11} & \bst{13.07} & \bst{5.23} & \sbst{9.80} & \sbst{7.12} & \bst{2.74} & \bst{1.67} \\ 
\zslexemnn &  & 20.95 & 4.12 & 16.70 & 12.48 & 4.95 & 9.50 & 6.93 & 2.58 & 1.56 \\ 
\zslexemnns &  & \sbst{21.93} & \sbst{4.33} & \sbst{16.94} & 12.68 & \sbst{5.14} & 8.87 & 6.73 & 2.64 & \sbst{1.63} \\ \hline
\zslsyncovo &  & 20.42 & 4.35 & 19.97 & 14.32 & 5.43 & 8.60 & 6.40 & 2.84 & 1.69 \\ 
\zslsyncstr & wv-v1 & 21.04 & 4.36 & 18.14 & 13.38 & 5.23 & 8.83 & 6.59 & 2.78 & 1.66 \\ 
\zslexemsyncovo & + & 23.00 & 5.12 & 23.23 & 17.00 & 6.33 & \bst{11.50} & \bst{8.17} & \sbst{3.16} & 2.01 \\ 
\zslexemsyncstr & hie & \bst{24.20} & \bst{5.39} & \bst{23.74} & \bst{17.24} & \bst{6.59} & \sbst{10.40} & \sbst{7.90} & \sbst{3.16} & \bst{2.11} \\ 
\zslexemnn &  & 21.34 & 4.83 & 22.39 & 16.32 & 6.03 & 10.00 & 7.48 & 2.98 & 1.89 \\ 
\zslexemnns &  & \sbst{23.33} & \sbst{5.14} & \sbst{23.65} & \sbst{17.06} & \sbst{6.48} & 9.97 & 7.70 & \bst{3.23} & \sbst{2.02} \\ \hline
\zslsyncovo &  & 21.22 & 4.49 & 20.68 & 14.97 & 5.40 & 10.77 & 7.35 & 2.98 & 1.71 \\ 
\zslsyncstr & wv-v2 & 20.21 & 4.24 & 18.22 & 13.26 & 5.07 & 8.47 & 6.01 & 2.65 & 1.62 \\ 
\zslexemsyncovo & + & \sbst{23.64} & \sbst{5.33} & 24.40 & \sbst{17.80} & \sbst{6.68} & 12.33 & \bst{8.38} & \bst{3.45} & \sbst{2.09} \\ 
\zslexemsyncstr & hie & \bst{24.48} & \bst{5.56} & \bst{24.63} & \bst{18.01} & \bst{6.87} & \sbst{12.77} & \sbst{7.99} & 3.32 & \bst{2.18} \\ 
\zslexemnn &  & 22.40 & 5.17 & 23.90 & 17.37 & 6.47 & \bst{13.07} & 7.83 & 3.16 & 2.05 \\ 
\zslexemnns &  & 22.70 & 5.21 & \bst{24.63} & \sbst{17.80} & \sbst{6.68} & 12.23 & 7.81 & \sbst{3.37} & 2.06 \\ \hline
\end{tabular}
\end{table*}

How much do different types of semantic representations affect the performance of our ZSL algorithms? We focus on the \imn dataset and investigate this question in detail. 
In the main \imn experiments, we consider the word vectors derived from a skip-gram model \cite{MikolovCCD13}. In this section, we obtain higher-quality word vectors and consider another type of semantic representations derived from a class hierarchy.

First, we train a skip-gram model in the same manner (the same corpus, the same vector dimension, etc.) as in Sect.~\ref{sExpSetupSem} but we let it train for 10 epochs instead of for one epoch that was used in \zsldevise. We call the word vectors from one-epoch and 10-epoch training  ``word vectors version 1 (wv-v1)" and ``word vectors version 2 (wv-v2)," respectively. Additionally, we derive 21,632 dimensional semantic vectors of the class names using multidimensional scaling (MDS) on the WordNet hierarchy \cite{Miller95}, following \cite{Lu16}. We denote such semantic vectors ``hie." As before, we normalize each semantic representation to have a unit $\ell_2$ norm unless stated otherwise. Finally, we consider the combination of either version of word vectors and the hierarchy embeddings. As both \method{SynC} and \method{EXEM} use the RBF kernel in computing the semantic relatedness among classes (cf. Eq.~(\ref{eSynCSimForm}) and Eq.~(\ref{eEXEMlearn})), we perform convex combination of the kernels from two types of semantic representations instead of directly concatenating them. The combination weight scalar is a hyper-parameter to be tuned.

\paragraph{Main Results:} In Table~\ref{tbZSLSemanticPCIMN}, we see how improved semantic representations lead to substantially improved ZSL performances. In particular, word vectors trained for a larger number of iterations can already improve the overall accuracy by an absolute 0.2-0.3\%. The hierarchy embeddings improve the performances \emph{further} by an absolute 0.1-0.2\%. Finally, we see that these two types of semantic representations are complementary; the combination of either version of word vectors and the hierarchy embeddings improves over either the word vectors or the hierarchy embeddings alone. In the end, the best result we obtain is 2.18\% by \method{EXEM (SynC$^\textrm{struct}$)} with ``wv-v2 + hie," achieving a 69\% improvement over the word vectors ``wv-v1."

\begin{figure*}
	\centering
	\includegraphics[width=1.0\textwidth]{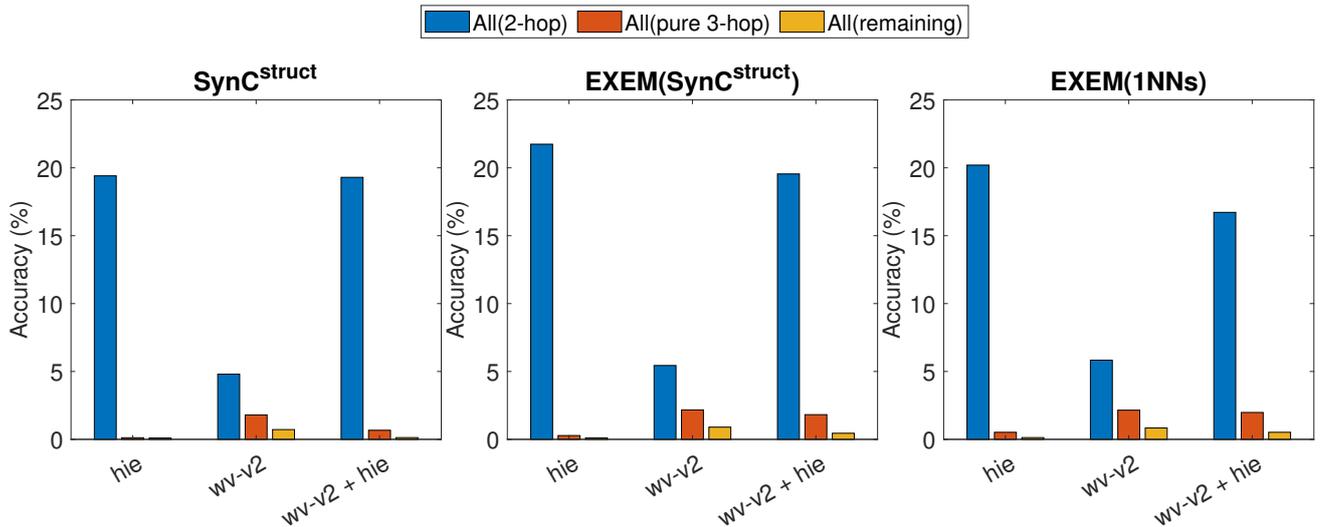}
	\caption{{Contribution of unseen classes to the performance (per-class accuracy) of \zslsyncstr, \zslexemsyncstr, and \zslexemnns on zero-shot classification to All (20,345 unseen classes) given the ``wv-v2," ``hie," ``wv-v2 + hie" semantic representations (cf. Table~\ref{tbZSLSemanticPCIMN} and Sect.~\ref{sExpSetupZSL}). We report the average accuracies on three \emph{disjoint} subsets of All: (i) All (2-hop), (ii) All (pure 3-hop), and (iii) All (remaining), corresponding respectively to the unseen classes within two, \textbf{exactly} three, and \textbf{more than} three tree hops from the 1K seen classes according to the WordNet hierarchy. See text for details.}}
	\label{fw2vMDS}
\end{figure*}

\begin{figure*}[!htb]
\centering
\includegraphics[width=0.40\textwidth]{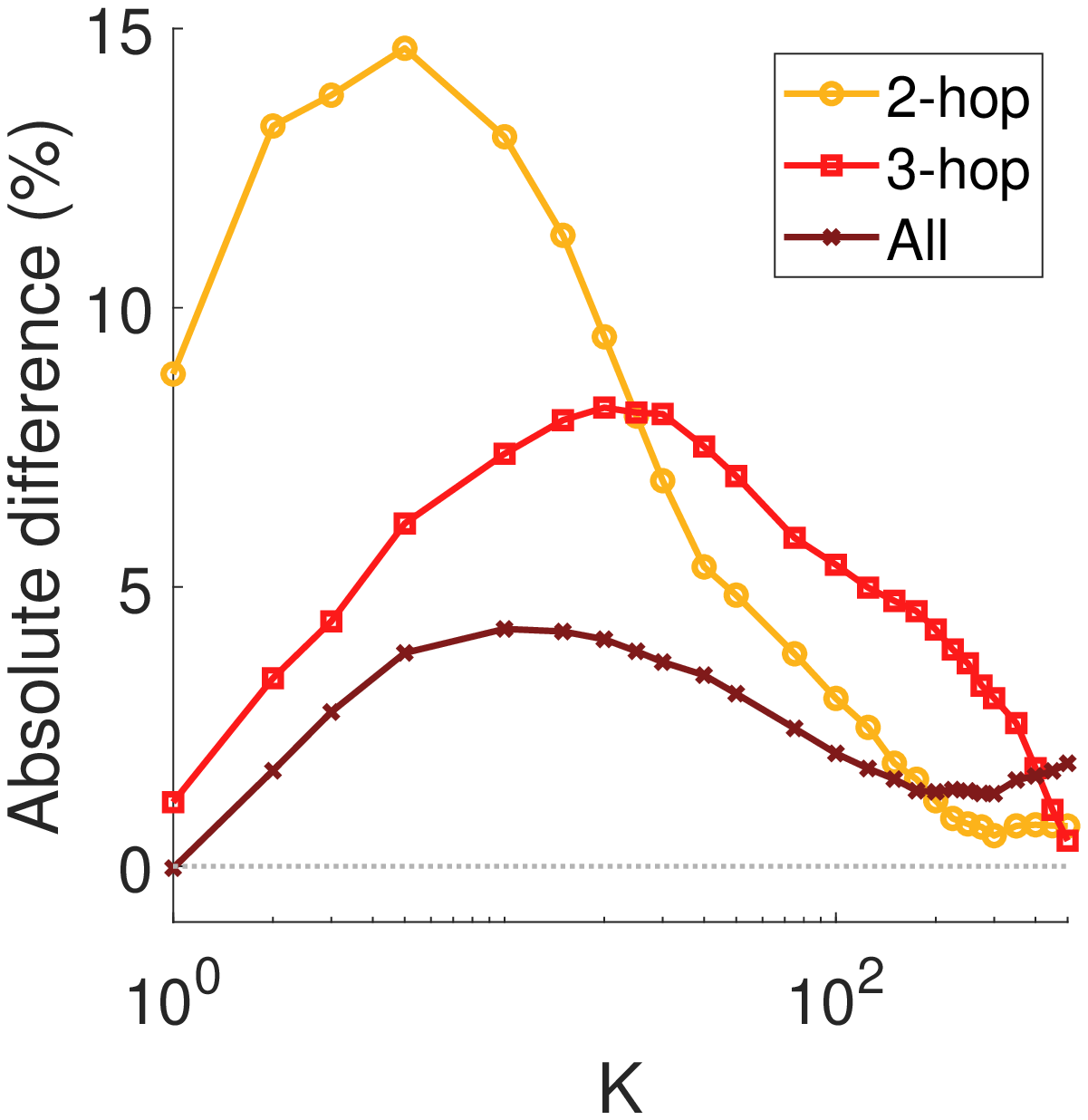}
\includegraphics[width=0.58\textwidth]{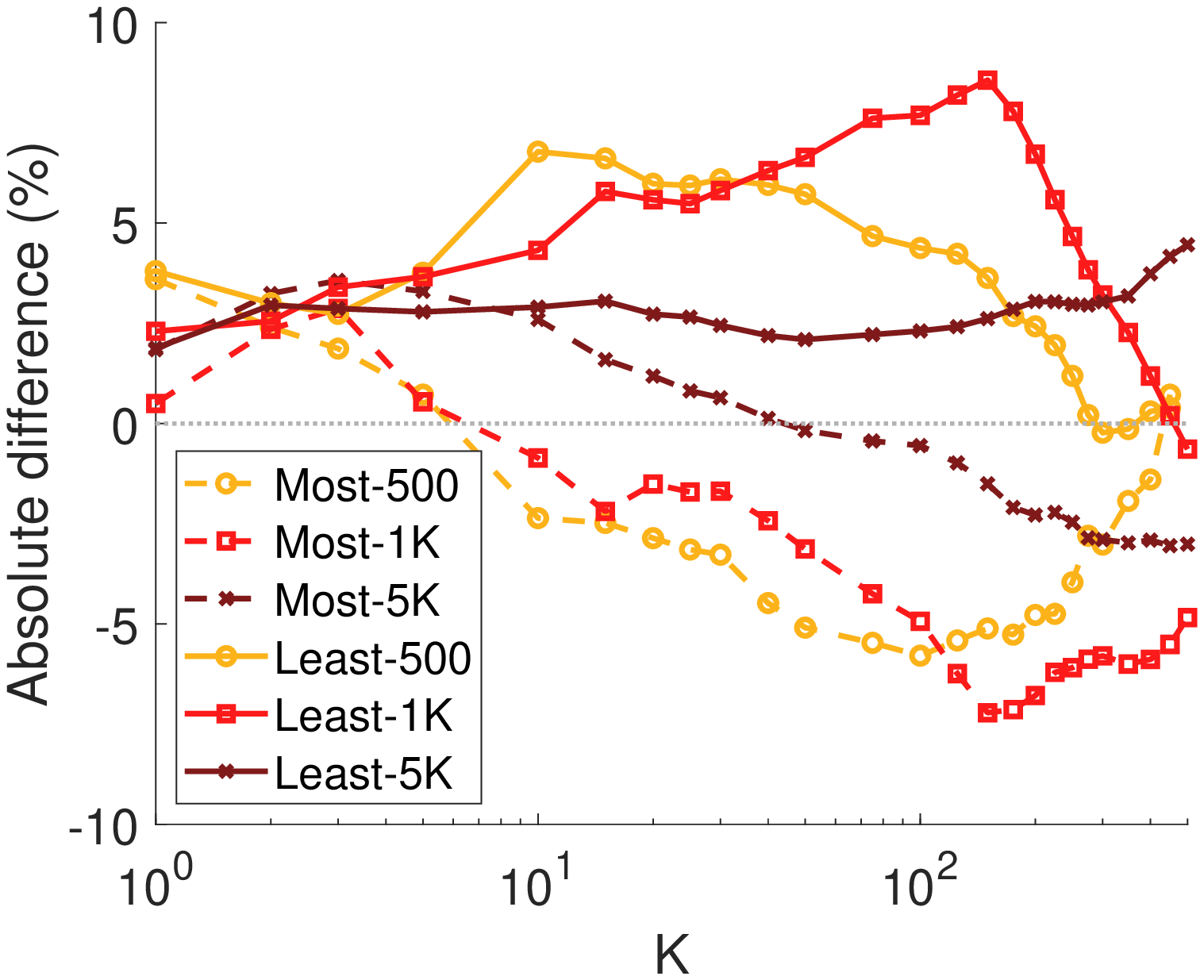}
\caption{{For a fixed $K$, a positive value of ``absolute difference" (\%) indicates ``hie" is better at encoding class similarity than ``wv-v2" \emph{within the neighborhoods of size $K$}, where ``better" means ``closer to class similarity based on visual exemplars." Specifically, we compute the overlap of $K$ nearest neighboring classes (i) between using ``hie" (the hierarchy embeddings based on WordNet) and using ``visual exemplars," and (ii) between using ``wv-v2" (word vectors) and using ``visual exemplars." We compute the overlaps within each test subset of unseen classes and then plot (i) minus (ii) for $K$ from 1 to 500. See text for details.}}  \label{fCompareW2VMDSKnn}
\end{figure*}

{
\paragraph{Word vectors vs. hierarchy embeddings} We then inspect ``wv-v2" and ``hie" in more detail. Firstly, the performance gaps between ``hie" and ``wv-v2" are reduced when we consider larger test subsets of unseen classes. Secondly, there are a few exceptions to the general trend that ``hie" is of higher quality than ``wv-v2" as semantic representations for ZSL. When evaluated on different test subsets (the columns of Table~\ref{tbZSLSemanticPCIMN}; see also Sect.~\ref{sExpSetupZSL} for their descriptions), ``hie" leads to noticeably worse performance on the 500 and 1K most populated unseen classes, and comparable performance on the 5K most populated ones. To better understand such observations, we provide two sets of analysis.

The first set of analysis breaks down the overall ZSL performance; which unseen classes most contribute to the performance of a ZSL algorithm on All (20,345 unseen classes) for both types of semantic representations and their combination (cf. Table 7)? We consider the following three \emph{disjoint} subsets of All: (i) All (2-hop), (ii) All (pure 3-hop), and (iii) All (remaining). Similar but different to the provided definitions in Sect.~\ref{sExpSetupZSL}, they correspond to unseen classes that are two, \textbf{exactly} three, and \textbf{more than} three tree hops from the 1K seen classes, respectively; the main difference is that the label space here is always taken to be that of \emph{All}. We report accuracies averaged over classes on these subsets for \zslsyncstr, \zslexemsyncstr, and \zslexemnns.

As shown in Fig.~\ref{fw2vMDS}, ``hie" outperforms ``wv-v2" on All (2-hop) on all three ZSL methods but the reverse is observed for All (pure 3-hop) and All (remaining). This explains our first observation regarding the reduced performance gap of larger test subsets. In addition, this suggests that the way we obtain hierarchy embeddings favors unseen classes that are \emph{semantically close} to seen ones. Fortunately, ``wv-v2" can provide significantly improved representations for semantically far away unseen classes where ``hie" are less useful, as illustrated in ``wv-v2 + hie."

The second set of analysis aims at understanding an intrinsic quality of semantic representations. As ZSL algorithms rely on class similarities from semantic representations, intuitively the neighborhood structures of semantic representations should play a critical role in the \emph{downstream} zero-shot classification task. We investigate this by comparing the two semantic representations' neighborhood structures. We use the degree of \emph{closeness} to the structure of visual exemplars (cf. Sect.~\ref{sEXEMPredExem}) as a proxy for semantic representations' \emph{intrinsic} quality.

Formally, for a constant $K$, let kNN$_A(u)$ be the $K$ nearest unseen classes of an unseen class $u$ based on semantic representations $A$. Let \%kNN-overlap($A1$, $A2$) be the average (over all unseen classes $u$) of the percentages of overlap between kNN$_{A1}(u)$ and kNN$_{A2}(u)$. In other words, this indicates the degree to which the neighboring classes of $A1$ are similar to those of $A2$. Given semantic representations $A$, \%kNN-overlap($A$, visual exemplars) is used to measure the quality of $A$. We compute the visual exemplars by averaging the ResNet features per unseen class without PCA (cf. Sect.~\ref{sEXEMPredExem}).

Fig.~\ref{fCompareW2VMDSKnn} shows the absolute difference \%kNN-overlap (``hie", visual exemplars) $-$ \%kNN-overlap (``wv-v2", visual exemplars) for $K$ from $1$ to $500$, where the Euclidean distance between classes is used in all cases. For each $K$, a positive number indicates that ``hie" is better at encoding class similarity than ``wv-v2" \emph{within the neighborhoods of size $K$} and this number should be close to zero once $K$ is big enough to cover almost all unseen classes. We observe that these results are correlated well with ZSL performances in Table~\ref{tbZSLSemanticPCIMN}; ``hie" significantly outperforms ``wv-v2" for \emph{semantically close} unseen classes (2-hop) while ``wv-v2" generally outperforms ``hie" for most populated unseen classes. Furthermore, we can attribute the superiority of ``wv-v2" over ``hie" on most populated unseen classes to \emph{distant neighborhoods} --- ``hie" never performs worse when we consider small values of $K$.
}

\subsubsection{Comparison to recently published results with per-sample accuracy}
\label{sExpResAddRecentResults}

\begin{table*}
\centering
\caption{\small Comparison between our approaches using the strongest visual features (ResNet-101) and semantic representations (wv-v2 + hie) and the best published results on \imn. Following those results, we use \textbf{``per-sample"} accuracy (\emph{italicized} text, in \%). For each scenario, the best is in \bst{red} and the second best in \sbst{blue}. (The comparison is subject to a slight variation. See the main text for details.)} \label{tbZSLSemanticPSIMN}
\vskip -0.5em
\begin{tabular}{c|c|c|c|c|c|c|}
Approach & Reported & Visual & Semantic & \multicolumn{2}{c|}{Hierarchy} & All \\ \cline{5-6}
&  by & features & types & 2-hop & 3-hop  & \\ \hline
\zslsyncovo & \tw & &  & \textit{25.70} & \textit{6.28} & \textit{2.84} \\ 
\zslsyncstr & \tw & & wv-v2 & \textit{25.00} & \textit{6.04} & \textit{2.73} \\ 
\zslexemsyncovo & \tw & ResNet-101 & + & \textit{25.74} & \textit{6.41} & \textit{3.01} \\ 
\zslexemsyncstr & \tw & & hie & \textit{\sbst{26.65}} & \textit{6.70} & \textit{3.12} \\ 
\zslexemnn & \tw & &  & \textit{26.42} & \textit{\sbst{6.92}} & \textit{\sbst{3.26}} \\ 
\zslexemnns & \tw & &  & \textit{\bst{27.02}} & \textit{\bst{7.08}} & \textit{\bst{3.35}} \\ \hline
\method{GCNZ} \cite{WangYG18} & \cite{WangYG18} & ResNet-50 & GloVe + hie & \textit{19.8} & \textit{4.1} & \textit{1.8} \\
\method{ADGPM} \cite{KampffmeyerCLWZX18} & \cite{KampffmeyerCLWZX18} & ResNet-50 & GloVe + hie & \textit{24.6} & \textit{-} & \textit{-} \\
\method{ADGPM} \cite{KampffmeyerCLWZX18} & \cite{KampffmeyerCLWZX18} & ResNet-50 fine-tuned & GloVe + hie & \textit{26.6} & \textit{6.3} & \textit{3.0} \\ \hline
\end{tabular}
\end{table*}

Recent studies, \method{GCNZ} \cite{WangYG18} and \method{ADGPM} \cite{KampffmeyerCLWZX18}, obtained very strong ZSL results on \imn. Both methods apply graph convolutional networks~\cite{KipfW17} to predict recognition models given semantic representations, where their ``graph" corresponds to the WordNet hierarchy \cite{Miller95}. They use ResNet-50 visual features and word vectors extracted using GloVe~\cite{PenningtonSM14}.

In Table~\ref{tbZSLSemanticPSIMN}, we report their best results as well as our results with the strongest visual features (ResNet-101) and semantic representations (the combination of ``wv-v2" and ``hie"). The comparison is subject to a slight variation due to differences in visual features (and whether we fine-tune them or not), semantic representations, and the number of unseen classes\footnote{\cite{WangYG18,KampffmeyerCLWZX18} extracted word vectors of class names by averaging the vectors of words in the synset name, enabling all 20,842 unseen classes to have word vectors. The number of 2-hop, 3-hop, and All classes are thus 1,589, 7,860, and 20,842, respectively.}. Nevertheless, in contrast to the conclusion in \cite{WangYG18,KampffmeyerCLWZX18} that these approaches outperform ours (i.e., \zslexemnns with GoogLeNet and ``wv-v1"), our results can be greatly improved by using stronger visual features and semantic representations, especially the latter. This suggests that the conclusion there may result from visual feature/semantic representation differences rather than the methodological superiority of their approaches.

\subsubsection{How far are we from ideal performance?}
\label{sExpResAddSemIdeal}

\begin{table*}
\centering
\caption{\small Performance of zero-shot learning with \emph{ideal} semantic representations on \imn. We compare ``wv-v2 + hie": the combination of ``wv-v2" and ``hie" vs. ``ideal": the average of visual features belong to each class. We use \textbf{``per-class"} accuracy (in \%) and ResNet features in all cases. For each scenario, the best is in \bst{red} and the second best in \sbst{blue}. We note that the numbers with ``wv-v2 + hie" are not exactly the same as in Table~\ref{tbZSLSemanticPCIMN} since we only test on 50\% of the data per unseen class.} \label{tbZSLSemanticIdealIMN}
\vskip -0.5em
\begin{tabular}{c|c|c|c|c|c|c|c|c|c|c|c|c|}
Approach & Semantic & \multicolumn{2}{c|}{Hierarchy} & \multicolumn{3}{c|}{Most populated} & \multicolumn{3}{c|}{Least populated} & All \\ \cline{3-10}
& types & 2-hop & 3-hop & 500 & 1K & 5K & 500 & 1K & 5K & \\ \hline
\zslsyncovo &  & 21.39 & 4.55 & 20.68 & 14.96 & 5.40 & 11.51 & 7.00 & 2.96 & 1.74 \\ 
\zslsyncstr & wv-v2 & 20.34 & 4.29 & 18.20 & 13.25 & 5.07 & 9.92 & 6.60 & 2.64 & 1.63 \\ 
\zslexemsyncovo & + & \sbst{23.83} & \sbst{5.40} & 24.40 & 17.78 & 6.67 & \bst{13.49} & \sbst{8.67} & \bst{3.56} & \sbst{2.11} \\ 
\zslexemsyncstr & hie & \bst{24.68} & \bst{5.62} & \sbst{24.63} & \bst{17.96} & \bst{6.87} & \bst{13.49} & \bst{8.82} & 3.39 & \bst{2.20} \\ 
\zslexemnn &  & 22.51 & 5.24 & 23.91 & 17.37 & 6.47 & \bst{13.49} & 8.40 & 3.15 & 2.07 \\ 
\zslexemnns &  & 22.97 & 5.32 & \bst{24.64} & \sbst{17.79} & \sbst{6.69} & 12.70 & 7.60 & \sbst{3.49} & 2.10 \\ \hline
\zslsyncovo &  & 41.45 & 20.94 & 51.92 & 45.39 & 25.37 & 29.37 & 26.00 & 18.69 & 12.53 \\ 
\zslsyncstr &  & 45.37 & 24.53 & \sbst{54.41} & 48.11 & 28.20 & 30.95 & 28.08 & 22.03 & 15.48 \\ 
\zslexemsyncovo & ideal & 43.67 & 23.29 & 53.44 & 46.80 & 26.98 & 31.75 & 27.97 & 21.57 & 14.68 \\ 
\zslexemsyncstr &  & \bst{45.72} & \sbst{24.80} & \bst{54.89} & \bst{48.34} & \sbst{28.44} & \sbst{32.14} & \bst{28.95} & \sbst{22.16} & \sbst{15.62} \\ 
\zslexemnn &  & 44.80 & 24.31 & 53.48 & 47.26 & 27.84 & 30.16 & 28.55 & 21.76 & 15.40 \\ 
\zslexemnns &  & \sbst{45.56} & \bst{25.04} & \sbst{54.41} & \sbst{48.19} & \bst{28.61} & \bst{32.54} & \sbst{28.75} & \bst{22.65} & \bst{16.09} \\ \hline
\end{tabular}
\end{table*}

It is clear that the success of zero-shot learning relies heavily on how accurate semantic information represents visual similarity among classes. We investigate this in more details. We focus on the strongest semantic representations on \imn and ask what would happen in the \emph{ideal} scenario where predicted visual exemplars of the unseen classes are very accurate. Concretely, for each class, ideal semantic representations can be obtained by averaging visual features of images belonging to that class~\cite{ChaoCGS16}. For seen classes, we use all the data to compute ideal semantic representations and to train the ZSL models. For unseen classes, we randomly reserve 50\% of the data along with their labels for computing ideal semantic representations. The remaining 50\% will be used as test data. Our goal is not to outperfrom the accuracies obtained in Table~\ref{tbZSLSemanticPCIMN} (the numbers are not comparable anyway due to the difference in data splitting). Rather, we aim to see how large the gap is between existing and ideal performances. In Table~\ref{tbZSLSemanticIdealIMN}, we see that the relative gap to the ideal performance is larger as the test set is semantically further from seen classes (from 2-hop to 3-hop to All) and as the test classes become more rare. These observations suggest that developing improved semantic representations (e.g., with more visual information) is a promising future direction for zero-shot learning.


\section{Related Work}
\label{sRelated}

\subsection{Zero-Shot Learning Background}

Morgado and Vasconcelos \cite{MorgadoV17} distinguish ``recognition using independent semantics (RIS)" and ``recognition using semantic embeddings (RULE)." Wang and Chen \cite{WangC17} group ZSL algorithms into ``direct mapping", ``model parameter transfer", and ``common space learning" approaches. Fu et al. \cite{FuXJXSG18} argue that solving zero-shot recognition involves ``embedding models" and ``recognition models in the embedding space," where the embedding models can be further categorized into semantic embedding, Bayesian models, embedding into common spaces, or deep embedding approaches. Xian et al. \cite{XianLSA17} categorize 13 ZSL methods into ``learning linear compatibility", ``learning nonlinear compatibility",  ``learning intermediate attribute classifiers," and ``hybrid models." To facilitate the discussions in our work, we divide zero-shot learning algorithms into the following two themes: two-stage approaches and unified approaches. This criterion is most similar to the one used by Wang and Chen \cite{WangC17}.

\subsubsection{Two-stage approaches}
\label{sRelatedTwoStage}

The theme of two-stage approaches is to identify and learn an intermediate subtask that is then used to infer the final prediction. Two popular subtasks are predicting the embeddings of images in the semantic space, and generating instances of each class given its corresponding semantic representation. It is possible that the selected intermediate subtask is trained jointly with the zero-shot recognition in a unified manner (Sect.~\ref{sRelatedUnified}), but this is not fully investigated in the literature and in some cases may lead to other technical difficulties.

\paragraph{Learning to predict semantic embeddings:}

Given an image, one can project it to the semantic embedding space, and then infer its class label by comparing the predicted semantic embedding to those of unseen classes using a similarity measure. The projection mapping can be trained using standard classification or regression models on image-semantic embedding pairs from the seen classes. The semantic embedding space is usually chosen to be the one where \emph{given} semantic representations live in. As for label inference, there are two popular approaches. One is based on probabilistic models of class labels based on semantic representations \cite{LampertNH09,LampertNH14,NorouziMBSSFCD14}. The other is based on nearest neighbor classifiers on the semantic space \cite{FarhadiEHF09,PalatucciPHM09,SocherGMN13,XuHG15}.

If we assume that semantic representations capture all information one needs to predict the class labels (i.e., they are highly discriminative), then focusing on accurately predicting the semantic embeddings would solve zero-shot learning. In practice, however, this paradigm suffers from the unreliability of semantic predictions.
Several techniques are as a result proposed to alleviate this problem.  
Jayaraman and Grauman \cite{JayaramanG14} propose a random forest based approach to take this unreliability into account. 
Al-Halah and Stiefelhagen \cite{AlHalahS15} construct the hierarchy of concepts underlying the attributes to improve reliability.
Gan et al. \cite{GanYG16} transform visual features to reduce the mismatches between attributes in different categories, thus enhancing reliability.

\paragraph{Learning to generate instances of each class:}

Recent advances in conditional generative models (e.g., \cite{ReedAYLSL16,YanYSL16}) lead to interest in exploiting them for generating labeled data from corresponding semantic representations. Once those examples are generated, one can employ any supervised learning technique to learn classifiers \cite{KumarAMR18,ZhuELPE18,XianLSA18,BucherHJ18}. Note that all these methods focus on directly generating features rather than image pixels. 

\subsubsection{Unified approaches}
\label{sRelatedUnified}

The other type of ZSL approaches focuses on the task of zero-shot classification directly. There are two main sub-themes, where the difference lies in whether the emphasis is on learning common space (or compatibility) or on learning model parameters, but the distinction between the two is thin.  

\paragraph{Common space or compatibility learning:}
This approach learns a common representation to which visual features and semantic representations are projected with the objective of maximizing the compatibility score of projected instances in this space. The difference among methods of this category lies in their choices of common spaces or compatibility functions.
The linear or bilinear compatibility functions are extensively used \cite{AkataPHS13,AkataRWLS15,FromeCSBDRM13,Bernardino15,LiGS15}. Some propose to use canonical correlation analysis (CCA)~\cite{FuHXFG14,FuHXG15,Lu16}. Nonlinear methods are scarce but have also been explored such as dictionary learning and sparse coding~\cite{KodirovXFG15,ZhangS16}.

\paragraph{Model parameter learning:}
One can also build the classifiers of unseen classes by relating them to seen ones via similarities computed from semantic representations~\cite{RohrbachSSGS10,RohrbachSS11,ElhoseinySE13,MensinkGS14,GavvesMTST15,LeiSFS15,YangH15,ZhangS15}. For example, Mensink et al. and Gan et al. \cite{MensinkGS14,GanLYZH2015} propose to construct classifiers of unseen objects by combining classifiers of seen objects, where the combining coefficients are determined based on semantic relatedness.

\subsection{Putting Our Methods in Context}
\label{sApproachComparison}

Both \method{SynC}, \method{EXEM (1NN)} and \method{EXEM (1NNs)} fall into model parameter learning approaches but the details in how they construct classifiers/exemplars are different. \method{EXEM (1NN)} and \method{EXEM (1NNs)} can also be viewed as learning to generate one instance for each class --- without modeling variations explicitly. \method{EXEM (\emph{ZSL method})} falls into the approach of learning to predict semantic embeddings but we show that the (projected) space of visual features is an extremely effective semantic embedding space.

We provide detailed discussions of each of our methods with respect to their most relevant work below.

\paragraph{Detailed discussions of \method{SynC}:}
\method{COSTA} \cite{MensinkGS14} combines \textbf{pre-trained} classifiers of seen classes to construct new classifiers. To estimate the semantic embedding of a test image, \method{ConSE} \cite{NorouziMBSSFCD14} uses the decision values of pre-trained classifiers of seen objects to weightedly average the corresponding semantic representations. Neither of them has the notion of base classifiers as in \method{SynC}, which we introduce to construct classifiers but nothing else. We thus expect using bases to be more effective in transferring knowledge between seen and unseen classes than overloading the pre-trained and fixed classifiers of the seen classes for dual duties. We note that \method{ALE} \cite{AkataPHS13} and \method{SJE} \cite{AkataRWLS15} can be considered as special cases of \method{SynC}. In \method{ALE} and \method{SJE}, each attribute corresponds to a base and each ``real'' object classifier is represented as a linear combination of those bases, where the weights are the real objects' ``descriptions'' in the form of attributes. This modeling choice is inflexible as the number of bases is fundamentally constrained by the number of attributes. Moreover, the model is strictly a subset of \method{SynC}\footnote{For interested readers, if we set the number of attributes as the number of phantom classes (each $\vb_r$ is the one-hot representation of an attribute), and use the Gaussian kernel with an isotropically diagonal covariance matrix in Eq.~(\ref{eSynCDist}) with properly set bandwidths (either very small or very large) for each attribute, we will recover the formulation in \cite{AkataPHS13,AkataRWLS15} when the bandwidths tend to zero or infinity.}. \method{SSE} and \method{JLSE} \cite{ZhangS16,ZhangS15} propose similar ideas of aligning the visual and semantic spaces but take different approaches from ours.

Our convex combination of base classifiers for synthesizing real classifiers can also be motivated from multi-task learning with shared representations~\cite{ArgyriouEP08}. While labeled examples of each task are required in~\cite{ArgyriouEP08}, our method has no access to \emph{data} of the unseen classes.

\paragraph{Detailed discussions of \method{EXEM}:}

\method{DeViSE} \cite{FromeCSBDRM13} and \method{ConSE} \cite{NorouziMBSSFCD14} predict an image's semantic embedding from its visual features and compare to unseen classes' semantic representations. We perform an ``inverse prediction'': given an unseen class's semantic representation, we predict the visual feature exemplar of that class.

One appealing property of \method{EXEM} is its \emph{scalability}: we learn and predict at the exemplar (class) level so the runtime and memory footprint of our approach depend only on the number of seen classes rather than the number of training data points. This is much more efficient than other ZSL algorithms that learn at the level of each individual training instance~\cite{FarhadiEHF09,LampertNH09,PalatucciPHM09,AkataPHS13,YuCFSC13,FromeCSBDRM13,SocherGMN13,NorouziMBSSFCD14,JayaramanG14,MensinkGS14,AkataRWLS15,Bernardino15,ZhangS15,ZhangS16,Lu16}.

Several methods propose to learn visual exemplars by preserving structures obtained in the semantic space, where \emph{exemplars} are used loosely here and do not necessarily mean class-specific feature averages. Examples of such methods include \method{SynC}, \method{BiDiLEL} \cite{WangC17} and \method{UVDS} \cite{LongLSSDH17}. However, \method{EXEM} \emph{predicts} them with a regressor such that they may or may not strictly follow the structure in the semantic space, and thus they are more flexible and could even better reflect similarities between classes in the visual feature space.

Similar in spirit to our work, Mensink et al. propose using nearest class mean classifiers for ZSL \cite{MensinkVPC13}. The Mahalanobis metric learning in this work could be thought of as learning a linear transformation of semantic representations (their ``zero-shot prior" means, which are in the visual feature space). Our approach learns a highly non-linear transformation. Moreover, our \method{EXEM (1NNs)} (cf. Sect.~\ref{sExpSetup}) learns a (simpler, i.e., diagonal) metric over the learned exemplars. Finally, the main focus of \cite{MensinkVPC13} is on \emph{incremental}, not zero-shot, learning settings (see also \cite{RistinGGB16,RebuffiKSL17}).

\method{DEm} \cite{ZhangXG17} uses a deep feature space as the semantic embedding space for ZSL. Though similar to \method{EXEM}, this approach does not compute the average of visual features (exemplars) but train neural networks to predict \emph{all} visual features from their semantic representations. Their model learning takes significantly longer time than ours.

There has been a recent surge of interests in applying deep learning models to generate images (see, e.g., \cite{MansimovPBS16,ReedAYLSL16,YanYSL16}). Most of these methods are based on probabilistic models (in order to incorporate the statistics of natural images). Unlike them, \method{EXEM} deterministically predicts visual features. Note that generating features directly is likely easier and more effective than generating realistic images first and then extracting visual features. Recently, researchers became interested in generating visual features of unseen classes using conditional generative models such as variants of generative adversarial networks (GANs) \cite{XianLSA18,ZhuELPE18} and variational autoencoders (VAEs) \cite{KumarAMR18} for ZSL.


\section{Conclusion}
\label{sDiscuss}

\method{SynC} is a concrete realization of a novel idea that casts zero-shot learning as learning manifold embeddings from graphs composed of object classes. In this classifier synthesis framework, we show how to parameterize the graphs with the locations of the phantom classes, and how to derive recognition models as convex combinations of base classifiers. \method{EXEM} is a two-stage zero-shot learning method that incorporates the task of visual exemplar predictions to automatically \emph{denoise} semantic representations. We show that this task can be done efficiently and effectively using kernelized support vector regression and PCA. We use \method{EXEM} to improve upon \method{SynC} and several other ZSL methods as well as to construct nearest-neighbor-style classifiers.

Our results demonstrate the effectiveness of our proposed approaches on both conventional and generalized zero-shot learning settings in diverse sets of scenarios. We also show that semantic representations significantly contribute to the performance of ZSL methods on the large-scale ImageNet benchmark --- we derive better semantic representations, achieve the state-of-the-art performance, and see a large gap between the effectiveness of existing and ``ideal" semantic representations. Our study also raises an important issue regarding the evaluation metrics in the generalized ZSL setting, suggesting that the AUSUC or the calibrated harmonic mean should be used instead of the uncalibrated harmonic mean. We believe that such insights will greatly benefit the community.

\vskip 1em
\noindent\textbf{Acknowledgments} This work is partially supported by USC Graduate Fellowships, NSF IIS-1065243, 1451412, 1513966/1632803/1833137, 1208500, CCF-1139148, a Google Research Award, an Alfred P. Sloan Research Fellowship, gifts from Facebook and Netflix, and ARO\# W911NF-12-1-0241 and W911NF-15-1-0484.

\begin{appendices}


\section{Details on How to Obtain Word Vectors on \imn}
\label{sSuppSkip}
We use the word2vec package\footnote{\url{https://code.google.com/p/word2vec/}}. We preprocess the input corpus with the word2phrase function so that we can directly obtain word vectors for both single-word and multiple-word terms, including those terms in the ImageNet \emph{synsets}; each class of \textbf{ImageNet} is a \emph{synset}: a set of synonymous terms, where each term is a word or a phrase. We impose no restriction on the vocabulary size. Following \cite{FromeCSBDRM13}, we use a window size of 20, apply the hierarchical softmax for predicting adjacent terms, and train the model for a single epoch. As one class may correspond to multiple word vectors by the nature of synsets, we simply average them to form a single word vector for each class.

\section{Hyper-parameter Tuning} 
\label{sSuppHyperTuning}
 
\begin{figure}[t]
\centering
\vspace{-.1in}
\includegraphics[width=0.49\textwidth]{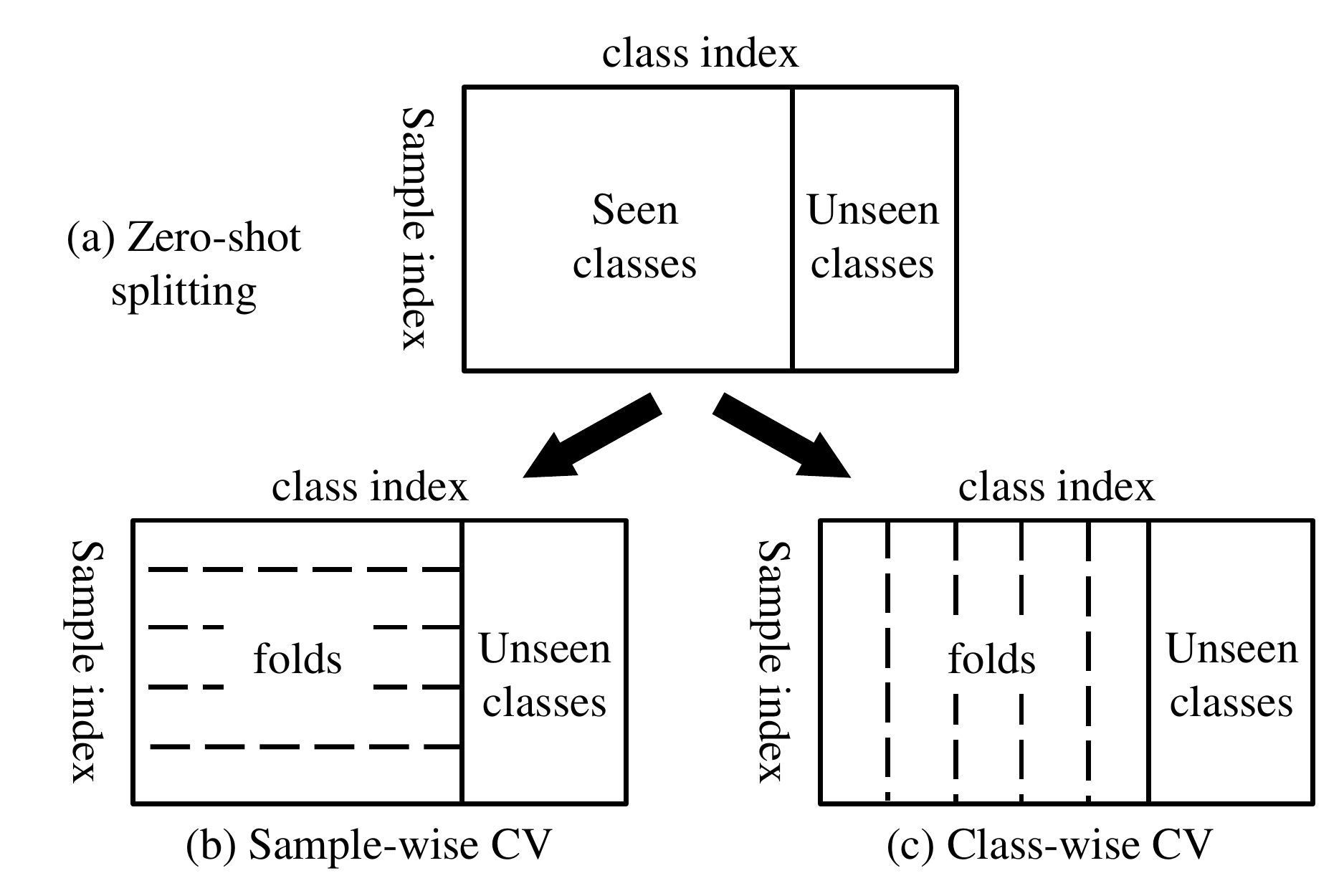}
\vspace{-.1in}
\caption{Data splitting for different cross-validation (CV) strategies: (a) the seen-unseen class splitting for (conventional) zero-shot learning, (b) the sample-wise CV, (c) the class-wise CV.}
\label{fCV}
\end{figure}

\subsection{For Conventional Zero-shot Learning}
\label{sCV_CZSL}
The standard approach for cross-validation (CV) in a classification task splits training data into several folds such that they share the same set of class labels. This strategy is less sensible in zero-shot learning as it does not imitate what actually happens at the test stage. We thus adopt the strategy in \cite{ElhoseinySE13,AkataRWLS15,Bernardino15,ZhangS15,Bernardino15}. In this scheme, we split training data into several folds such that the class labels of these folds are disjoint. We then hold out data from one fold as \emph{pseudo-unseen} classes, train our models on the remaining folds (which belong to the remaining classes), and tune hyper-parameters based on a certain performance metric on the held-out fold. For clarity, we denote the standard CV as \emph{sample}-wise CV and the zero-shot CV scheme as \emph{class}-wise CV. Fig. \ref{fCV} illustrates the two scenarios.

We use this strategy to tune hyper-parameters in both our approaches (\method{SynC} and \method{EXEM}) and the baselines.
In \method{SynC}, the main hyper-parameters are the regularization parameter $\lambda$ in Eq. (\ref{eSynCObj}) and the scaling parameter $\sigma$ in Eq. (\ref{eSynCDist}). When learning semantic representations (Eq. (\ref{eSynCSemObj})), we also tune $\eta$ and $\gamma$. To reduce the search space during CV, we first fix $\vb_r = \va _r$ for $r = 1,\ldots,\cR$ and tune $\lambda, \sigma$. Then we fix $\lambda$ and $\sigma$ and tune $\eta$ and $\gamma$. The metric is the classification accuracy.

In \method{EXEM}, we tune \textbf{(a)} projected dimensionality $\cst{d}$ for PCA and \textbf{(b)} $\lambda$, $\nu$, and the RBF-kernel bandwidth in SVR\footnote{For GoogLeNet features, we follow~\cite{ChangpinyoCS17} to set $\lambda=1$ and $\cst{d}=500$ for all experiments.}. Since \method{EXEM} is a two-stage approach, we consider the following two performance metrics. The first one minimizes the distance between the predicted exemplars and the ground-truth (average of the hold-out data of each class after the PCA projection) in $\R^{\cst{d}}$. We use the Euclidean distance in this case. We term this measure ``CV-distance." This approach does not assume the downstream task at training and aims to measure the quality of predicted exemplars by its \emph{faithfulness}. The other approach ``CV-accuracy" maximizes the per-class classification accuracy on the hold-out fold. This measure can easily be obtained for \method{EXEM (1NN)} and \method{EXEM (1NNs)}, which use simple decision rules that have no further hyper-parameters to tune. Empirically, we found that CV-accuracy generally leads to slightly better performance. The results reported in the main text for these two approaches are thus based on this measure. On the other hand, \method{EXEM (\emph{ZSL method})} (where \emph{ZSL method} = \method{SynC}, \method{ConSE}, \method{ESZSL}) requires further hyper-parameter tuning. For computational purposes, we use CV-distance\footnote{For CV-distance, we set $\cst{d}=500$ for all experiments. This is because the smaller $\cst{d}$ is, the smaller the distance is.} for tuning hyper-parameters of the regressors, followed by the hyper-parameter tuning for \emph{ZSL methods} using the predicted exemplars. As \method{SynC} and \method{ConSE} construct their classifiers based on the distance values between class semantic representations, we do not expect a significant performance drop in this case.

\subsection{For Generalized Zero-shot Learning}
To perform \emph{class}-wise CV in the generalized zero-shot learning (GZSL) setting, we further separate each fold into two splits, each with either 80\% or 20\% of data. We then hold out one fold, train models on the $80\%$ splits of the remaining folds, and tune hyper-parameters based on a certain performance metric on (i) the $80\%$ split of the hold-out fold and (ii) the $20\%$ splits of the training (i.e., remaining) folds. In this way we can mimic the GZSL setting in hyper-parameter tuning. Specifically, for metrics with calibration (cf. Table~\ref{tbGZSLHarmonic}), we first compute AUSUC using (i) and (ii) to tune the hyper-parameters mentioned in Sect.~\ref{sCV_CZSL}, and select the calibration factor $\gamma$ that maximizes the harmonic mean. For the uncalibrated harmonic mean, we follow~\cite{XianLSA17} to tune hyper-parameters in the same way as in the conventional ZSL setting.


\section{Experimental Results on \imn with Previous Experimental Setups}
\label{sSuppExpIMN}

The first ZSL work on \imn and much of its follow-up considers only 2-hop, 3-hop, and All test sets and other evaluation metrics. We include our results here in Table~\ref{tbZSLExpandedIMN}, \ref{tbZslMainImnMds}, and \ref{tbZSLSemanticPSFullIMN} to aid comparison with such work. As mentioned in Sect.~\ref{sExpSetupZSL}, we also consider Flat hit@K (F@K) and Hierarchical precision@K (HP@K). F@K is defined as the percentage of test images for which the model returns the true label in its top K predictions. HP@K is defined as the percentage of overlapping (i.e., precision) between the model's top K predictions
and the ground-truth list. For each class, the ground-truth list of its K closest categories is generated based on the ImageNet
hierarchy. Note that F@1 is the \emph{per-sample} multi-way classification accuracy.

When computing Hierarchical precision@K (HP@K), we use the algorithm in the Appendix of \cite{FromeCSBDRM13} to compute the ground-truth list, a set of at least $K$ classes that are considered to be correct. This set is called $hCorrectSet$ and it is computed for each $K$ and class $c$. See Algorithm~\ref{aHCorrect} for more details. The main idea is to expand the radius around the true class $c$ until the set has at least $K$ classes.
\vspace{.1in}
\begin{algorithm}
\caption{Algorithm for computing $hCorrectSet$ for HP@K \cite{FromeCSBDRM13}}
\begin{algorithmic}[1]
	\State Input: $K$, class $c$, ImageNet hierarchy
	\State $hCorrectSet \leftarrow \emptyset$
	\State $R \leftarrow 0$
	\While {NumberElements($hCorrectSet) < K$}
		\State $radiusSet \leftarrow$ all nodes in the hierarchy which are $R$ hops from $c$
		\State $validRadiusSet \leftarrow$ ValidLabelNodes($radiusSet$)
		\State $hCorrectSet \leftarrow$ $hCorrectSet \cup validRadiusSet$
		\State $R \leftarrow R + 1$
	\EndWhile
	\vspace{0.1in}\State \Return $hCorrectSet$
\end{algorithmic}
\label{aHCorrect}
\end{algorithm}

Note that $validRadiusSet$ depends on which classes are in the label space to be predicted (i.e., depending on whether we consider 2-hop, 3-hop, or All. We obtain the label sets for 2-hop and 3-hop from the authors of \cite{FromeCSBDRM13,NorouziMBSSFCD14}. We implement Algorithm~\ref{aHCorrect} to derive $hCorrectSet$ ourselves.

\begin{table*}
\centering
\tabcolsep 5pt
\caption{Expanded results of Table~\ref{tbZSLMainIMN}. The metric is \textbf{``per-sample"} accuracy (\emph{italicized} text) for F@K to aid comparison with previous published results. Comparison of ZSL methods on \imn using word vectors of the class names as semantic representations. For both types of metrics (in \%), the higher the better. The best is in \bst{red}. AlexNet is by \cite{AlexNet}. The number of actual unseen classes are given in parentheses.}
\label{tbZSLExpandedIMN}
\begin{tabular}{c|c|c|c|ccccc|cccc}
Test & Approach & Reported & Visual & \multicolumn{5}{c|}{F@K} & \multicolumn{4}{c}{HP@K}\\ \cline{5-13}
data  & K= & by & features & 1 & 2 & 5 & 10 & 20 & 2 & 5 & 10 & 20 \\ \hline
2-hop
& \zsldevise & \cite{FromeCSBDRM13}& AlexNet & \textit{6.0} & \textit{10.1} & \textit{18.1} & \textit{26.4} & \textit{36.4} & 15.2 & 19.2 & 21.7 & 23.3 \\
(1,549)
& \zslconse & \cite{NorouziMBSSFCD14}& AlexNet  & \textit{9.4} & \textit{15.1} & \textit{24.7} & \textit{32.7} & \textit{41.8} & 21.4 & 24.7 & 26.9 & 28.4 \\  \hline
& \zslconse & \sync& GoogLeNet & \textit{8.3} & \textit{12.9} & \textit{21.8} & \textit{30.9} & \textit{41.7} & 21.5 & 23.8 & 27.5 & 31.3 \\ \cline{2-13}
& \zslsyncovo & \sync& GoogLeNet & \textit{10.5} & \textit{16.7} & \textit{28.6} & \textit{40.1} & \textit{52.0} & 25.1 & 27.7 & 30.3 & 32.1 \\
2-hop
& \zslsyncstr & \sync& GoogLeNet & \textit{9.8} & \textit{15.3} & \textit{25.8} & \textit{35.8} & \textit{46.5} & 23.8 & 25.8 & 28.2 & 29.6 \\
(1,509)
& \zslexemsyncovo & \exem& GoogLeNet & \textit{11.8} & \textit{18.9} & \textit{31.8} & \textit{43.2} & \textit{54.8} & 25.6 & 28.1 & 30.2 & 31.6 \\
& \zslexemnn & \exem& GoogLeNet	& \textit{11.7} & \textit{18.3} & \textit{30.9} & \textit{42.7} & \textit{54.8} & 25.9 & 28.5 & \bst{31.2} & \bst{33.3} \\
& \zslexemnns & \exem& GoogLeNet & \textit{\bst{12.5}} & \textit{\bst{19.5}} & \textit{\bst{32.3}} & \textit{\bst{43.7}} & \textit{\bst{55.2}} & \bst{26.9} & \bst{29.1} & 31.1 & 32.0 \\ \hline \hline 
3-hop
& \zsldevise & \cite{FromeCSBDRM13}& AlexNet  & \textit{1.7} & \textit{2.9} & \textit{5.3} & \textit{8.2} & \textit{12.5} & 3.7 & 19.1 & 21.4 & 23.6 \\
(7,860)
& \zslconse & \cite{NorouziMBSSFCD14}& AlexNet  & \textit{2.7} & \textit{4.4} & \textit{7.8} & \textit{11.5} & \textit{16.1} & 5.3 & 20.2 & 22.4 & 24.7 \\ \hline 
& \zslconse & \sync& GoogLeNet & \textit{2.6} & \textit{4.1} & \textit{7.3} & \textit{11.1} & \textit{16.4} & 6.7 & 21.4 & 23.8 & 26.3 \\ \cline{2-13}
& \zslsyncovo & \sync& GoogLeNet & \textit{2.9} & \textit{4.9} & \textit{9.2} & \textit{14.2} & \textit{20.9} & 7.4 & 23.7 & 26.4 & 28.6 \\
3-hop
& \zslsyncstr & \sync& GoogLeNet & \textit{2.9} & \textit{4.7} & \textit{8.7} & \textit{13.0} & \textit{18.6} & 8.0 & 22.8 & 25.0 & 26.7 \\
(7,678)
& \zslexemsyncovo & \exem& GoogLeNet & \textit{3.4} & \textit{5.6} & \textit{10.3} & \textit{15.7} & \textit{22.8} & 7.5 & 24.7 & 27.3 & 29.5 \\ 
& \zslexemnn & \exem& GoogLeNet & \textit{3.4} & \textit{5.7} & \textit{10.3} & \textit{15.6} & \textit{22.7} & 8.1 & \bst{25.3} & \bst{27.8} & \bst{30.1}	 \\
& \zslexemnns & \exem& GoogLeNet & \textit{\bst{3.6}} & \textit{\bst{5.9}} & \textit{\bst{10.7}} & \textit{\bst{16.1}} & \textit{\bst{23.1}} & \bst{8.2} & 25.2 & 27.7 & 29.9 \\
\hline \hline 
All
& \zsldevise & \cite{FromeCSBDRM13}& AlexNet  & \textit{0.8} & \textit{1.4} & \textit{2.5} & \textit{3.9} & \textit{6.0} & 1.7 & 7.2 & 8.5 & 9.6 \\
(20,842)
& \zslconse & \cite{NorouziMBSSFCD14}& AlexNet  & \textit{1.4} & \textit{2.2} & \textit{3.9} & \textit{5.8} & \textit{8.3} & 2.5 & 7.8 & 9.2 & 10.4 \\ \hline
& \zslconse & \sync& GoogLeNet & \textit{1.3} & \textit{2.1} & \textit{3.8} & \textit{5.8} & \textit{8.7} & 3.2 & 9.2 & 10.7 & 12.0 \\ \cline{2-13}
& \zslsyncovo & \sync& GoogLeNet & \textit{1.4} & \textit{2.4} & \textit{4.5} & \textit{7.1} & \textit{10.9} & 3.1 & 9.0 & 10.9 & 12.5 \\ 
All
& \zslsyncstr & \sync& GoogLeNet & \textit{1.5} & \textit{2.4} & \textit{4.4} & \textit{6.7} & \textit{10.0} & 3.6 & 9.6 & 11.0 & 12.2 \\
(20,345)
& \zslexemsyncovo & \exem& GoogLeNet & \textit{1.6} & \textit{2.7} & \textit{5.0} & \textit{7.8} & \textit{11.8} & 3.2 & 9.3 & 11.0 & 12.5 \\ 
& \zslexemnn & \exem& GoogLeNet & \textit{1.7} & \textit{2.8} & \textit{5.2} & \textit{8.1} & \textit{12.1} & \bst{3.7} & \bst{10.4} & \bst{12.1} & \bst{13.5} \\
& \zslexemnns & \exem& GoogLeNet  & \textit{\bst{1.8}} & \textit{\bst{2.9}} & \textit{\bst{5.3}} & \textit{\bst{8.2}} & \textit{\bst{12.2}} & 3.6 & 10.2 & 11.8 & 13.2\\
\hline
\end{tabular}
\end{table*}

\begin{table*}
\centering
\caption{Expanded results of the third section of Table \ref{tbZSLSemanticPCIMN}. The metric is \textbf{``per-sample"} accuracy (\emph{italicized} text) to aid comparison with previous published results. Comparison of ZSL methods using ``hie", the WordNet-hierarchy embeddings by multidimensional scaling \cite{Lu16}, as semantic representations. The higher, the better (in \%). The best is in \bst{red}.} 
\label{tbZslMainImnMds}
\begin{tabular}{c|c|c|c|ccccc}
Test & Approach & Reported & Visual & \multicolumn{5}{c}{F@K} \\ \cline{5-9}
data & K= & by & features & 1 & 2 &5 & 10 & 20 \\ \hline 
& \zslcca & \cite{Lu16}& GoogLeNet & \textit{1.8} & \textit{3.0} & \textit{5.2} & \textit{7.3} & \textit{9.7}  \\ \cline{2-9}
All
& \zslsyncovo & \sync& GoogLeNet & \textit{\bst{2.0}} & \textit{3.4} & \textit{6.0} & \textit{8.8} & \textit{12.5}  \\ 
(20,842)
& \zslexemsyncovo & \exem& GoogLeNet & \textit{\bst{2.0}} & \textit{3.3} & \textit{6.1} & \textit{9.0} & \textit{12.9} \\
& \zslexemnn	&\exem & GoogLeNet & \textit{\bst{2.0}} & \textit{\bst{3.4}} & \textit{\bst{6.3}} & \textit{\bst{9.2}} & \textit{13.1}  \\
& \zslexemnns &\exem & GoogLeNet & \textit{\bst{2.0}} & \textit{\bst{3.4}} & \textit{6.2} & \textit{\bst{9.2}} & \textit{\bst{13.2}} \\
\hline
\end{tabular}
\end{table*}

\begin{table*}
\centering
\caption{Expanded results of Table~\ref{tbZSLSemanticPCIMN} with \textbf{``per-sample"} accuracy (\emph{italicized} text) used to differentiate this accuracy from the ``per-class" one. Except for the metric, the setting in Table~\ref{tbZSLSemanticPCIMN} is still employed to compare different types of semantic representations on \imn. We consider (i) ``wv-v1": word vectors of the class names trained for one epoch used in Table~\ref{tbZSLMainIMN}, (ii) ``wv-v2": word vectors of the class names trained for 10 epochs, (iii) ``hie": the WordNet-hierarchy embeddings obtained using multidimensional scaling \cite{Lu16}, (iv) ``wv-v1 + hie": the combination of (i) and (iii), (v) ``wv-v2 + hie": the combination of (ii) and (iii). We use ResNet features in all cases. For each scenario, the best is in \bst{red} and the second best in \sbst{blue}.} \label{tbZSLSemanticPSFullIMN}
\vskip -0.5em
\begin{tabular}{c|c|c|c|c|c|c|c|c|c|c|c|c|}
Approach&  Semantic & \multicolumn{2}{c|}{Hierarchy} & \multicolumn{3}{c|}{Most populated} & \multicolumn{3}{c|}{Least populated} & All \\ \cline{3-10}
& types & 2-hop & 3-hop & 500 & 1K & 5K & 500 & 1K & 5K & \\ \hline
\zslsyncovo &  & \textit{12.42} & \textit{3.35} & \textit{16.23} & \textit{11.18} & \textit{3.55} & \textit{5.68} & \textit{4.05} & \textit{1.42} & \textit{1.64} \\ 
\zslsyncstr &  & \textit{11.47} & \textit{3.31} & \textit{14.78} & \textit{10.38} & \textit{3.50} & \textit{4.69} & \textit{3.58} & \textit{1.31} & \textit{1.67} \\ 
\zslexemsyncovo & wv-v1 & \textit{14.01} & \textit{4.13} & \textit{\bst{19.10}} & \textit{\bst{13.42}} & \textit{4.58} & \textit{\sbst{6.42}} & \textit{4.73} & \textit{\bst{1.75}} & \textit{2.03} \\ 
\zslexemsyncstr &  & \textit{\sbst{14.26}} & \textit{\sbst{4.18}} & \textit{18.82} & \textit{\sbst{13.31}} & \textit{\sbst{4.64}} & \textit{6.17} & \textit{4.85} & \textit{\sbst{1.74}} & \textit{2.06} \\ 
\zslexemnn &  & \textit{13.78} & \textit{4.17} & \textit{18.06} & \textit{12.74} & \textit{4.51} & \textit{\bst{6.54}} & \textit{\bst{5.04}} & \textit{1.68} & \textit{\sbst{2.15}} \\ 
\zslexemnns &  & \textit{\bst{14.90}} & \textit{\bst{4.36}} & \textit{\sbst{18.94}} & \textit{13.22} & \textit{\bst{4.65}} & \textit{6.30} & \textit{\sbst{4.88}} & \textit{1.72} & \textit{\bst{2.18}} \\ \hline
\zslsyncovo &  & \textit{15.37} & \textit{4.30} & \textit{18.84} & \textit{13.06} & \textit{4.46} & \textit{5.56} & \textit{4.36} & \textit{1.93} & \textit{2.06} \\ 
\zslsyncstr &  & \textit{14.76} & \textit{4.25} & \textit{17.59} & \textit{12.58} & \textit{4.33} & \textit{6.54} & \textit{4.39} & \textit{1.78} & \textit{2.09} \\ 
\zslexemsyncovo & wv-v2 & \textit{16.76} & \textit{4.87} & \textit{\bst{21.43}} & \textit{\bst{14.89}} & \textit{5.23} & \textit{6.42} & \textit{\sbst{5.53}} & \textit{\bst{2.38}} & \textit{2.37} \\ 
\zslexemsyncstr &  & \textit{\sbst{17.05}} & \textit{\sbst{4.94}} & \textit{\sbst{21.27}} & \textit{14.86} & \textit{\bst{5.35}} & \textit{6.42} & \textit{4.79} & \textit{2.31} & \textit{2.39} \\ 
\zslexemnn &  & \textit{16.52} & \textit{\sbst{4.94}} & \textit{20.63} & \textit{14.45} & \textit{5.21} & \textit{\bst{7.28}} & \textit{5.34} & \textit{2.15} & \textit{\sbst{2.49}} \\ 
\zslexemnns &  & \textit{\bst{17.44}} & \textit{\bst{5.08}} & \textit{21.12} & \textit{\bst{14.89}} & \textit{\bst{5.35}} & \textit{\sbst{6.79}} & \textit{\bst{5.56}} & \textit{\sbst{2.34}} & \textit{\bst{2.50}} \\ \hline
\zslsyncovo &  & \textit{23.25} & \textit{5.29} & \textit{15.46} & \textit{11.66} & \textit{4.86} & \textit{9.88} & \textit{6.24} & \textit{2.32} & \textit{2.37} \\ 
\zslsyncstr &  & \textit{22.86} & \textit{5.20} & \textit{14.67} & \textit{11.27} & \textit{4.78} & \textit{9.14} & \textit{5.90} & \textit{2.29} & \textit{2.33} \\ 
\zslexemsyncovo & hie & \textit{23.78} & \textit{5.40} & \textit{\sbst{16.96}} & \textit{\sbst{12.82}} & \textit{5.20} & \textit{\bst{10.62}} & \textit{\bst{6.98}} & \textit{\sbst{2.74}} & \textit{\sbst{2.42}} \\ 
\zslexemsyncstr &  & \textit{\bst{24.46}} & \textit{\bst{5.54}} & \textit{\bst{17.10}} & \textit{\bst{13.10}} & \textit{\bst{5.35}} & \textit{\sbst{10.37}} & \textit{\bst{6.98}} & \textit{\bst{2.79}} & \textit{\bst{2.48}} \\ 
\zslexemnn &  & \textit{23.36} & \textit{5.25} & \textit{16.67} & \textit{12.50} & \textit{5.06} & \textit{10.12} & \textit{6.39} & \textit{2.49} & \textit{2.35} \\ 
\zslexemnns &  & \textit{\sbst{24.17}} & \textit{\sbst{5.41}} & \textit{16.91} & \textit{12.69} & \textit{\sbst{5.25}} & \textit{9.51} & \textit{6.64} & \textit{2.69} & \textit{\sbst{2.42}} \\ \hline
\zslsyncovo &  & \textit{25.34} & \textit{6.21} & \textit{20.02} & \textit{14.50} & \textit{5.57} & \textit{8.89} & \textit{6.18} & \textit{2.83} & \textit{2.86} \\ 
\zslsyncstr & wv-v1 & \textit{25.69} & \textit{6.14} & \textit{18.18} & \textit{13.51} & \textit{5.37} & \textit{9.38} & \textit{6.43} & \textit{2.70} & \textit{2.77} \\ 
\zslexemsyncovo & + & \textit{25.39} & \textit{6.32} & \textit{23.16} & \textit{17.01} & \textit{6.43} & \textit{\bst{11.85}} & \textit{\bst{7.85}} & \textit{\sbst{3.29}} & \textit{2.97} \\ 
\zslexemsyncstr & hie & \textit{\sbst{26.37}} & \textit{\sbst{6.53}} & \textit{\bst{23.64}} & \textit{\bst{17.27}} & \textit{\bst{6.69}} & \textit{\sbst{11.11}} & \textit{7.66} & \textit{3.27} & \textit{3.04} \\ 
\zslexemnn &  & \textit{25.46} & \textit{6.52} & \textit{22.36} & \textit{16.38} & \textit{6.14} & \textit{10.49} & \textit{7.29} & \textit{3.02} & \textit{\sbst{3.05}} \\ 
\zslexemnns &  & \textit{\bst{27.44}} & \textit{\bst{6.94}} & \textit{\sbst{23.55}} & \textit{\sbst{17.13}} & \textit{\sbst{6.60}} & \textit{10.74} & \textit{\sbst{7.82}} & \textit{\bst{3.36}} & \textit{\bst{3.24}} \\ \hline
\zslsyncovo &  & \textit{25.70} & \textit{6.28} & \textit{20.69} & \textit{15.07} & \textit{5.55} & \textit{11.36} & \textit{6.98} & \textit{2.92} & \textit{2.84} \\ 
\zslsyncstr & wv-v2 & \textit{25.00} & \textit{6.04} & \textit{18.22} & \textit{13.37} & \textit{5.23} & \textit{9.38} & \textit{6.15} & \textit{2.62} & \textit{2.73} \\ 
\zslexemsyncovo & + & \textit{25.74} & \textit{6.41} & \textit{24.32} & \textit{17.82} & \textit{6.78} & \textit{12.22} & \textit{\bst{7.97}} & \textit{\sbst{3.51}} & \textit{3.01} \\ 
\zslexemsyncstr & hie & \textit{\sbst{26.65}} & \textit{6.70} & \textit{\sbst{24.50}} & \textit{\bst{18.02}} & \textit{\bst{6.97}} & \textit{\sbst{12.96}} & \textit{\sbst{7.88}} & \textit{3.42} & \textit{3.12} \\ 
\zslexemnn &  & \textit{26.42} & \textit{\sbst{6.92}} & \textit{23.82} & \textit{17.39} & \textit{6.59} & \textit{\bst{13.21}} & \textit{7.35} & \textit{3.24} & \textit{\sbst{3.26}} \\ 
\zslexemnns &  & \textit{\bst{27.02}} & \textit{\bst{7.08}} & \textit{\bst{24.53}} & \textit{\sbst{17.83}} & \textit{\sbst{6.82}} & \textit{12.47} & \textit{7.54} & \textit{\bst{3.55}} & \textit{\bst{3.35}} \\ \hline
\end{tabular}
\end{table*}


\section{Analysis on SynC}
\label{sSuppExpResAddSynC}

In this section, we focus on \method{SynC$^\textrm{o-vs-o}$} together with GoogLeNet features and the standard split (SS). We look at the effect of modifying the regularization term, learning base semantic representations, and varying the number of base classes and their correlations.

\begin{table}
	\centering
	\caption{Comparison between regularization with $\vw_c$ and $\vv_c$ on \method{SynC$^\textrm{o-vs-o}$}.}
	\label{tSynCRegularization}
	\begin{tabular}{c|c|c|c}
		Datasets & Visual features & $\twonorm{\vw_c}$ & $\twonorm{\vv_r}$\\ \hline
		\awa & GoogLeNet & 69.7\% & 71.7\% \\ 
		\cub & GoogLeNet & 53.4\% & 56.4\%\\ 
		\sun & GoogLeNet & 62.8\% & 67.5\%\\ \hline
	\end{tabular}
\end{table}

\subsection{Different Forms of Regularization} In Eq.~(\ref{eSynCObj}) and (\ref{eSynCSemObj}), $\twonorm{\vw_c}$ is the regularization term. Here we consider modifying that term to $\twonorm{\vv_r}$ --- regularizing the bases directly. Table~\ref{tSynCRegularization} shows that $\twonorm{\vv_r}$ leads to better results. However, we find that learning with $\twonorm{\vv_r}$ converges much slower than with $\twonorm{\vw_c}$. Thus, we use $\twonorm{\vw_c}$ in our main experiments (though it puts our methods at a disadvantage).

\subsection{Learning Phantom Classes' Semantic Representations} So far we adopt the version of \method{SynC} that sets the number of base classifiers to be the number of seen classes $\cS$, and sets $\vb_r = \va_c$ for $r=c$. 
Here we study whether we can learn optimally the semantic representations for the phantom classes that correspond to base classifiers. The results in Table~\ref{tSynCLearnSpace} suggest that learning representations could have a positive effect.

\begin{table}
	\centering
	\caption{Effect of learning semantic representations.}
	\label{tSynCLearnSpace}
	\begin{tabular}{c|c|c|c}
		Datasets &Visual features & w/o learning & w/ learning\\ \hline
		\awa & GoogLeNet & 69.7\% & 71.1\% \\ \hline
		\cub & GoogLeNet & 53.4\% & 54.2\%\\ \hline
		\sun & GoogLeNet & 62.8\% & 63.3\%\\ \hline
	\end{tabular}
	\vskip -1em
\end{table}

\subsection{How Many Base Classifiers Are Necessary?} 
\begin{figure}
	\centering
	\includegraphics[width=0.95\columnwidth]{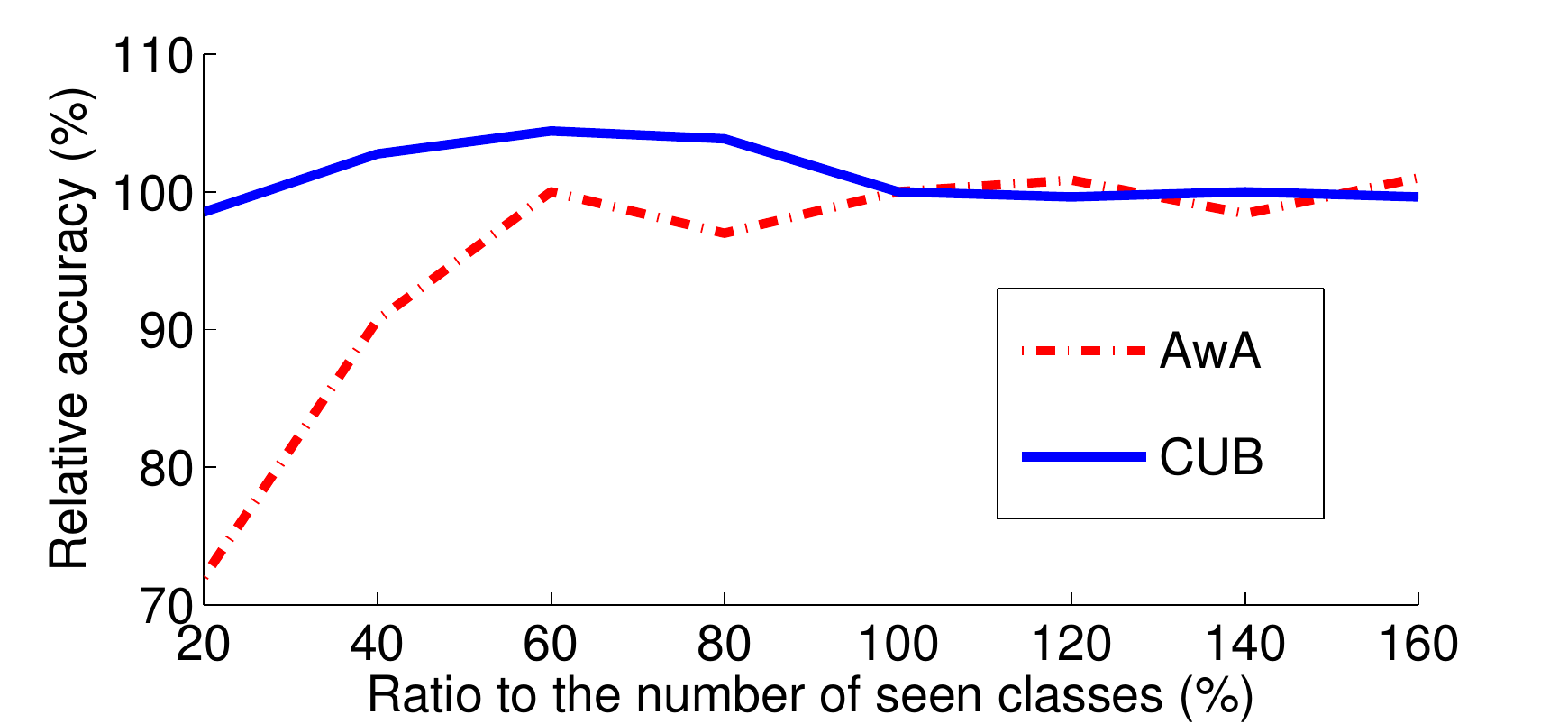}
	\caption{We vary the number of phantom classes $\cR$ as a percentage of the number of seen classes $\cS$ and investigate how much that will affect classification accuracy (the vertical axis corresponds to the ratio with respect to the accuracy when $\cR = \cS$). The base classifiers are learned with \method{SynC$^\textrm{o-vs-o}$}.}
	\label{fSynCNumPhantom}
	\vskip -1em
\end{figure}

In Fig.~\ref{fSynCNumPhantom}, we investigate how many base classifiers are needed --- so far, we have set that number to be the number of seen classes out of convenience. The plot shows that in fact, a smaller number ($\sim 60\%$) is enough for our algorithm to reach the plateau of the performance curve. Moreover, increasing the number of base classifiers does not seem to have an overwhelming effect.

Note that the semantic representations $\vb_r$ of the phantom classes are set equal to $\va_r, \forall r\in\{1, \cdots, \cR\}$ at 100\% (i.e., $\cR=\cS$). For percentages smaller than 100\%, we perform $K$-means and set $\vb_r$ to be the cluster centroids after $\ell_2$ normalization (in this case, $\cR = K$). For percentages larger than 100\%, we set the first $\cS$ $\vb_r$ to be $\va_r$, and the remaining $\vb_r$ as the random combinations of $\va_r$ (also with $\ell_2$ normalization on $\vb_r$).

\begin{figure}
\begin{center}
\includegraphics[width=0.95\columnwidth]{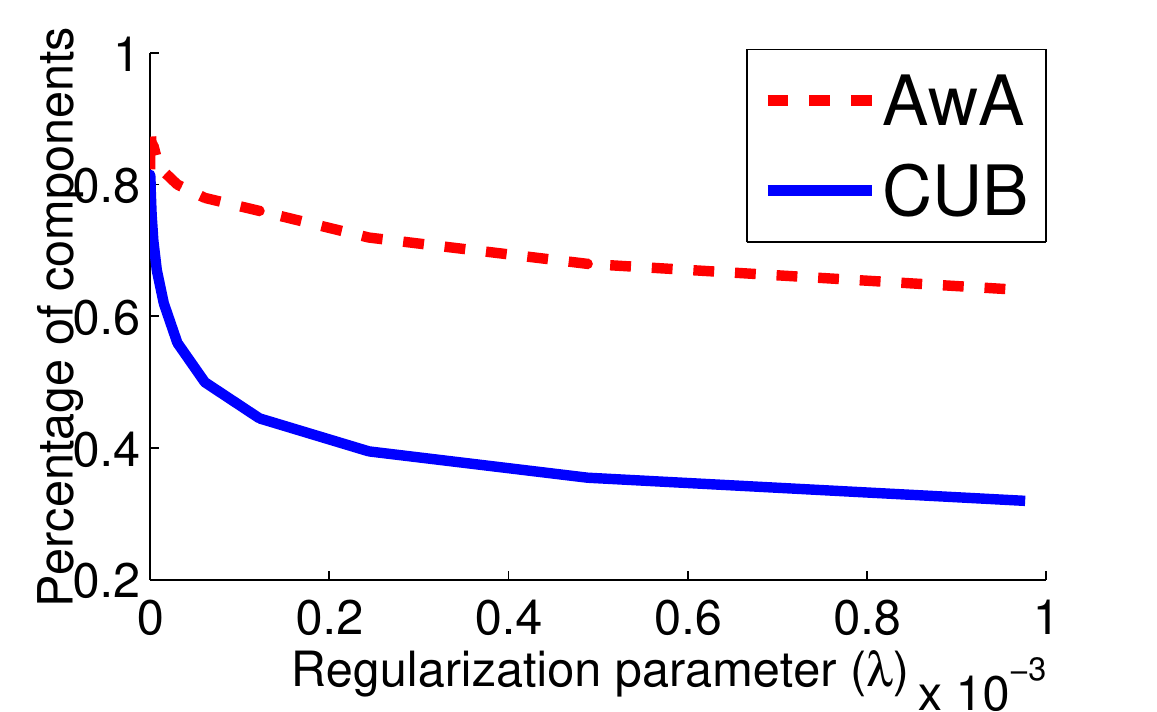}
\end{center}
\caption{Percentages of basis components required to capture 95\% of variance in classifier matrices for \awa and \cub.}
\label{fSynCBasis}
\end{figure}

We have shown that even by using fewer base (phantom) classifiers than the number of seen classes (e.g., around 60 \%), we get comparable or even better results, especially for \cub. We surmise that this is because \cub is a fine-grained recognition benchmark and has higher correlations among classes, and provide analysis in Fig.~\ref{fSynCBasis} to justify this.

We train one-versus-other classifiers for each value of the regularization parameter on both \awa and \cub, and then perform PCA on the resulting classifier matrices. We then plot the required number (in percentage) of PCA components to capture 95\% of variance in the classifiers. Clearly, \awa requires more. This explains why we see the drop in accuracy for \awa but not \cub when using even fewer base classifiers. Particularly, the low percentage for \cub in Fig.~\ref{fSynCBasis} implies that fewer base classifiers are possible. Given that \cub is a fine-grained recognition benchmark, this result is not surprising in retrospection as the classes are highly correlated.

\section{Analysis on EXEM}
\label{sSuppExpResAddExem}

In this section, we provide more analysis on \method{EXEM}. We focus on GoogLeNet features and the standard split (SS). We provide both qualitative and quantitative measures of predicted exemplars. We also investigate neural networks for exemplar prediction functions and the effect of PCA.

\subsection{Quality of Predicted Exemplars}

\begin{table}
\centering
\caption{We compute the Euclidean distance matrix between the \emph{unseen} classes based on semantic representations ($\mD_{\va_u}$), predicted exemplars ($\mD_{\vpsi(\va_u)}$), and real exemplars ($\mD_{\vv_u}$). Our method leads to $\mD_{\vpsi(\va_u)}$ that is better correlated with $\mD_{\vv_u}$ than $\mD_{\va_u}$ is. See text for more details.}
\vskip .5em
\label{tbExemCorrDis}
\begin{tabular}{c|c|c}
Dataset & \multicolumn{2}{c}{Correlation to $\mD_{\vv_u}$}  \\ \cline{2-3}
name & Semantic distances & Predicted exemplar\\
& $\mD_{\va_u}$ & distances $\mD_{\vpsi(\va_u)}$ \\ \hline
\awa & 0.862 & \textbf{0.897} \\ \hline
\cub & 0.777 $\pm$ 0.021 & \textbf{0.904} $\pm$ 0.026 \\ \hline
\sun & 0.784 $\pm$ 0.022 & \textbf{0.893} $\pm$ 0.019 \\ \hline
\end{tabular}
\vskip -1em
\end{table}

\begin{figure*}[!htb]
\centering
\includegraphics[width=0.48\textwidth]{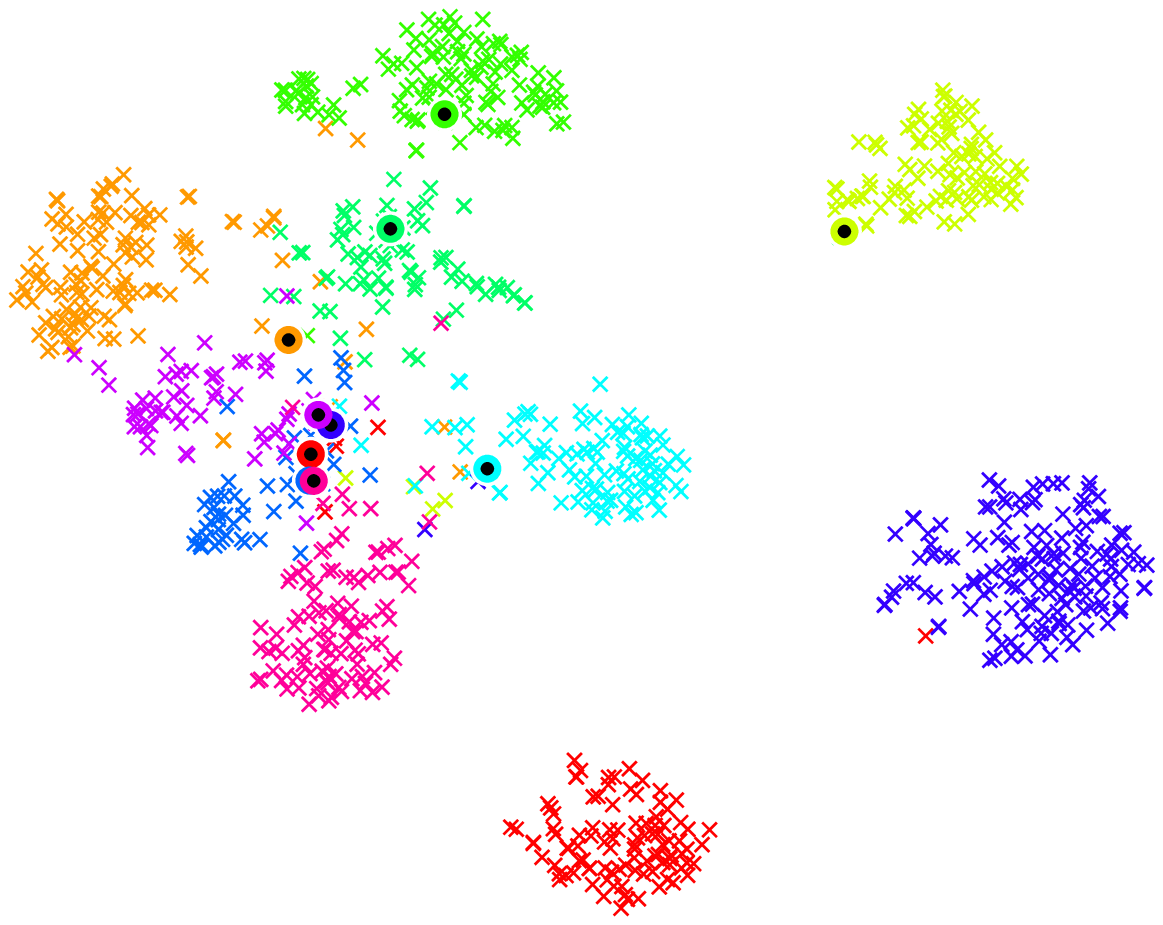}
\includegraphics[width=0.48\textwidth]{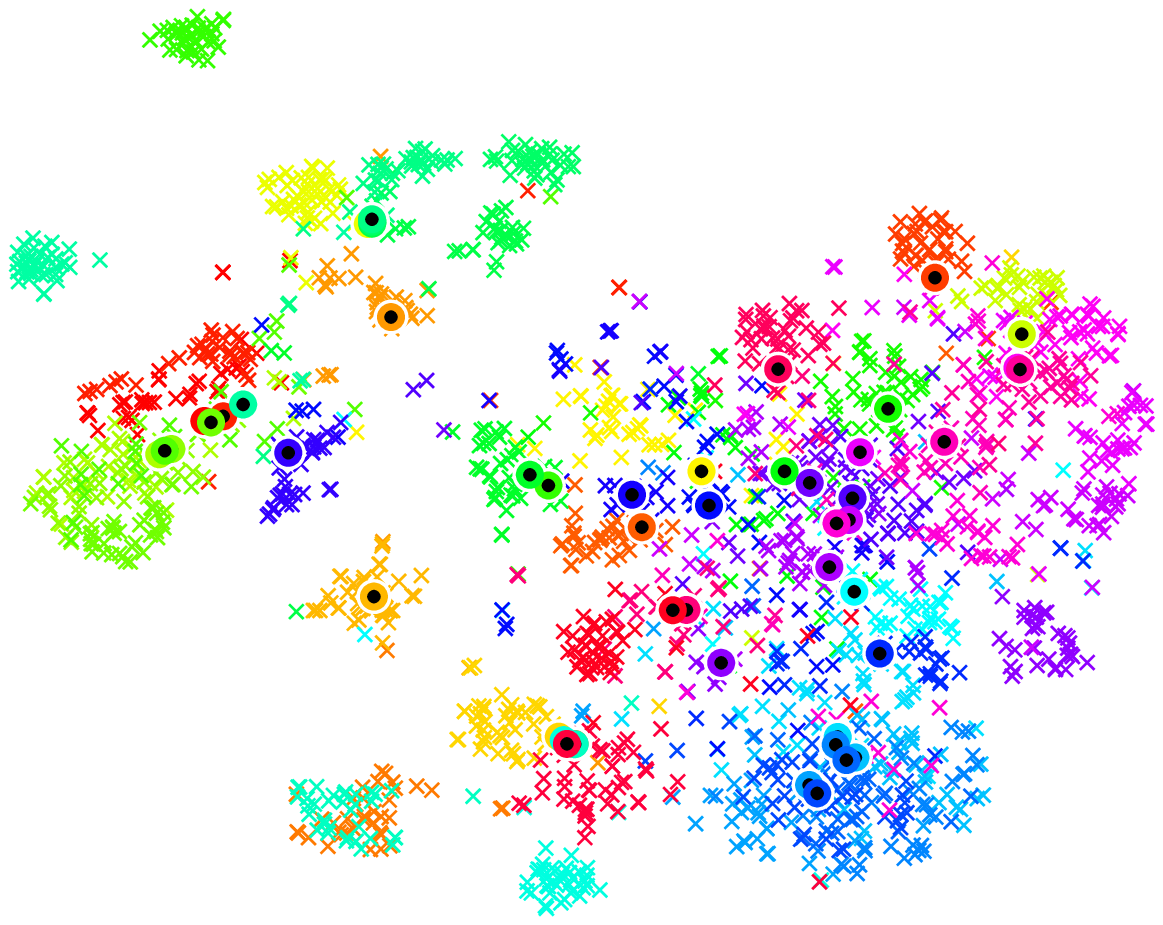}
\includegraphics[width=0.48\textwidth]{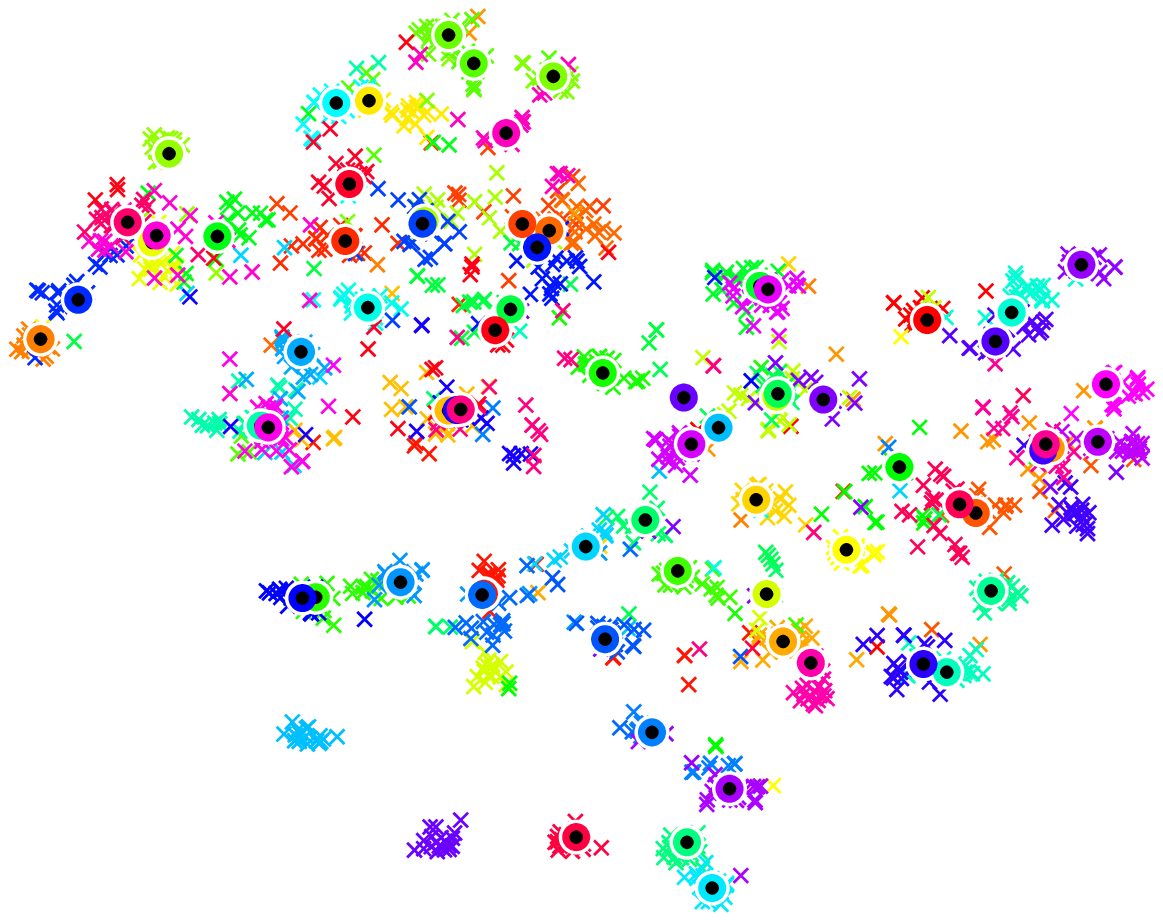}
\includegraphics[width=0.48\textwidth]{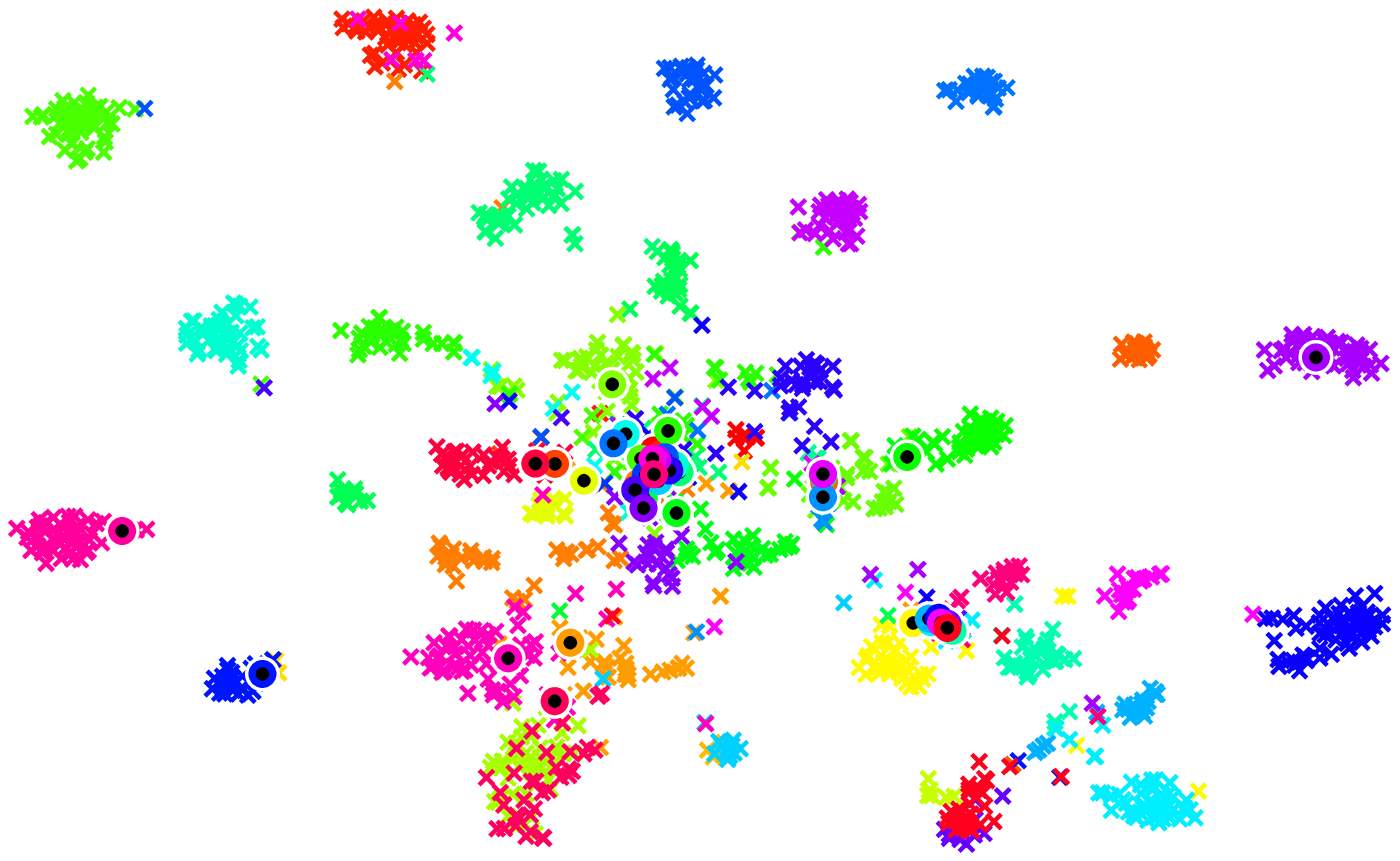}
\caption{t-SNE \cite{tSNE} visualization of randomly selected real images (crosses) and predicted visual exemplars (circles) for the \emph{unseen} classes on (from left to right, then from top to down) \awa, \cub, \sun, and \imn. Different colors of symbols denote different unseen classes. Perfect predictions of visual features would result in well-aligned crosses and circles of the same color. Plots for \textbf{CUB} and \textbf{SUN} are based on their first splits of SS. Plots for \imn are based on randomly selected 48 unseen classes from 2-hop and word vectors as semantic representations. Best viewed in color.} \label{fExemTsne}
\end{figure*}

We first show that predicted visual exemplars better reflect visual similarities between classes than semantic representations.
Let $\mD_{\va_u}$ be the pairwise Euclidean distance matrix between \emph{unseen} classes computed from semantic representations (i.e., $\cU$ by $\cU$), $\mD_{\vpsi(\va_u)}$ the distance matrix computed from predicted exemplars, and $\mD_{\vv_u}$ the distance matrix computed from real exemplars (which we do not have access to).
Table~\ref{tbExemCorrDis} shows that the Pearson correlation coefficient\footnote{We treat rows of each distance matrix as data points and compute the Pearson correlation coefficients between matrices.} between $\mD_{\vpsi(\va_u)}$ and $\mD_{\vv_u}$ is much higher than that between $\mD_{\va_u}$ and $\mD_{\vv_u}$. Importantly, we improve this correlation without access to any data of the unseen classes.

\begin{table}
\centering
\caption{Overlap of k-nearest classes (in \%) on \awa, \cub, \sun. We measure the overlap between those searched by real exemplars and those searched by semantic representations (i.e., attributes) or predicted exemplars. We set $K$ to be 40 \% of the number of unseen classes. See text for more details.} \label{tbExemKNN}
\begin{tabular}{c|c|c|c}
\text{Distances for kNN using} & \awa & \cub & \sun \\ 
& ($K$=4) & ($K$=20)  & ($K$=29) \\  \hline
Semantic representations & 57.5 & 68.9 & 75.2 \\ \hline
Predicted exemplars \hspace{2pt} & 67.5 & 80.0 & 82.1 \\ \hline
\end{tabular}
\end{table}

Besides the correlation used in Table~\ref{tbExemCorrDis}, we can also use \%kNN-overlap defined in Sect.~\ref{sExpResAddSem} as another evidence that predicted exemplars better reflect visual similarities (as defined by real exemplars) than semantic representations. Recall that \%kNN-overlap($A1$, $A2$) is the average (over all unseen classes $u$) of the percentages of overlap between two sets of k-nearest neighbors kNN$_{A1}(u)$ and kNN$_{A2}(u)$. In Table~\ref{tbExemKNN}, we report \%kNN-overlap (semantic representations, real exemplars) and \%kNN-overlap (predicted exemplar, real exemplars). We set $K$ to be 40\% of the number of unseen classes, but we note that the trends are consistent for different values of $K$.

We then show some t-SNE~\cite{tSNE} visualization of predicted visual exemplars of the \emph{unseen} classes. Ideally, we would like them to be as close to their corresponding real images as possible. In Fig.~\ref{fExemTsne}, we demonstrate that this is indeed the case for many of the unseen classes. For those unseen classes (each of which denoted by a color), their real images (crosses) and our predicted visual exemplars (circles) are well-aligned.

The quality of predicted exemplars (here based on the distance to the real images) depends on two main factors: the predictive capability of semantic representations and the number of semantic representation-visual exemplar pairs available for training, which in this case is equal to the number of seen classes $\cS$. On \awa where we have only $40$ training pairs, the predicted exemplars are surprisingly accurate, mostly either placed in their corresponding clusters or at least closer to their clusters than predicted exemplars of the other unseen classes. Thus, we expect them to be useful for discriminating among the unseen classes. On \imn, the predicted exemplars are not as accurate as we would have hoped, but this is expected since the word vectors are purely learned from text. 

We also observe relatively well-separated clusters in the semantic embedding space (in our case, also the visual feature space since we only apply PCA projections to the visual features), confirming our assumption about the existence of clustering structures.
On \cub, we observe that these clusters are more mixed than on other datasets. This is not surprising given that it is a fine-grained classification dataset of bird species.

\begin{table}
\centering
\caption{Comparison between \method{EXEM (1NN)} with support vector regressors (SVR) and with 2-layer multi-layer perceptron (MLP) for predicting visual exemplars. Results on \cub are for the first split of SS. Each number for MLP is an average over 3 random initialization.}
\vskip .5em
\label{tExemSVRvsMLP}
\begin{tabular}{c|c|c|c|c}
Dataset & Exemplar & No PCA & PCA & PCA  \\ 
name & predicted by &$\cst{d}$=1024  & $\cst{d}$=1024 & $\cst{d}$=500\\ \hline
\awa & SVR & \textbf{77.8} & 76.2 & \textbf{76.2}\\ \cline{2-5}
		 & MLP & 76.1$\pm$0.5 & \textbf{76.4}$\pm$0.1 & 75.5$\pm$1.7\\ \hline
\cub & SVR & \textbf{57.1} & \textbf{59.4} & \textbf{59.4} \\ \cline{2-5}
		 & MLP & 53.8$\pm$0.3 & 54.2$\pm$0.3 & 53.8$\pm$0.5\\ \hline
\end{tabular}
\end{table}

\subsection{Exemplar Prediction Function}

We compare two approaches for predicting visual exemplars: kernel-based support vector regressors (SVR) and 2-layer multi-layer perceptron (MLP) with ReLU nonlinearity.  
MLP weights are $\ell_2$ regularized, and we cross-validate the regularization constant.

Similar to \cite{ZhangXG17}, our multi-layer perceptron is of the form:
\begin{equation} \label{eqMLP}
\frac{1}{\cS} \sum_{c=1}^{\cS} \twonorm{\vv_c - \mW_2 \cdot \textrm{ReLU}(\mW_1 \cdot \va_c)} + \lambda \cdot R(\mW_1, \mW_2),
\end{equation}
where $R$ denotes the $\ell_2$ regularization, $\cS$ is the number of seen classes, $\vv_c$ is the visual exemplar of class $c$, $\va_c$ is the semantic representation of class $c$, and the weights $\mW_1$ and $\mW_2$ are parameters to be optimized. 

Following \cite{ZhangXG17}, we randomly initialize the weights $\mW_1$ and $\mW_2$, and set the number of hidden units for \textbf{AwA} and \textbf{CUB} to be 300 and 700, respectively. We use Adam optimizer with a learning rate 0.0001 and minibatch size of $\cS$. We tune $\lambda$ on the same splits of data as in other experiments with class-wise CV (Sect.~\ref{sSuppHyperTuning}). Our code is implemented in TensorFlow \cite{tensorflow}.

Table~\ref{tExemSVRvsMLP} shows that SVR performs more robustly than MLP.
One explanation is that MLP is prone to overfitting due to the small training set size (the number of seen classes) as well as the model selection challenge imposed by ZSL scenarios.
SVR also comes with other benefits; it is more efficient and less susceptible to initialization.

\subsection{Effect of PCA}

\begin{table*}
\centering
\caption{Accuracy of \method{EXEM (1NN)} on \awa, \cub, and \sun when predicted exemplars are from original visual features (No PCA) and PCA-projected features (PCA with $\cst{d}$ = 1024, 500, 200, 100, 50, 10).}
\label{tbExemPCA}
\begin{tabular}{c|c|c|c|c|c|c|c}
Dataset & No PCA & PCA & PCA & PCA & PCA & PCA & PCA \\ 
name & $\cst{d}$=1024  & $\cst{d}$=1024 & $\cst{d}$=500 & $\cst{d}$=200 & $\cst{d}$=100 & $\cst{d}$=50 & $\cst{d}$=10\\ \hline
\awa & \textbf{77.8} & 76.2 & 76.2 & 76.0 & 75.8 & 76.5 & 73.4\\ \hline
\cub & 55.1 & 56.3 & 56.3& \textbf{58.2} & 54.7 & 54.1 & 38.4\\ \hline
\sun & 69.2 & \textbf{69.6} & \textbf{69.6}& \textbf{69.6} & 69.3 & 68.3 & 55.3\\ \hline
\end{tabular}
\end{table*}
   
Table~\ref{tbExemPCA} investigates the effect of PCA. In general, \method{EXEM (1NN)} performs comparably with and without PCA.
Moreover, we see that our approach is extremely robust, working reasonably over a wide range of (large enough) $\cst{d}$ on all datasets.
Clearly, a smaller PCA dimension leads to faster computation due to fewer regressors to be trained.

\end{appendices}

\bibliographystyle{ieee}
\footnotesize
\bibliography{main}   

\end{document}